\documentclass[10pt]{article} 
\usepackage[accepted]{tmlr}

\usepackage{amsmath,amsfonts,bm}

\def\eqref#1{equation~\ref{#1}}

\def\1{\bm{1}}

\DeclareMathAlphabet{\mathsfit}{\encodingdefault}{\sfdefault}{m}{sl}
\SetMathAlphabet{\mathsfit}{bold}{\encodingdefault}{\sfdefault}{bx}{n}

\DeclareMathOperator*{\argmin}{arg\,min}

\DeclareMathOperator{\sign}{sign}

\usepackage{hyperref}
\usepackage{url}

\usepackage{graphicx}
\usepackage{subfigure}
\usepackage{amsmath}
\usepackage{amsthm}
\usepackage{amssymb}
\usepackage{booktabs}
\usepackage[T1]{fontenc}
\usepackage{dsfont}

\usepackage{color,xcolor}
\definecolor{R3_color}{rgb}{0.00, 0.00, 1.00}

\newtheorem{theorem}{Theorem}

\newtheorem{proposition}[theorem]{Proposition}

\newcommand{\bE}{\mathbb{E}}

\newcommand{\cF}{\mathcal{F}}
\newcommand{\cX}{\mathcal{X}}
\newcommand{\cY}{\mathcal{Y}}
\newcommand{\cR}{\mathcal{R}}

\newcommand{\bx}{{x}}
\newcommand{\bxprime}{{x}'}

\newcommand{\epsball}{\mathcal{B}_\epsilon}
\newcommand{\xadv}{\tilde{{x}}}
\newcommand{\yadv}{\tilde{y}}

\newcommand{\CW}{C$\&$W$_{\infty}$}
\usepackage{algorithm}
\usepackage{etoolbox}
\let\classAND\AND
\let\AND\relax
\usepackage{algorithmic}

\let\AND\classAND
	\AtBeginEnvironment{algorithmic}{\let\AND\algoAND}
\usepackage{multirow}

\title{NoiLIn: Improving Adversarial Training \\and Correcting Stereotype of Noisy Labels}

\author{\name Jingfeng Zhang\thanks{Equal contribution} \email jingfeng.zhang@riken.jp\\
      \addr RIKEN Center for Advanced Intelligence Project (AIP)      \AND
      \name Xilie Xu$^{*}$ \email xuxilie@comp.nus.edu.sg \\
      \addr School of Computing, National University of Singapore
      \AND
      \name Bo Han\email bhanml@comp.hkbu.edu.hk\\
      \addr Department of Computer Science, Hong Kong Baptist University
      \AND
      \name Tongliang Liu \email tongliang.liu@sydney.edu.au \\
      \addr Trustworthy Machine Learning Lab, University of Sydney
      \AND    
      \name Lizhen Cui \email clz@sdu.edu.cn \\
      \addr School of Software $\&$ Joint SDU-NTU Centre for Artificial Intelligence Research (C-FAIR), Shandong University
      \AND      
      \name Gang Niu\email gang.niu@riken.jp\\
      \addr RIKEN Center for Advanced Intelligence Project (AIP)
      \AND
      \name Masashi Sugiyama\email sugi@k.u-tokyo.ac.jp\\
      \addr RIKEN Center for Advanced Intelligence Project (AIP) \\
      Graduate School of Frontier Sciences, the University of Tokyo
    }

\def\month{06}  
\def\year{2022} 
\def\openreview{\url{https://openreview.net/forum?id=zlQXV7xtZs}} 

\begin{document}

\maketitle

\begin{abstract}

{Adversarial training (AT) formulated as the minimax optimization problem can effectively enhance the model's robustness against adversarial attacks. 
The existing AT methods mainly focused on manipulating the inner maximization for generating quality adversarial variants or manipulating the outer minimization for designing effective learning objectives. 
However, empirical results of AT always exhibit the robustness at odds with accuracy and the existence of the cross-over mixture problem, which motivates us to study some label randomness for benefiting the AT.}
First, we thoroughly investigate noisy labels (NLs) injection into AT's inner maximization and outer minimization, respectively and obtain the observations on when NL injection benefits AT. 
Second, based on the observations, we propose a simple but effective method---NoiLIn that randomly injects NLs into training data at each training epoch and dynamically increases the NL injection rate once robust overfitting occurs. 
Empirically, NoiLIn can significantly mitigate the AT's undesirable issue of robust overfitting and even further improve the generalization of the state-of-the-art AT methods.
Philosophically, NoiLIn sheds light on a new perspective of learning with NLs: NLs should not always be deemed detrimental, and even in the absence of NLs in the training set, we may consider injecting them deliberately.
Codes are available in \text{https://github.com/zjfheart/NoiLIn}.

\end{abstract}

\section{Introduction}
\label{sec:intro}
Security-related areas require deep models to be robust against \textit{adversarial attack}~\citep{szegedy}. To obtain \textit{adversarial robustness}, \textit{adversarial training} (AT)~\citep{Madry_adversarial_training} would be currently the most effective defense that has so far not been comprehensively compromised~\citep{Athalye_ICML_18_Obfuscated_Gradients}, in which AT is formulated as a minimax optimization problem with the \textit{inner maximization} to generate adversarial data within small neighborhoods of their natural counterparts  and the \textit{outer minimization} to learn from the generated adversarial data.   
\begin{figure*}[t!]
	\centering
	\vspace{-0mm}
	\includegraphics[width=0.80\linewidth]{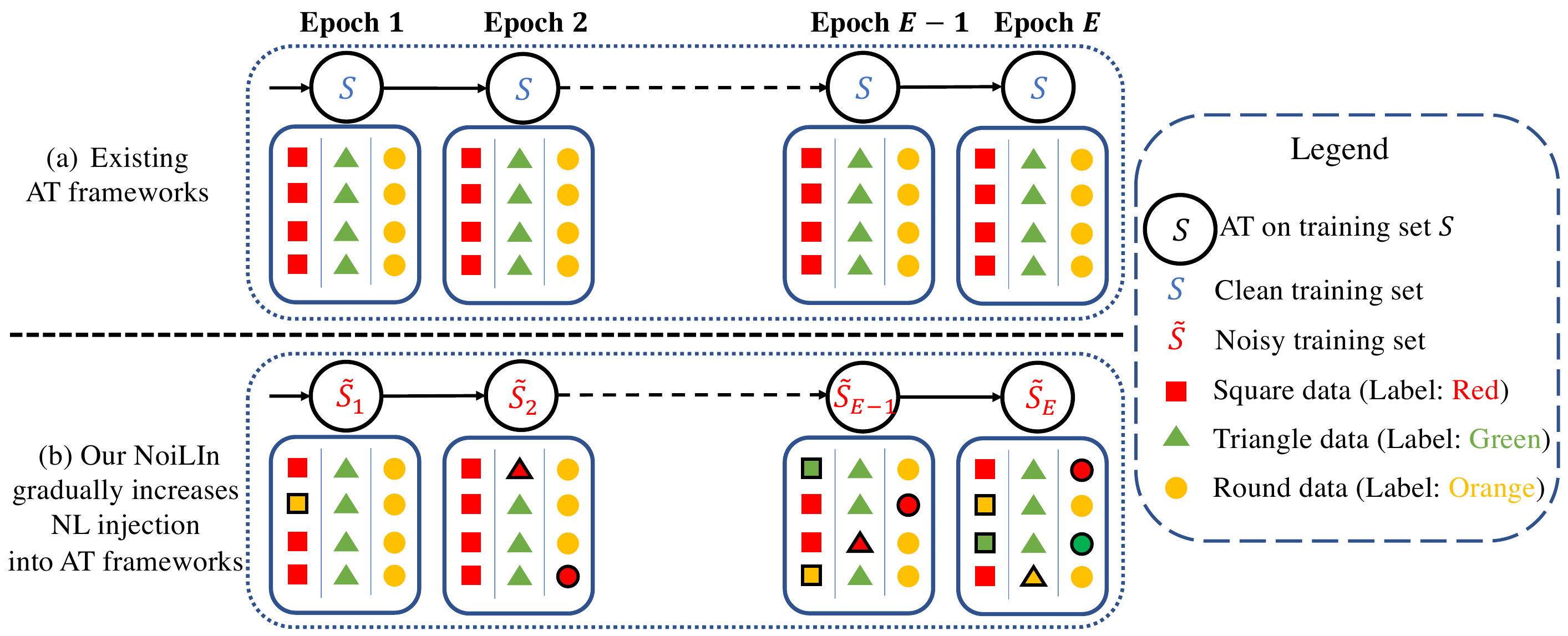}
	\vspace{-2mm}
	\caption{Comparisons between the existing AT framework (panel (a)) and our AT-NoiLIn framework (panel (b)). At each training epoch, AT-NoiLIn randomly flips labels of a portion of data (marked in the black edges), generating noisy-label training set $\tilde{S}$ for the learning. 
	}
\vspace{-2mm}
	\label{fig:alg_illustrate}
\end{figure*}

{However, AT empirically shows a severe tradeoff between the natural accuracy of natural test data and the robust accuracy of adversarial test data~\citep{tsipras19_robustness_at_odd}. 
Besides, \cite{zhang2020fat} showed AT has a \textit{cross-over mixture problem}, where the adversarial variants of the natural data cross over the decision boundary and fall in the other-class areas. 
Even, \cite{sanyal2020benign} showed the benchmark dataset (such as CIFAR~\citep{krizhevsky2009learning_cifar10}) contains some noisy data points. Furthermore. \cite{donhauser2021interpolation} showed an unorthodox way to yield an estimator with a smaller robust risk, i.e., flipping the labels of a fixed fraction of the training data.
}


{The above facts motivate us to explore manipulating labels for benefits that were largely overlooked by the existing AT studies.} 
To enhance adversarial robustness further, the existing AT methods have mainly focused on manipulating the inner maximization or the outer minimization.
For example, by manipulating the inner maximization for generating quality adversarial data, \cite{zhang2020fat} generated \textit{friendly adversarial data} that are near the decision boundaries but are wrongly classified. 
On the other hand, by manipulating the outer minimization for designing the model's loss functions for the learning, \cite{Zhang_trades} designed TRADES that has two loss terms, i.e., the cross entropy loss on the natural data and the Kullback-Leibler (KL) divergence loss on the adversarial data. Since \textit{NLs were often deemed to hurt the training}~\citep{angluin1988learning}, little efforts has been made to investigate noisy labels (NLs) in AT



To correct this negative stereotype of NLs, firstly, this paper thoroughly investigates NL injections in AT's every component.  $(i)$ We inject NLs in inner maximization. In each training minibatch, we randomly choose a portion of data whose adversarial variants are generated according to the flipped labels; the labels in outer minimization are intact for the learning. 
The label-flipped adversarial variants are far from the decision boundaries, thus the label-flipped adversarial data become no longer adversarial to the current model. In other words, label-flipped adversarial data serve more like natural data. As a result, when AT learns label-flipped adversarial data, the deep model's generalization gets increased but the robustness gets decreased.%
%
%
$(ii)$ We inject NLs in outer minimization. We generate the adversarial data in each training minibatch according to the intact labels but randomly flip a portion of labels for the learning. 
NL injection in outer minimization leads to high \textit{data diversity} for the learning, i.e., the deep model learns from different labels sets over the training process. From another perspective, NL injection in outer minimization implicitly averages deep models trained with different label sets. 
Therefore, NLs serve as regularization that can largely alleviate the AT's undesirable issue of robust overfitting~\citep{rice2020overfitting} and even slightly enhances AT's best-epoch-checkpoint robustness.
$(iii)$ We inject NLs in both inner maximization and outer minimization. In each training epoch, we randomly choose a portion of data to flip their labels; then, we conduct standard adversarial training (SAT~\citep{Madry_adversarial_training}) on the noisy set. In other words, adversarial variants of this portion of data are generated according to the flipped labels, and their flipped labels are also used for the learning. The empirical results are very similar to case $(ii)$. 

Inspired by the above three observations, secondly, we propose a simple method ``\textit{NoiLIn}'' that dynamically adjusts NL injection into AT (see Figure~\ref{fig:alg_illustrate} for illustrations). In each training epoch, we first randomly flip a portion of labels of the training set, and then execute an AT method (such as SAT, TRADES~\citep{Zhang_trades}, WAP~\citep{wu2020adversarial}) using the noisy-label set.
As the training progresses, we increase the flipping portion if there occurs robustness degradation (evaluated on a validation set). 


Our contributions are summarized as follows.
(a) We study NLs from a novel perspective, i.e., injecting NLs into the AT's training process; we observe that NL injection can even benefit AT. 
(b) Based on our observations, we propose to gradually increase the NL injection rate over the training process (i.e., NoiLIn). 
NoiLIn can significantly relieve the issue of robust overfitting of some AT methods such as SAT and TRADES. 
Using large-scale Wide ResNet~\cite{zagoruyko2016WRN} on the CIFAR-10 dateset~\cite{krizhevsky2009learning_cifar10}, NoiLIn can even improve the generalization of current state-out-of-art (SOTA) TRADES-AWP~\cite{wu2020adversarial} by surprisingly $2.2\%$ while maintaining its robustness (measured by auto attack (AA)~\cite{croce2020reliable}). 
(c) Philosophically, we should not always consider NLs to be detrimental. Even in the absence of NLs in the training set, we may consider injecting them deliberately.

\section{Background and Related Work}
There are many ways to improve the model's adversarial robustness, such as robustness certification~\citep{hein2017formal,Weng_Evaluating_robustness_extreme_value_approach,Eric_Wong_provable_defence_convex_polytope,icml/MirmanGV18_internal_bound_propagation,Tsuzuku_Lipschitz_margin_training_scalable_certification,Lecuyer_certified_robustness_with_DP,DBLP:conf/iclr/XiaoTSM19,Jeremy_cohen_certified_robustness_random_smoothing,balunovic2020adversarial_certified_robustness,zhang2020towards_certifiable,icml_20_second_order,zhang2021towards}, Liptschiz regularization~\citep{cisse2017parseval,moosavi2019robustness_curvature,local_linearilization}, incorporating attack modules~\citep{yan2018deep}, and detecting adversarial data~\citep{DBLP:conf/iclr/MetzenGFB17,DBLP:conf/iccv/LiL17,DBLP:conf/ccs/Carlini017,DBLP:conf/aaai/TianYC18,DBLP:conf/iclr/Ma0WEWSSHB18,DBLP:conf/nips/LeeLLS18,DBLP:conf/nips/PangDDZ18,DBLP:conf/uai/SmithG18,DBLP:conf/icml/RothKH19,DBLP:conf/cvpr/LiuZZHLZY19,DBLP:conf/iclr/YinKR20,DBLP:conf/eurosp/SperlKCLB20,DBLP:conf/cvpr/CohenSG20,DBLP:conf/iclr/SheikholeslamiL21,conf/icassp/ChenC00MZY21,DBLP:conf/aaai/YangCHWJ20,DBLP:conf/iclr/QinFSRCH20,DBLP:conf/aaai/Tian0LD21,DBLP:conf/aaai/WuAWY21}. Nevertheless, adversarial training (AT) still stands out as the most effective defense against the adaptive and strong attacks, attracting significant attention. This section briefly reviews the background of AT and provides related work. 

\paragraph{Adversarial Training (AT)}
AT aims to a) classify the natural data  $x$ correctly, and b) make the decision boundaries ``thick'' so that no data are encouraged to fall nearby the decision boundaries. AT's learning objective is as follows.  

Let $(\cX,L_{\infty})$ be the input feature space $\cX$ with the infinity distance metric $L_{\infty}(\bx,\bxprime)=\|\bx-\bxprime\|_\infty$, and
\begin{equation}
	\label{eq:perturbation_ball}
	\epsball[\bx] = \{\bxprime \in \cX \mid L_{\infty}(\bx,\bx')\le\epsilon\} 
\end{equation}
be the closed ball of radius $\epsilon>0$ centered at $\bx$ in $\cX$.
Given a dataset $S = \{ ({x}_i, y_i)\}^n_{i=1}$, where ${x}_i \in \cX$ and $y_i \in \cY =  \{0, 1, \ldots, C-1\}$,
AT's learning objective is formulated as a minimax optimization problem, i.e.,
\begin{equation}
	\label{eq:adv_obj}
	\underbrace{\min_{f\in\cF} \frac{1}{n}\sum_{i=1}^n \ell(f(\xadv_i),y_i)}_{\text{outer minimization}},\quad
	\underbrace{ \xadv_i = \arg\max\nolimits_{\xadv\in\epsball[\bx_i]} \ell(f(\xadv),y_i)}_{\text{inner maximization}}.
\end{equation}
Eq.~(\ref{eq:adv_obj}) implies the AT's realization with one step of \textit{inner maximization} on generating adversarial data $\xadv_i$ and one step of \textit{outer minimization} on fitting the model $f$ to the generated adversarial data $(\xadv_i,y_i)$. 
\paragraph{Projected Gradient Descent (PGD)}
To generate adversarial data, SAT~\cite{Madry_adversarial_training}\footnote{{Throughout this paper, we use AT to denote adversarial training methods in general and use SAT to specify the standard AT method by \cite{Madry_adversarial_training}. }} uses PGD to approximately solve the inner maximization.
SAT formulates the problem of finding adversarial data as a constrained optimization problem. Namely, given a starting point ${x}^{(0)} \in \cX$ and step size $\alpha > 0$, PGD works as follows:
\begin{equation}
	\label{PGD-k}
	{x}^{(t+1)} = \Pi_{\mathcal{B}[{x}^{(0)}]} \big( {x}^{(t)} +\alpha \sign (\nabla_{{x}^{(t)}} \ell(f_{\theta}({x}^{(t)}), y )  )  \big ) , \forall { t \geq 0}
\end{equation}
until a certain stopping criterion is satisfied. Commonly, the criterion can be a fixed number of iterations $K$. In Eq.~\eqref{PGD-k}, $\ell$ is the loss function in Eq.~\eqref{eq:adv_obj}; ${x}^{(0)}$ refers to natural data or natural data corrupted by a small Gaussian or uniform random noise; $y$ is the corresponding true label; ${x}^{(t)}$ is adversarial data at step $t$; and $\Pi_{\epsball[{x}_0]}(\cdot)$ is the projection function that projects the adversarial data back into the $\epsilon$-ball centered at ${x}^{(0)}$ if necessary. 

\paragraph{Broad Studies on AT} Existing literature studied/improved AT in many aspects such as customizing the inner maximization for simulating the better adversary~\citep{Goodfellow14_Adversarial_examples,Ali_NIPS19_adversarial_training_for_free,Lu_yiping_NIPS19_yopo,wong2020fast_zico_kolter,Babu_cvpr_2020,Cai_CAT,Wang_Xingjun_MA_FOSC_DAT,zhang2020fat,sriramanan2020guided} or the outer minimization for designing better loss functions/regularizer~\citep{farnia2018generalizable,song_iclr2019_domain_adaptation,Zhang_trades,ding2020mma,wang2020improving_MART,wu2020adversarial,zhang2021geometryaware,pang2021bag,chen2021robust}, designing/searching robust network structures~\citep{xie2020smooth,DBLP:conf/iclr/XieY20,guo2020meets,sehwag2020hydra,pmlr-v139-yan21e,pmlr-v139-du21f,li2021neural,huang2021exploring,li2021neural}, employing multiple models~\citep{Pang_ICML_19_AT_Ensemble,yang2020dverge}, augmenting the training data~\citep{Tramer_iclr_18,schmidt2018adversarial_more_data,DeepMind_useto,carmon2019unlabeled,najafi2019robustness,song_iclr20_RBS,Lee_2020_CVPR,Gong_2021_CVPR,gowal2021improving,wu2020wider}, and analyzing AT's intriguing properties~\citep{tsipras19_robustness_at_odd,robust_features_nips2019_madry,stutz2019disentangling,Dong_Yin_adv_gen,rice2020overfitting,zhang2019interpreting,raghunathan2020understanding,yang2020closer}. 
Besides, recent studies showed AT could benefit other domains such as pre-training\citep{chen_CVPR_pretrain,Jiang_NIPS_pretrain}, out-of-distribution generalization~\citep{yi2021improved},  inpainting~\citep{khachaturov2021markpainting}, interpretability~\citep{ross2018improving}, fairness~\citep{xu2021robust} and so on.
%
%
%
%
%
%
%
%
%

\paragraph{Interaction between NLs and AT}
\label{bg:NL}
To avoid confusion, we clarify the differences between this paper and the existing studies of AT on NLs. 
NLs practically exist in the training set~\citep{natarajan2013learning}, and therefore, some studies have explored the effects of AT on NLs. 
\citet{DeepMind_useto} showed that NLs degraded both generalization and robustness of AT, but robustness suffers less from NLs than generalization. 
\citet{sanyal2020benign} showed robust training avoids memorization of NLs. Furthermore, \citet{zhu2021understanding} showed AT has a smoothing effect that can naturally mitigate the negative effect of NLs. 
Nevertheless, all those studies assumed NLs exist in the training set, which is detrimental to AT. 
In comparison, we assume the training set is noise-free. We treat NLs as friends and deliberately inject NLs to benefit AT in terms of relieving robust overfitting of SAT~\citep{Madry_adversarial_training} and TRADES~\citep{Zhang_trades}, even improving TRADES-AWP~\citep{wu2020adversarial}'s generalization while maintaining its peak robustness. 

\paragraph{Relation with DisturbLabel~\citep{xie2016disturblabel}} DisturbLabel, the most relevant work to NoiLIn, also randomly selects a small subset of data in each training epoch and then sets their ground-truth labels to be incorrect. 
\citet{xie2016disturblabel}  studied only standard training (ST) for generalization. Differently, our work focuses on AT's robustness and generalization. Besides, compared with DisturbLabel that injects a small ratio NLs (20$\%$ or less) to alleviate ST's overfitting, AT needs injecting larger NLs ratio (e.g., 40$\%$ or more) to alleviate the AT's robust overfitting (see experiments in Figure~\ref{fig:robust-overfitting}). Compared with ST, AT encounters worsen situations of data overlaps of different classes, thus requiring stronger label randomness. 

\paragraph{Relation with Memorization in AT~\citep{dong2022exploring}} The independent and concurrent work by \cite{dong2022exploring} explored various AT methods (i.e., training stability of SAT and TRADES methods) under completely random labels.  Besides, \cite{dong2022exploring} proposed to mitigate SAT's and TRADES's issue of robust overfitting via adding temporal ensembling (TE)~\citep{laine2016temporal} as an additional learning objective to penalize the overconfident predictions. This shares a similar idea to injecting learned smoothing~\citep{chen2021robust} (i.e., adding several additional learning objectives) for mitigating the robust overfitting.
We argue that our NoiLIn is simpler but is no worse than the above peer methods. TE and smoothness injection methods~\citep{dong2022exploring,chen2021robust} both need add additional learning objectives, which introduce more hyperparameters than NoiLIn's for finetuning the performance. Besides, without introducing additional learning objectives, NoiLIn saves computational resources. 

{\paragraph{Relation with Label Manipulations  Benefiting Robustness.}  
There are debates on whether label smoothing and logit squeezing~\citep{shafahi2019label}, and logit pairing~\citep{engstrom2018evaluating,mosbach2018logit} genuinely benefit adversarial robustness or mask gradients for overly reporting robustness. This paper avoids such a debate by evaluating our NoiLIn method using the strongest AutoAttack (AA)~\citep{croce2020reliable}, which should thwart the concerns of gradient obfuscations~\citep{Athalye_ICML_18_Obfuscated_Gradients}.}

\section{A Closer Look at NL injection in AT}
\label{sec:NL_AT}

\begin{figure}[tp!]
	\centering
	\vspace{-0mm}
	\subfigure [CIFAR-10 dataset]{
			\includegraphics[width=0.25\textwidth]{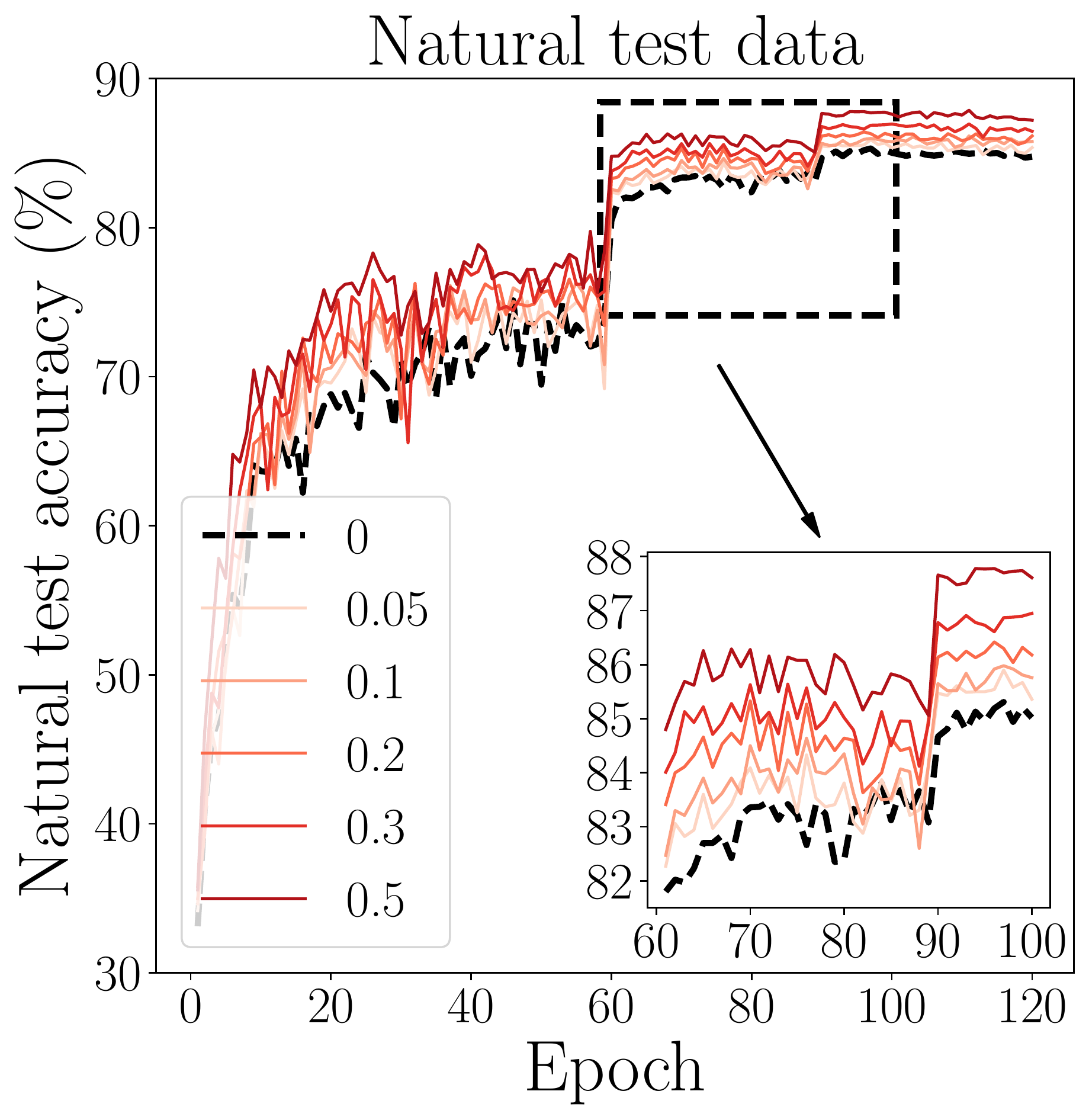}
		\includegraphics[width=0.25\textwidth]{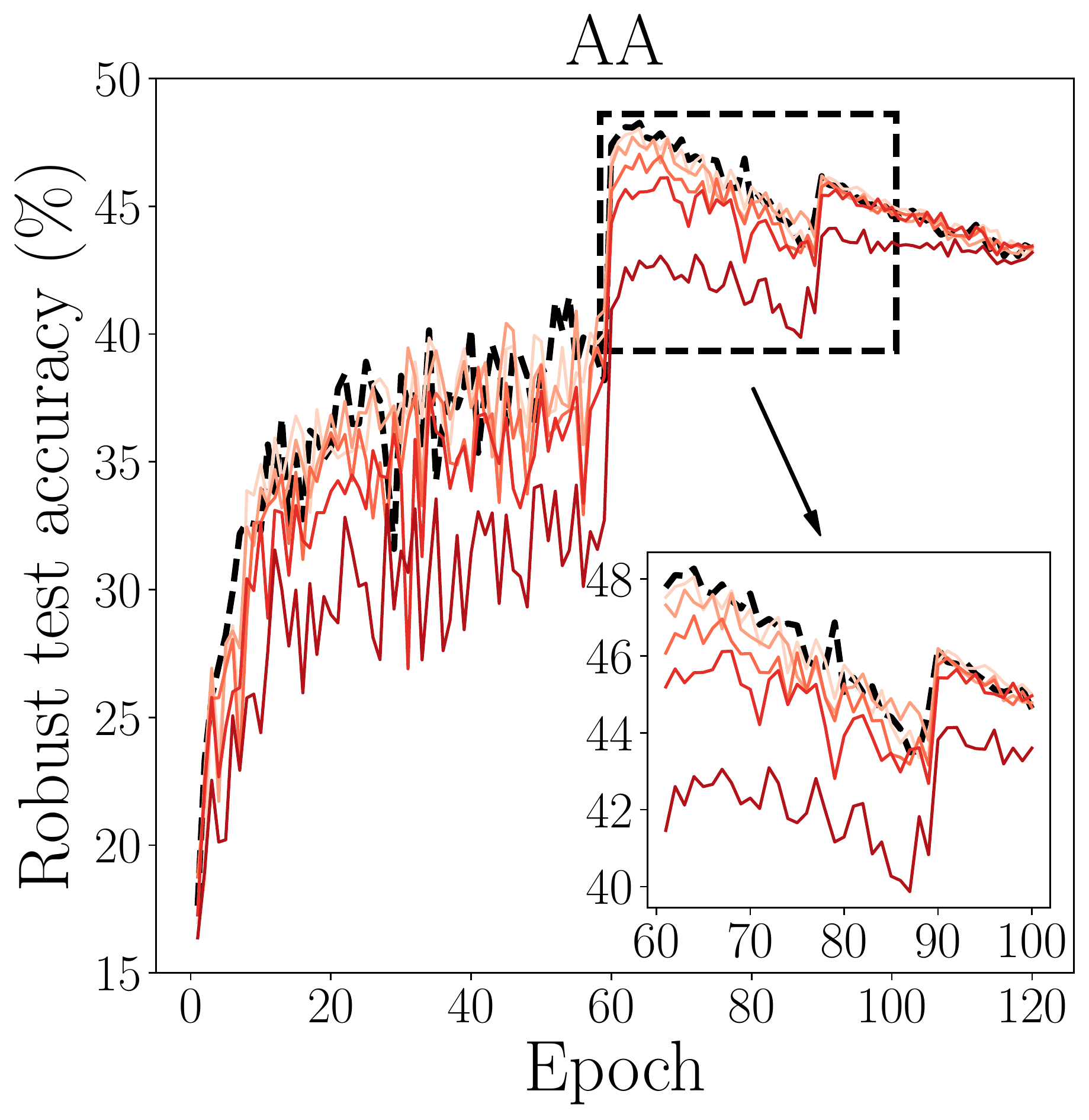}

		\label{fig:inner_NC_symmetric}
	}
	\subfigure[Supporting schematics]{
	\includegraphics[width=0.35\textwidth]{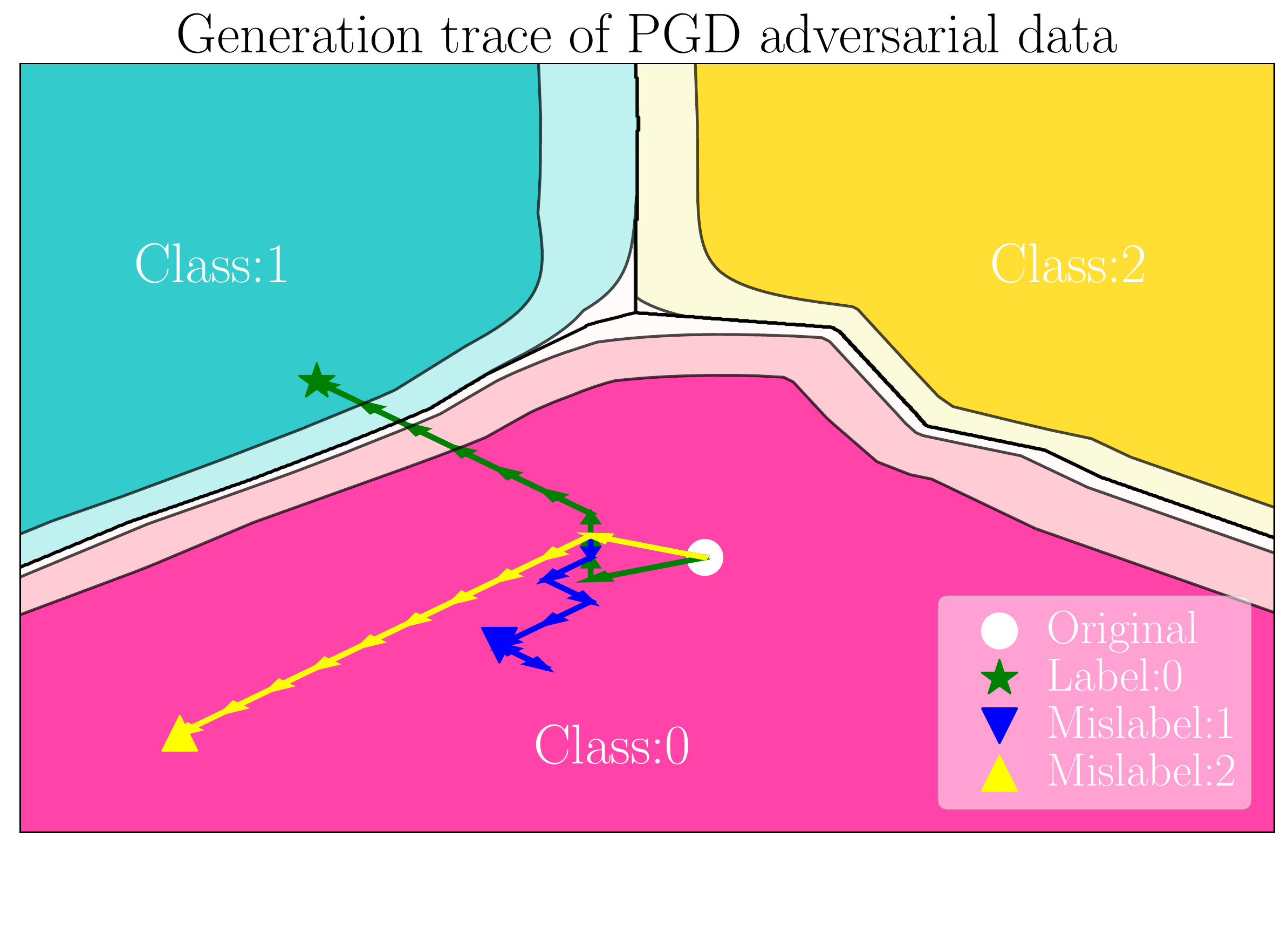}
		\label{fig:NL_inner_toy_example}
	}
	\vspace{-2mm}
	\caption{Figure~\ref{fig:inner_NC_symmetric} shows the learning curves of injecting various levels of NL in inner maximization. (The number in the legend represents NL injection rate $\eta$; the black dash line represents SAT.) 
		Figure~\ref{fig:NL_inner_toy_example} shows the generation trace of adversarial data on a synthetic ternary classification. (The white round denotes natural data. The blue trace is generated with correct label; both yellow and blue traces are generated with noisy label (NL). The color gradient represents the prediction confidence: the deeper color represents higher prediction confidence.) 
		NL injection in inner maximization prevents generating effectively adversarial data.	 }
	\vspace{-2mm}
\end{figure}

Before introducing our algorithm---NoiLIn, we get started with a thorough understanding of NL injection in AT. Specifically, AT (see Eq.~(\ref{eq:adv_obj})) has two label-related parts: one is inner maximization for generating adversarial data; another is outer minimization for learning the generated adversarial data. In this section, we inject NLs in the two parts separately and obtain some intriguing observations. 

\subsection{NL Injection in Inner Maximization}
\label{sec:NL_inner_experiment}
\paragraph{Experiment details}
We conducted experiments of injecting NLs in inner maximization on the CIFAR-10 dataset~\cite{krizhevsky2009learning_cifar10}. 
In Figure~\ref{fig:inner_NC_symmetric}, we compare AT with NLs in inner maximization (red lines with different color degrees) and SAT (black dashed lines). We inject symmetric-flipping NLs~\cite{van2015learning}, where labels $y$ are flipped at random with the uniform distribution. 
In each training mini-batch, we randomly flip $\eta$ portion of labels of training data; then, the adversarial data are generated according to the flipped labels. The noise rate $\eta$ is chosen from $\{0, 0.05, 0.1, 0.2, 0.3, 0.5\}$. Note that it is exactly SAT when $\eta=0$. 
The perturbation bound is set to $\epsilon_{\mathrm{train}} = 8/255$; 
the number of PGD steps is set to $K=10$, and the step size is set to $\alpha=2/255$.
The labels of outer minimization are intact for the learning.
We trained ResNet-18~\cite{he2016deep} using stochastic gradient descent (SGD) with 0.9 momentum for 120 epochs with the initial learning rate of 0.1 divided by 10 at Epochs 60 and 90, respectively.

We evaluate the robust models based on the two evaluation metrics, i.e., \textit{natural test accuracy} on natural test data and \textit{robust test accuracy} on adversarial data generated by AutoAttack (AA)~\cite{croce2020reliable}, respectively. 
The adversarial test data are bounded by $L_{\infty}$ perturbations with $\epsilon_{\mathrm{test}}=8/255$. 

For completeness, in Appendix~\ref{appendix:inner_NL}, we demonstrate other attacks (such as PGD and CW attacks), and we also inject \textit{pair-flipping} NLs~\cite{han2018co}, where labels $y$ are flipped between adjacent classes that are prone to be mislabeled.


\vspace{-0mm}
\paragraph{Observation ($i$)} \textit{NL injection in inner maximization improves AT's generalization but degrades AT's robustness.} 
As shown in Figure~\ref{fig:inner_NC_symmetric}, with the increasing of $\eta$, AT's generalization consistently increases (left panel), and AT's robustness consistently decreases (right panel). 

Figure~\ref{fig:NL_inner_toy_example} illustrates the effects of NLs on generating adversarial data. 
We draw the generation traces of PGD adversarial data on a two-dimensional ternary classification dataset. 
We randomly choose a start point (white dot) whose correct label is ``0'' and plot the generation traces of adversarial data using correct label ``0'' (green trace), wrong label ``1'' (blue trace), and wrong label ``2'' (yellow trace), respectively. We find that NLs (wrong label ``1'' or ``2'') mislead the generation traces that should have pointed at the decision boundary. Instead, generation traces with NLs fall in the internal areas of class ``0''. Therefore, the generated NL-adversarial data are similar to the natural data of class ``0''.
To further justify the above phenomenon, Figure~\ref{fig:NL_inner_cifar} (in Appendix~\ref{appendix:inner_NL}) uses the CIFAR-10 dataset to corroborate that label-flipped adversarial data are similar to natural data. 

Therefore, the reasons of Observation ($i$) are as follows. 
When PGD (see Eq.~\ref{PGD-k}) generates adversarial data, NLs mislead PGD to find wrong directions to the decision boundary. Then, the label-flipped adversarial data becomes no longer adversarial to the current model, and they serve more like their natural counterparts. 
NL injection in inner maximization at each minibatch equals randomly replacing a part of adversarial data with natural data.
Consequently, AT learning more natural data and less adversarial data leads to better generalization and worse robustness. 


\subsection{NL Injection in Outer Minimization}
\label{sec:AT_outer_NL}
\begin{figure}[tp!]
	\centering
	\vspace{-0mm}
	\subfigure [CIFAR-10 dataset]{
		\includegraphics[width=0.25\textwidth]{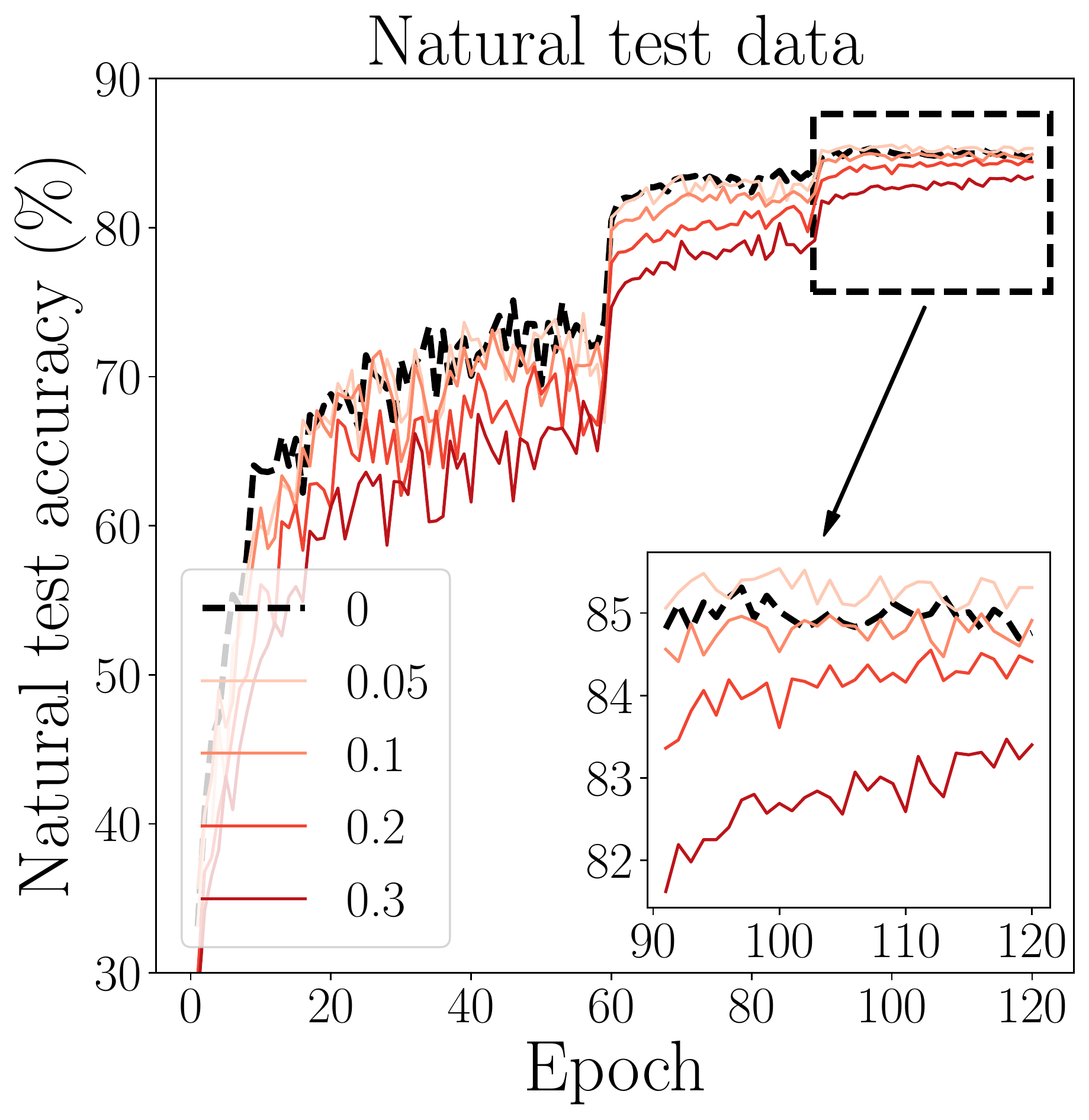}
	  \includegraphics[width=0.25\textwidth]{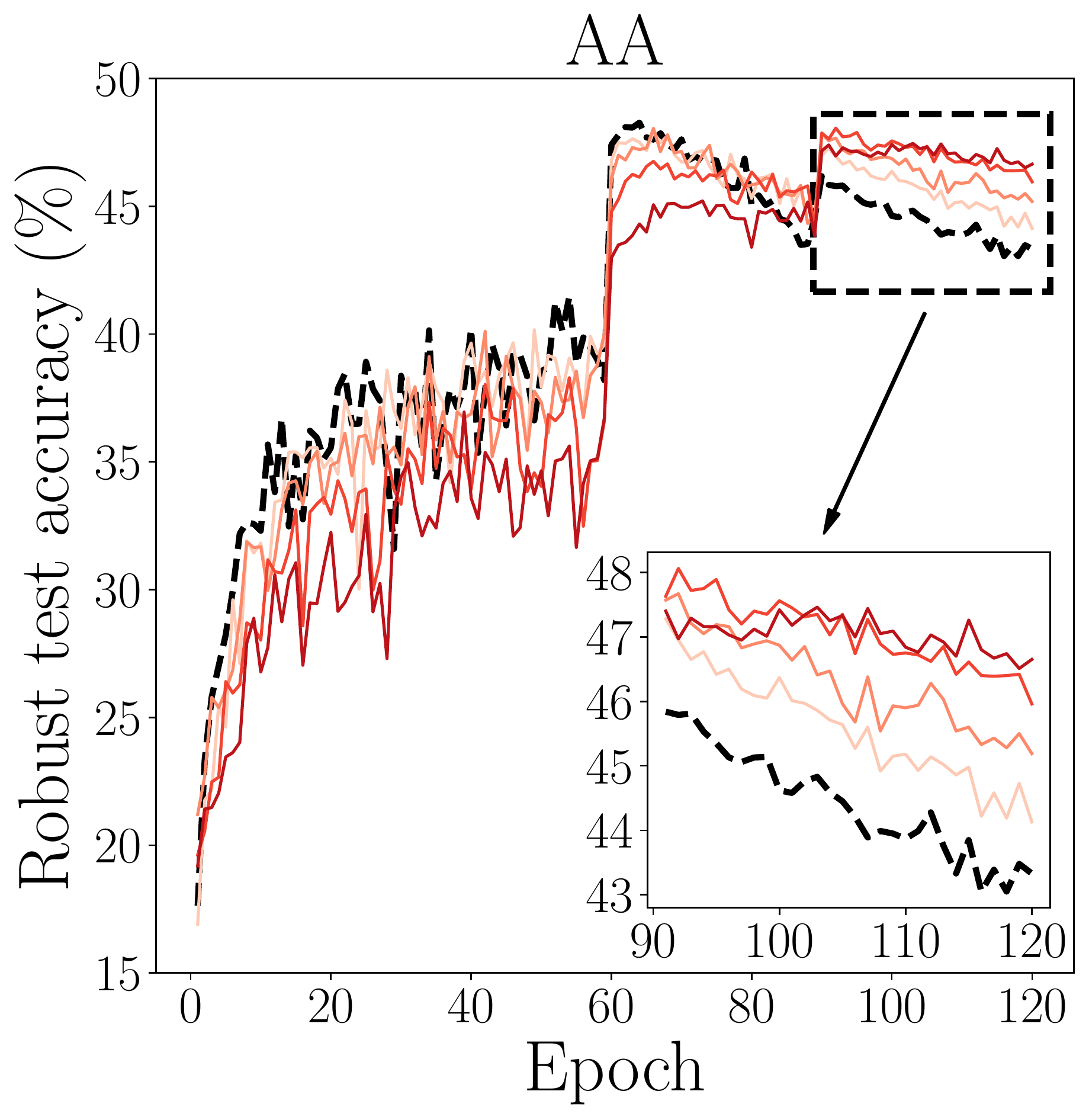}
		\label{fig:outer_NL}
	}
	\subfigure[Supporting schematics]{
		\includegraphics[width=0.39\textwidth]{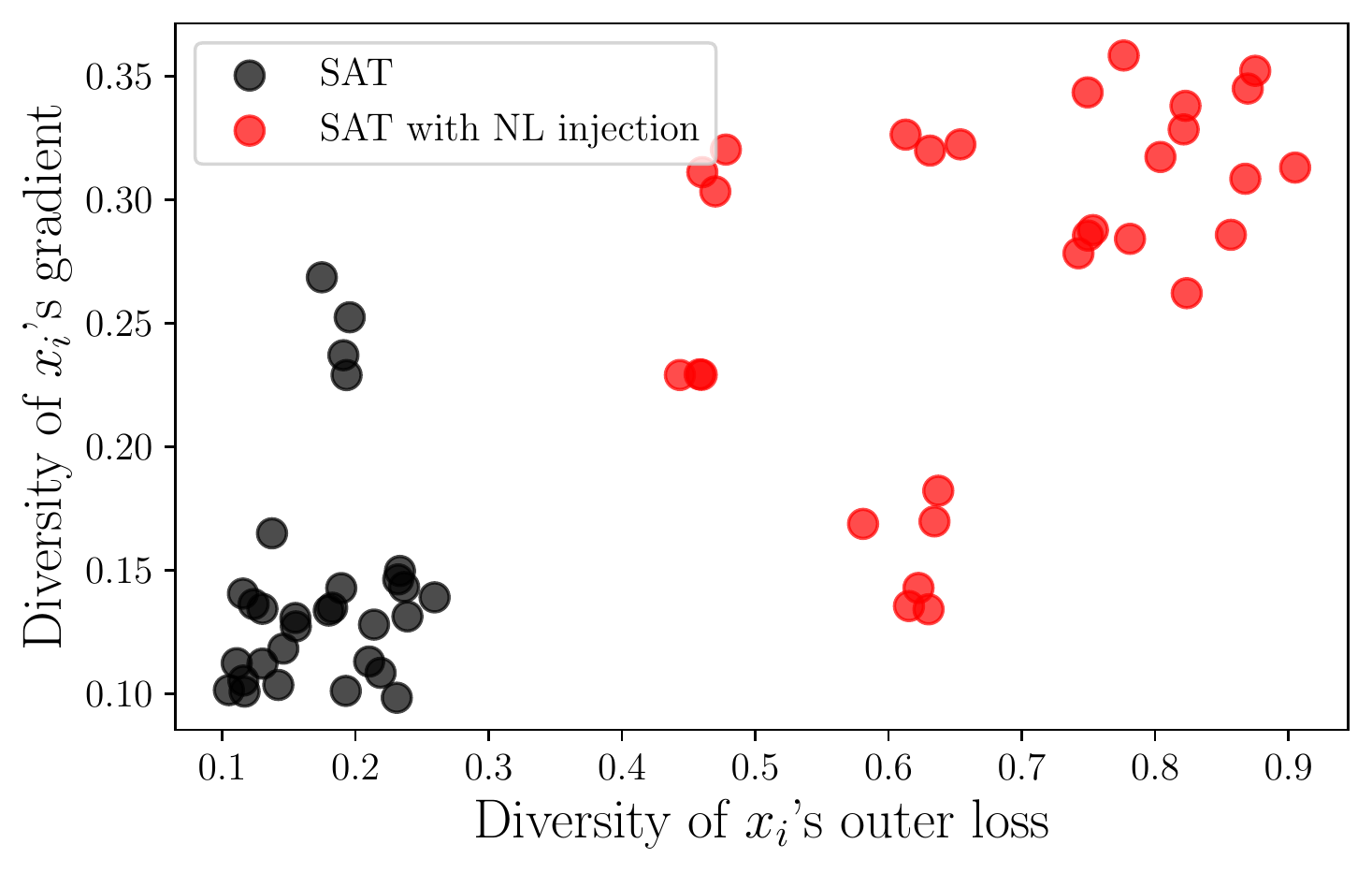}
		\label{data_diversity}
	}
	\vspace{-2mm}
	\caption{Figure~\ref{fig:outer_NL} shows the learning curves of injecting various levels of NL in outer minimization. (The number in the legend represents NL injection rate $\eta$. The black dash line represents SAT.) 
	Figure~\ref{data_diversity} shows data diversity comparisons between SAT (black dots) and SAT with NL injection (right dots). (Red dots are scattered at upper-right corner, which justifies that NL injection in outer minimization leads to high data diversity over the training epochs.) 
	NL injection in outer minimization serves as a strong regularization that prevents robust overfitting. 
	}
	\vspace{-2mm}
\end{figure}

\paragraph{Experiment details}
Figure~\ref{fig:outer_NL} shows the results of a series of experiments with various NL levels $\eta$ injected in outer minimization. 
Adversarial data are generated according to the intact labels. At every training minibatch, we randomly choose an $\eta$ portion of adversarial data to flip their labels for the learning. 
The other settings (e.g., the learning rate schedule, optimizer, network structures, dataset, $\epsilon_{\mathrm{train}}$ and $\epsilon_{\mathrm{test}}$, $K$, $\alpha$) are kept the same as Section~\ref{sec:NL_inner_experiment}.
Note that we do not show $\eta > 0.3$ because injecting too much noise (especially at the early training stage) significantly destabilizes and even kills the training.

In Figure~\ref{fig:outer_NL}, we show the learning curves of injecting symmetric-flipping NLs~\cite{van2015learning} . 
For completeness, in Appendix~\ref{appendix:outer_NL}, we show other attacks (such as the PGD and CW attacks) and also show pair-flipping NL~\cite{han2018co} injection.
In Appendix~\ref{appendix:outer_NL}, we find an intriguing phenomenon that pair-flipping NL injection obfuscates gradients of the CW attack. Specifically, pair-flipping NL injection significantly boosts the model's robustness against the CW attack.


\vspace{-0mm}
\paragraph{Observation ($ii$)} \textit{NL injection in outer minimization alleviates robust overfitting.} As shown in Figure~\ref{fig:outer_NL}, we find that at several last epochs, NL injection leads to consistently better robustness than SAT (black dashed line). Besides, the robustness at the final epoch improves with the increase of noise rate $\eta$. This shows that NL injection in outer minimization can significantly alleviate the undesirable issue of AT's robust overfitting~\cite{rice2020overfitting}. 

We interpret the Observation ($ii$) from data augmentation perspective. NL injection in outer minimization forces AT to learn the adversarial data $\xadv$ with the flipped label $\yadv$ occasionally. 
Considering an adversarial data point $(\xadv, y)$ and its flipped label $\yadv$, the objective loss function is $\ell(f(\xadv),\yadv)$. Taking the MSE loss as an example, we can rewrite the loss $\ell(f(\xadv),\yadv) = ||f(\xadv) - \yadv||_{2}^{2}$ as $|| (f(\xadv) - \yadv + y) - y ||_{2}^{2}$, in which we view $\tilde{f}(\xadv) := f(\xadv) - \yadv + y$ as noisy output.  
Through solving $\argmin_{\xadv_{\mathrm{aug}}}||\tilde{f}(\xadv) - f(\xadv_{\mathrm{aug}})||$, we basically project noisy output onto input space and augment original adversarial data $\xadv$ with $\xadv_{\mathrm{\mathrm{aug}}}$. 
Therefore, NL injection could increase the data diversity: at each training epoch, the model comes across different versions of data. 

We manifest the above perspective through conducting a proof-of-concept experiment 
on the CIFAR-10 dataset. We compare SAT and SAT with NL injection.
We randomly select 30 training data and collect the statistics of each data point at each epoch (120 epochs in total). 
Each data point $x_i$ corresponds to a 120-dimensional vector. 
$x_i$'s data diversity is reflected by the variance of its 120-dimensional vector.

Figure~\ref{data_diversity} illustrates the two statistics of each data point: 1) $x_i$'s \textit{outer loss} (horizontal axis)---the loss of the generated adversarial data, i.e., $\ell (f(\xadv_i),\yadv_i)$ and 2) $x_i$'s \textit{gradients} (vertical axis)---the $\ell_2$-norm of the weight gradient on $(\xadv_i,\yadv_i)$, i.e., $||\nabla_{\theta}\ell(f(\xadv_i),\yadv_i)||_2$.
For notational simplicity, over the training process, $\yadv_i$ is always equal to $y_i$ for SAT but not always equal to $y_i$ for SAT with NL injection (because $y_i$ is sometimes flipped to $\yadv$). 

Figure~\ref{data_diversity} shows that red rounds are scattered at the upper-right corner, and black squares are clustered at the lower-left corner, which justifies that SAT with NL injection has higher data diversity than SAT. 
{Figure~\ref{data_diversity} echoes the empirical observations in Figure~\ref{fig:outer_NL}, which implies that the high data diversity may impede the the undesirable issue of robust overfitting.}

\paragraph{NL Injection in both Inner Maximization and Outer Minimization} 
\paragraph{Observation (iii)} \textit{NL injection in data has a similar effect as Observation (ii).} 
In each training minibatch,  we randomly choose a portion of data to flip their labels. Then, we conduct SAT on them. We empirically find the performance is very similar to Observation (ii), with even a slightly better robustness. To avoid repetition and save space, we leave those results in Appendix~\ref{appendix:NN_fixed}. 
\paragraph{Observation (iv)}
There exists another case where NL injection in inner maximization mismatches that in outer minimization.  Specifically, at every minibatch, we choose a portion of data to flip labels in inner maximization and another portion of data to flip labels in outer minimization. 
In this setting, we observe slight degradation of robustness because label-flipped adversarial data are no longer serve the regularization purpose and are no longer adversarial as well to enhance robustness. The results can be found in Appendix~\ref{appendix:NN_pair}.

\section{Method}
The four observations in Section~\ref{sec:NL_AT} give us some insights on designing our own methods: Obs. (i) and Obs. (iv) actually harms AT's robustness, but robustness is usually AT's main purpose.  
Obs. (ii) and Obs. (iii) have the similar effects that effectively mitigate robust overfitting.
Inspired by the above Obs. (ii) or Obs. (iii), we propose our Algorithm~\ref{alg:AT_NN} (i.e., NoiLIn) that dynamically increases NL injection rate once the robust overfitting occurs. The simple strategy can be incorporated into various effective AT methods such as SAT~\citep{Madry_adversarial_training}, TRADES~\citep{Zhang_trades}, TRADES-AWP~\citep{wu2020adversarial}.
As a result, NoiLIn can fix robust overfitting issues of SAT and TRADES and further enhance generalization of TRADES-AWP\footnote{TRADES-AWP does not have robust overfitting issues because the AWP method has already fixed it. Besides, AWP has further enhanced TRADES's robustness, thus TRADES-AWP becomes the state-of-the-art AT method.} (see experiments parts in Section~\ref{sec:noilin-wrn}).

\vspace{-0mm}
\begin{algorithm}[t!]
	\caption{NoiLIn: automatically increasing \underline{Noi}sy \underline{L}abels \underline{In}jection rate in AT methods}
	\label{alg:AT_NN}
	\begin{algorithmic}
		\STATE {\bfseries Input:} network $f_{\mathbf{\theta}}$, training set $S_{\mathrm{train}}$, validation set $S_{\mathrm{valid}}$, total epochs $E$, initial noise rate $\eta_{\mathrm{min}}$, maximal noise rate $\eta_{\mathrm{max}}$, sliding window size $\tau$, boosting rate $\gamma$
		\STATE {\bfseries Output:} adversarially robust network $f_{\mathbf{\theta}}$  
		\STATE $\eta = \eta_{\mathrm{min}}$
		\FOR{Epoch  $e= 1$, $\dots$, $E$}
		\STATE Randomly flip $\eta$ portion of labels of training dataset $S_{\mathrm{train}}$ to get $\tilde{S}$
		\STATE Update $f_{\mathbf{\theta}}$ using $\tilde{S}$ by an AT method (such as SAT, TRADES, TRADES-AWP)
		\STATE Obtain robust validation accuracy $\mathcal{A}_{e}$ using $S_{\mathrm{valid}}$
		\IF{$\sum_{i=e-\tau}^{e}\mathcal{A}_i < \sum_{j=e-\tau-1}^{e-1}\mathcal{A}_j $ }
		\STATE$\eta = \min(\eta \cdot (1+\gamma), \eta_{\mathrm{max}})$ 
		{// Boost NL injection rate $\eta$ if robust overfitting occurs.} 
		\ENDIF
		\ENDFOR
	\end{algorithmic}
\end{algorithm} 
\vspace{-0mm}

As stated in Algorithm~\ref{alg:AT_NN}, at the beginning of each epoch, we randomly flip $\eta$ portion of labels of the training dataset $S_{\mathrm{train}}$ to get a noisy dataset $\tilde{S}$; then, we execute an existing AT method (e.g., SAT and TRADES) on the noisy dataset $\tilde{S}$. 
We monitor the training process using a clean validation set.
We increase noise injection rate $\eta$ by a small ratio $\gamma$ once the robust overfitting occurs.

%
\section{Experiment}
\label{sec:exp}
\begin{figure*}[t!]
	\centering
	\vspace{-0mm}
	\includegraphics[width=0.32\textwidth]{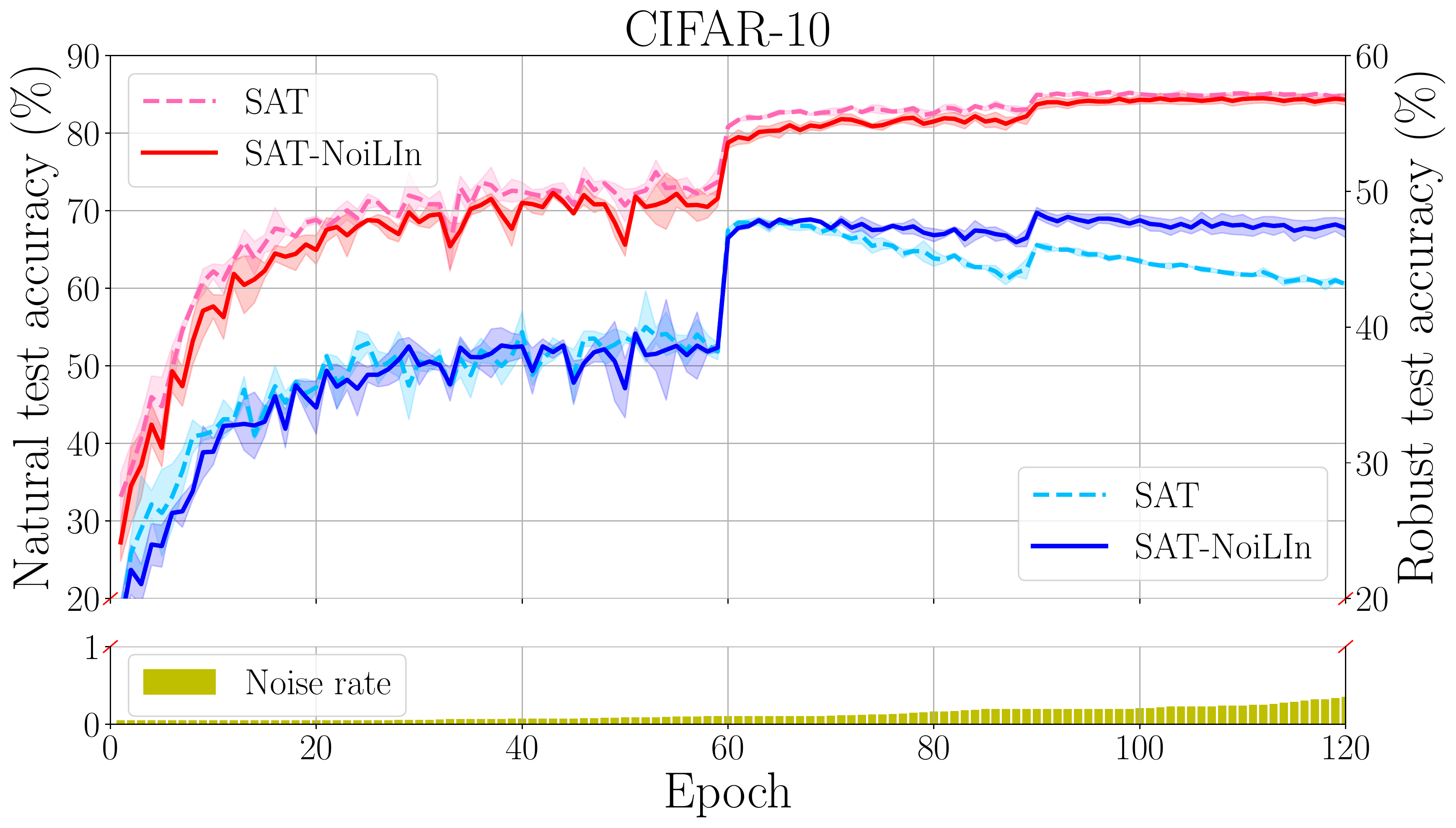}
	\includegraphics[width=0.32\textwidth]{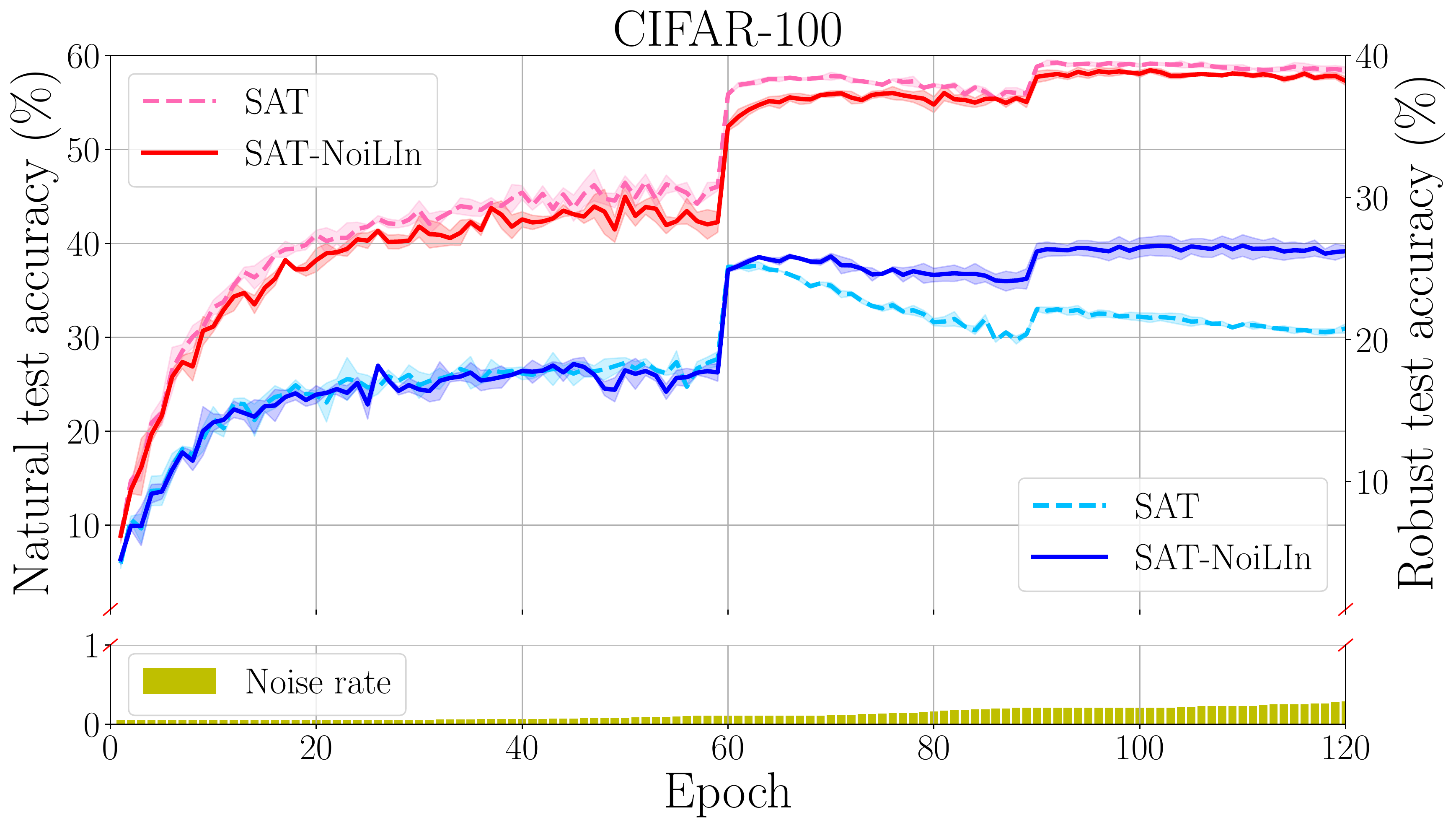}
	\includegraphics[width=0.32\textwidth]{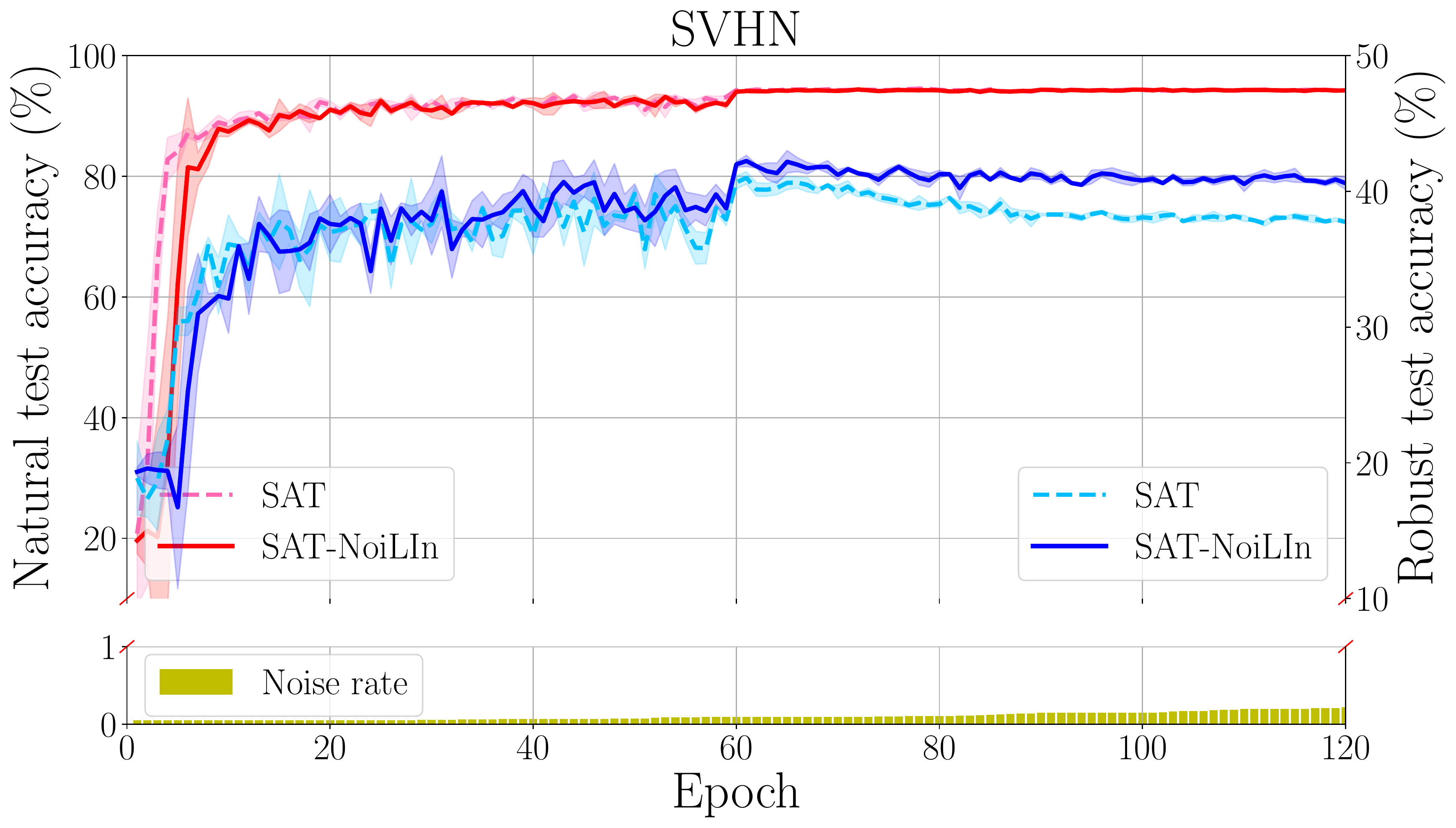}  \\
	\includegraphics[width=0.32\textwidth]{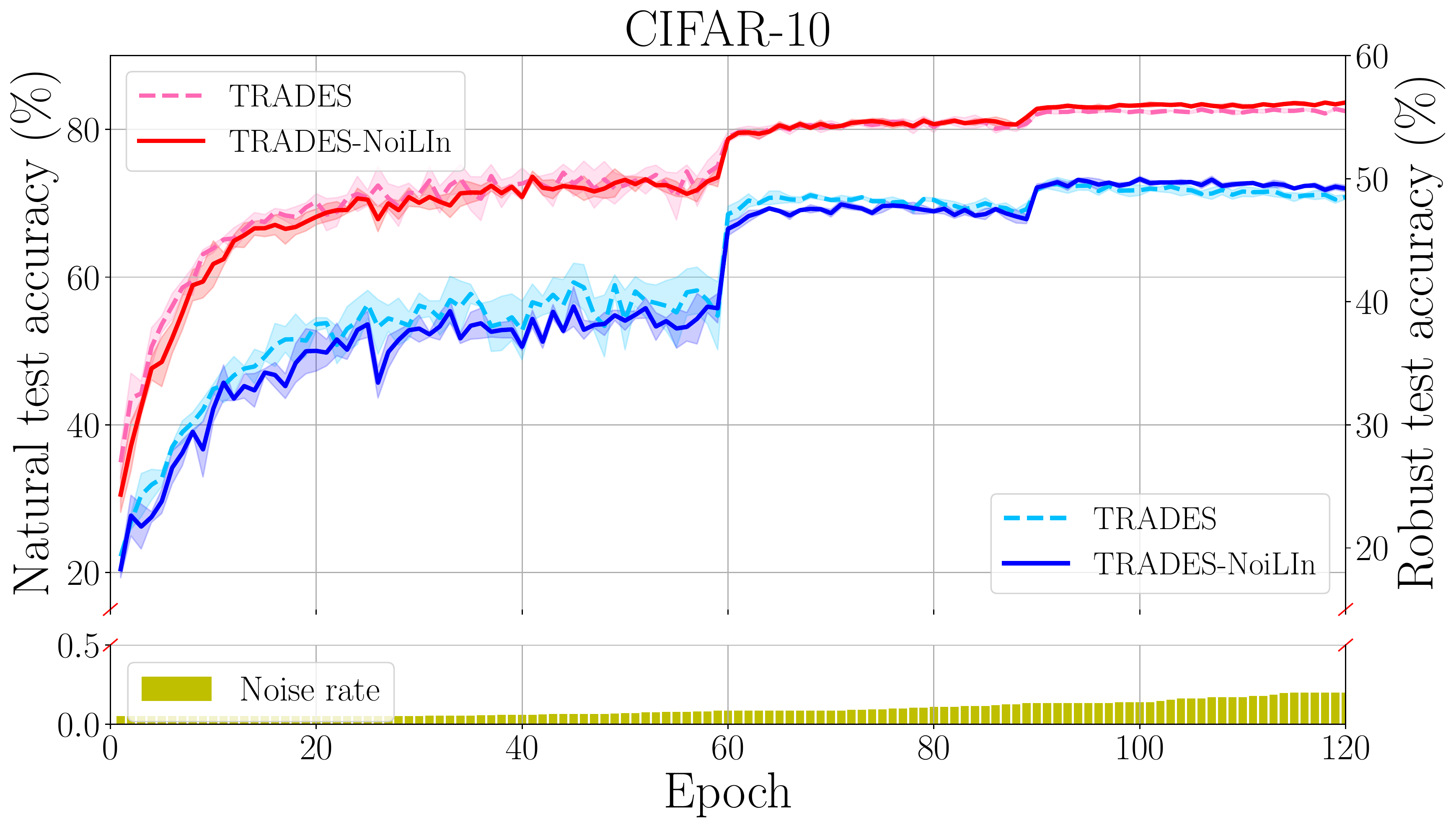}
	\includegraphics[width=0.32\textwidth]{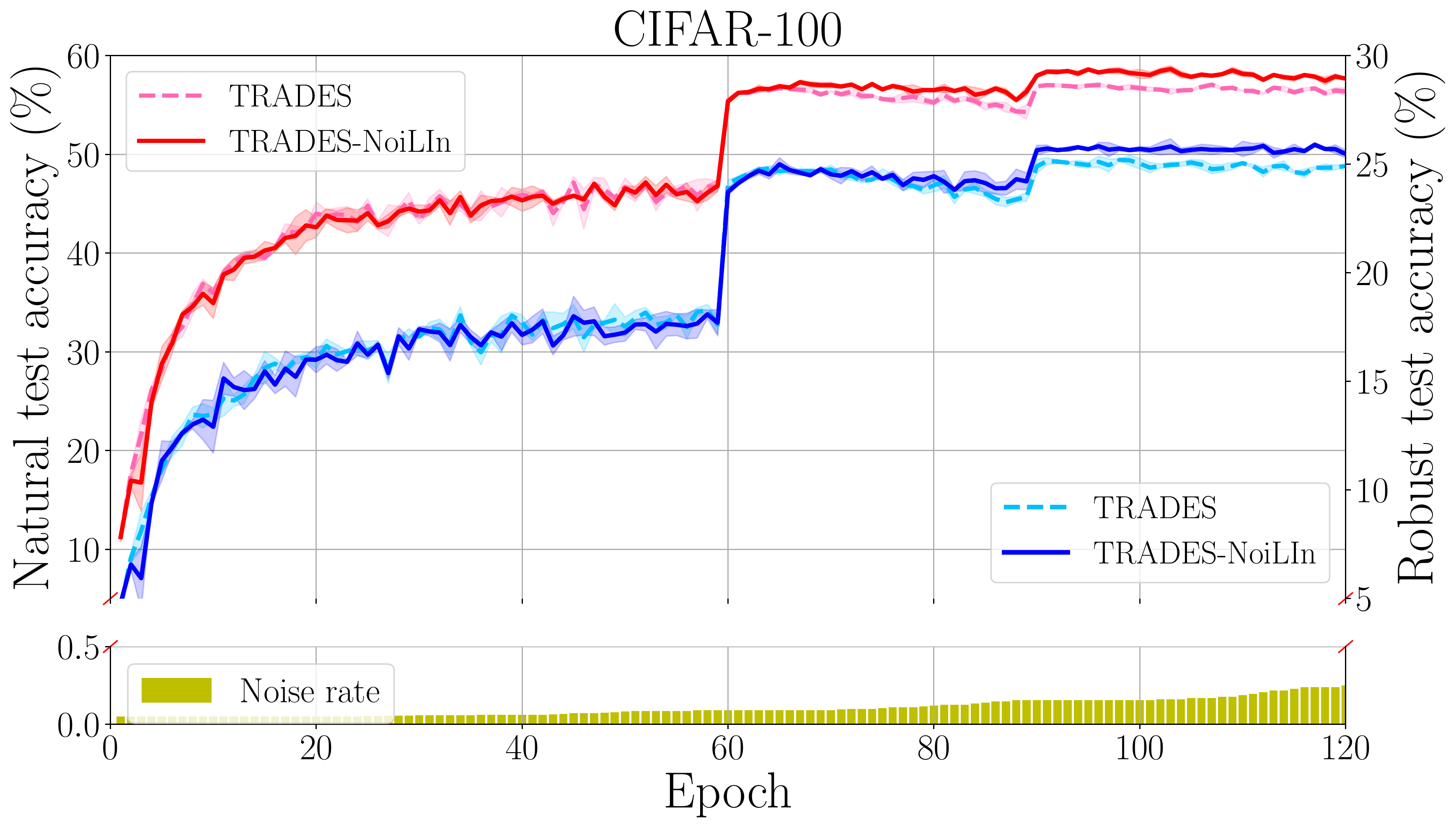}
	\includegraphics[width=0.32\textwidth]{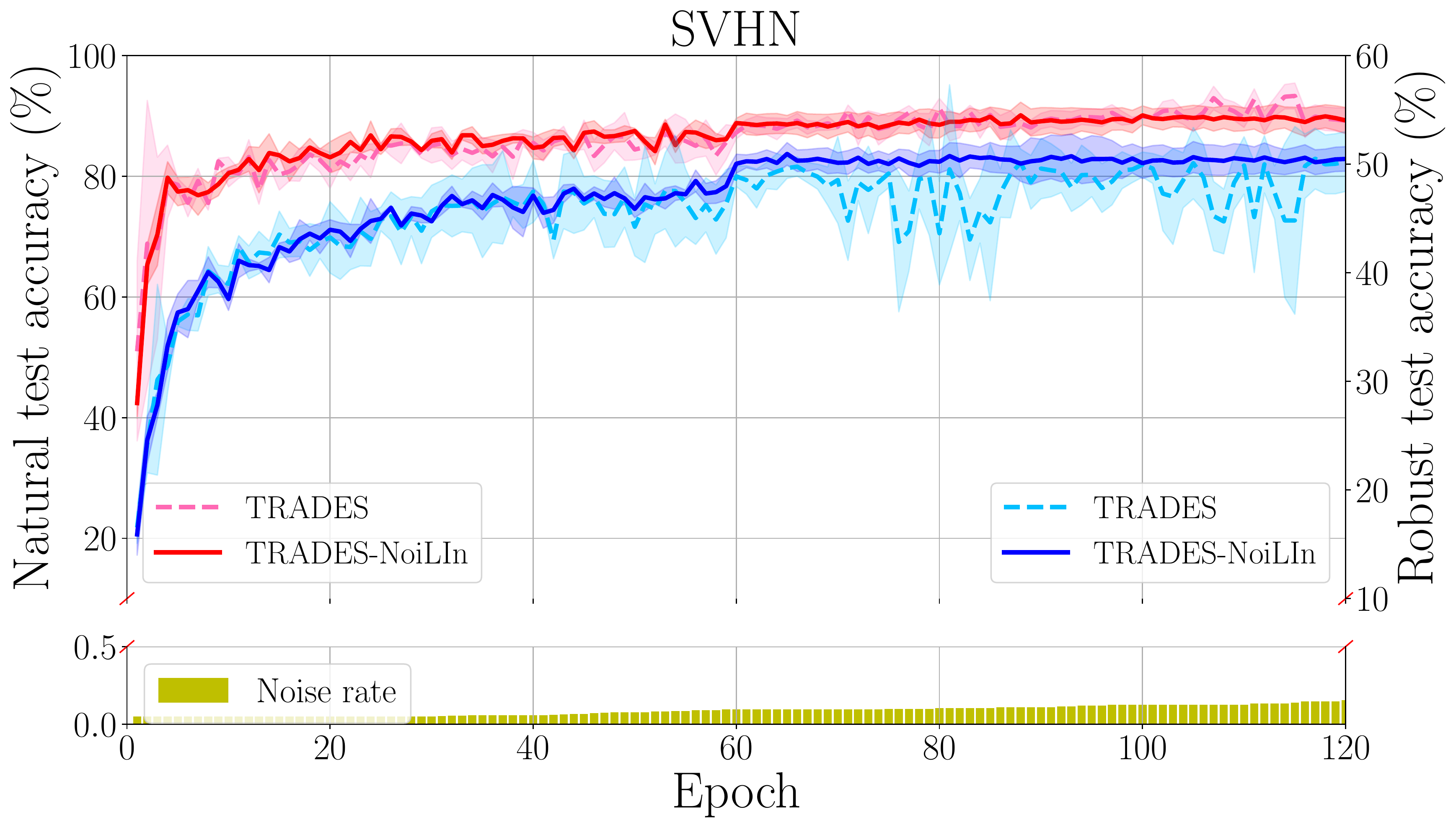}
	\vspace{-2mm}
	\caption{Evaluations on ResNet-18. We compare generalization (red lines) and robustness (blue lines, evaluated by the strongest AA attack) between our NoiLIn (solid lines) and two AT methods such as SAT and TRADES (dash lines) on three datasets---CIFAR-10, CIFAR-100, and SVHN. The yellow histogram below each figure reflects the change of NL injection rate over the training process. The shaded color reflects the standard deviations over the five repeated trails.  To maintain figures' neatness, we put the training accuracy curves in Appendix~\ref{appendix:NN_pair} (Figure~\ref{fig:robust-overfitting-with-trainingacc}).}
	\label{fig:robust-overfitting}
	\vspace{-2mm}
\end{figure*}
\subsection{NoiLin can Alleviate Robust Overfitting}
\label{sec:robust-overfitting}
Robust overfitting~\citep{rice2020overfitting} is an undesirable phenomenon in some AT methods particularly such as SAT and TRADES, where after the first learning rate decay, the model's robustness stops increasing but begins to decrease.

In Figure~\ref{fig:robust-overfitting}, we incorporate our NoiLIn using symmetric-flipping NLs into two typical AT methods, i.e., SAT and TRADES. We adversarially train ResNet-18 on three datasets, i.e., CIFAR-10, CIFAR-100 and SVHN.  
We took 1000 data from training set for validation and took the remaining data for the training.
The robust validation accuracy ($\mathcal{A}_{e}$ in Algorithm~\ref{alg:AT_NN}) was evaluated on PGD-10 attack for the consideration of speed.
For the hyperparameter setting of AT-NoiLIn, the noise rate $\eta$ was initialized as $\eta_{\mathrm{min}}=0.05$ with $\eta_{\mathrm{max}}=0.6$, $\tau=10$ and $\gamma=0.1$.
For the hyperparameter setting of TRADES-NoiLIn, we set $\eta_{\mathrm{min}}=0.05$, $\eta_{\mathrm{max}}=0.4$, $\tau=10$ and $\gamma=0.05$. We trained ResNet-18 using SGD with 0.9 momentum for 120 epochs with the initial learning rate of 0.1 (on CIFAR datasets and 0.01 on SVHN) and divided by 10 at Epochs 60 and 90, respectively. We compare the generalization evaluated on natural test data (red lines) and robustness evaluated on AA adversarial data~\citep{croce2020reliable} (blue lines) between typical AT methods (such as SAT and TRADES; dash lines) and NoiLIn counterparts (such as SAT-NoiLIn and TRADES-NoiLIn; solod lines). 
Besides, at the bottom of each panel, we also demonstrate the dynamic schedule of noise rate $\eta$ using the yellow histogram. 

Figure~\ref{fig:robust-overfitting} empirically validates the efficacy of NoiLIn to alleviate robust overfitting.
When the learning rate decays, the robust test accuracy of NoiLIn keeps steady and even rises slightly instead of immediately dropping. Besides, we observe that when the robust overfitting occurs, the noise rate (yellow pillars) gradually rises to deliver a stronger regularization to combat robust overfitting. 

\subsection{Performance Evaluations on Wide ResNet}
\label{sec:noilin-wrn}
To manifest the power of NoiLIn, we adversarially train Wide ResNet~\citep{zagoruyko2016WRN} on CIFAR-10 by incorporating our NoiLIn strategy using symmetric-flipping NLs to three common and effective AT methods, i.e., SAT, TRADES and TRADES-AWP (adversarial weight perturbation). 
In Table~\ref{tab:wrn_NN}, we used WRN-32-10 for AT-NoiLIn that keeps same as \cite{Madry_adversarial_training} and used WRN-34-10 for TRADES-NoiLIn and TRADES-AWP-NoiLIn that keeps same as \cite{Zhang_trades,wu2020adversarial}. 
The hyperparameters of AWP exactly follows~\cite{wu2020adversarial}. Other training settings (e.g., learning rate, optimizer, the hyperparameters for scheduling noise rate) keeps the same as Section~\ref{sec:robust-overfitting}. 
In Table~\ref{tab:wrn_NN}, we also compare independent and contemporary work, i.e.,``AT+ TE (temporal ensembing)''~\citep{dong2022exploring}.   

\begin{table}[t!]
	\centering
	\vspace{-0mm}
	\caption{Evaluations on Wide ResNet. 
		We reported the test accuracy of the best checkpoint and that of the last checkpoint as well as the gap between them---``best$\pm$std. /last$\pm$std. (gap)''.}
	\label{tab:wrn_NN}
	\resizebox{\columnwidth}{!} { 
		\begin{tabular}{c|ccc}
			\hline
			Defense & Natural & C$\&$W$_{\infty}$-100 & AA \\ \hline
			SAT~\citep{Madry_adversarial_training}  & \textbf{87.00\scriptsize{$\pm$0.556}}/\textbf{87.13\scriptsize{$\pm$0.167}} (\textbf{$+$0.13}) & {53.47\scriptsize{$\pm$0.174}}/{46.75\scriptsize{$\pm$0.272}} ($-6.72$) & {51.06\scriptsize{$\pm$0.471}}/{45.29\scriptsize{$\pm$0.247}} ($-5.77$) \\
			SAT+TE~\citep{dong2022exploring} & {85.84\scriptsize{$\pm$0.155}}/{86.25\scriptsize{$\pm$0.114}} ($+0.41$) & {55.91\scriptsize{$\pm$0.128}}/{53.23\scriptsize{$\pm$0.154}} ($-2.68$)  & {52.21\scriptsize{$\pm$0.262}}/{50.32\scriptsize{$\pm$0.243}} ($-1.89$) \\
			SAT-NoiLIn & {85.92\scriptsize{$\pm$0.400}}/{86.43\scriptsize{$\pm$0.343}} {($+0.51$)}   & \textbf{55.99\scriptsize{$\pm$0.352}}/\textbf{53.79\scriptsize{$\pm$0.579}} (\textbf{$-$2.20}) & \textbf{52.50\scriptsize{$\pm$0.278}}/\textbf{50.62\scriptsize{$\pm$0.555}} (\textbf{$-$1.88}) \\ \hline
			TRADES~\citep{Zhang_trades} & {84.92\scriptsize{$\pm$0.221}}/{85.16\scriptsize{$\pm$0.062}} ($+0.25$)& {53.69\scriptsize{$\pm$0.054}}/{49.89\scriptsize{$\pm$0.121}} ($-3.80$) & {52.54\scriptsize{$\pm$0.313}}/{47.95\scriptsize{$\pm$0.419}} ($-4.59$)\\
			TRADES+TE~\citep{dong2022exploring} & {85.25\scriptsize{$\pm$0.265}}/{85.78\scriptsize{$\pm$0.222}} ($+0.53$) & {53.30\scriptsize{$\pm$0.308}}/{51.42\scriptsize{$\pm$0.092}} ($-1.88$) & {52.56\scriptsize{$\pm$0.401}}/{50.09\scriptsize{$\pm$0.341}} ($-2.47$) \\
			TRADES-NoiLIn & \textbf{84.39\scriptsize{$\pm$0.142}}/\textbf{85.89\scriptsize{$\pm$0.076}} (\textbf{$+$1.50}) &  \textbf{54.37\scriptsize{$\pm$0.246}}/\textbf{51.69\scriptsize{$\pm$0.201}} (\textbf{$-$2.68}) & \textbf{53.14\scriptsize{$\pm$0.326}}/\textbf{50.16\scriptsize{$\pm$0.050}} (\textbf{$-$2.88})\\
			\hline
			TRADES-AWP~\citep{wu2020adversarial} & {84.48\scriptsize{$\pm$0.377}}/{84.96\scriptsize{$\pm$0.040}} {($+0.48$)} & {58.88\scriptsize{$\pm$0.087}}/{57.60\scriptsize{$\pm$0.194}} ($-1.28$) & {55.88\scriptsize{$\pm$0.182}}/{54.91\scriptsize{$\pm$0.207}} ($-0.97$) \\
			TRADES-AWP-NoiLIn & \textbf{86.69\scriptsize{$\pm$0.153}}/\textbf{87.13\scriptsize{$\pm$0.372}} (\textbf{$+$0.44}) & \textbf{59.98\scriptsize{$\pm$0.319}}/\textbf{59.54\scriptsize{$\pm$0.348}} (\textbf{$-$0.44}) & \textbf{56.12\scriptsize{$\pm$0.139}}/\textbf{55.89\scriptsize{$\pm$0.324}} (\textbf{$-$0.23}) \\
			\hline
		\end{tabular}
	}
	\vspace{-2mm}
\end{table}
On the current SOTA method---TRADES-AWP, our NoiLIn could further enhance its natural test accuracy by surprisingly $2\%$ while maintaining (even slightly improving) its robustness.  
Besides, on natural test accuracy, NoiLIn seems to have a 
bigger negative impact on SAT than TRADES. 
In terms of learning objectives, SAT has one cross-entropy loss on adversarial data; TRADES has two loss, i.e., cross-entropy loss on natural data $ +~6 \times$KL divergence loss between natural data and adversarial data, in which KL loss does not involve labels and only cross-entropy loss involves labels. Therefore, NoiLIn has less negative impact on TRADES than AT on natural accuracy.




\subsection{Ablation Study}
\label{sec:AT_NN_benefits}

In this section, we conduct ablation studies. We show NoiLIn is more effective in larger perturbation bounds $\epsilon$, is less sensitive to weight decay. Besides, we compare NoiLIn with label smoothing (LS), and conduct extensive ablation studies such as NoiLIn with various learning rate schedulers and various noisy-label training set. 

\paragraph{NoiLIn under larger $\epsilon$.}
NoiLIn is more effective for larger $\epsilon$. 
Robust overfitting becomes even worse when $\epsilon$ becomes larger. 
We compare SAT and SAT-NoiLIn under larger $\epsilon$ using ResNet-18 on CIFAR-10 dataset. The training settings of SAT and SAT-NoiLIn keep same as Section~\ref{sec:robust-overfitting} except $\epsilon$. We choose $\epsilon_{\mathrm{train}}$ from $\{$8/255,10/255,12/255,14/255,16/255$\}$. 
The step size $\alpha$ for PGD was $\epsilon/4$. 
Robust test accuracy was evaluated on adversarial data bounded by $L_{\infty}$ perturbations with $\epsilon_{\mathrm{test}} = \epsilon_{\mathrm{train}}$. 

\begin{figure}[t!]
	\centering
	\vspace{-0mm}
	\subfigure [Robust overfitting is more severe under larger $\epsilon=16/255$. ]{
		\includegraphics[width=0.235\textwidth]{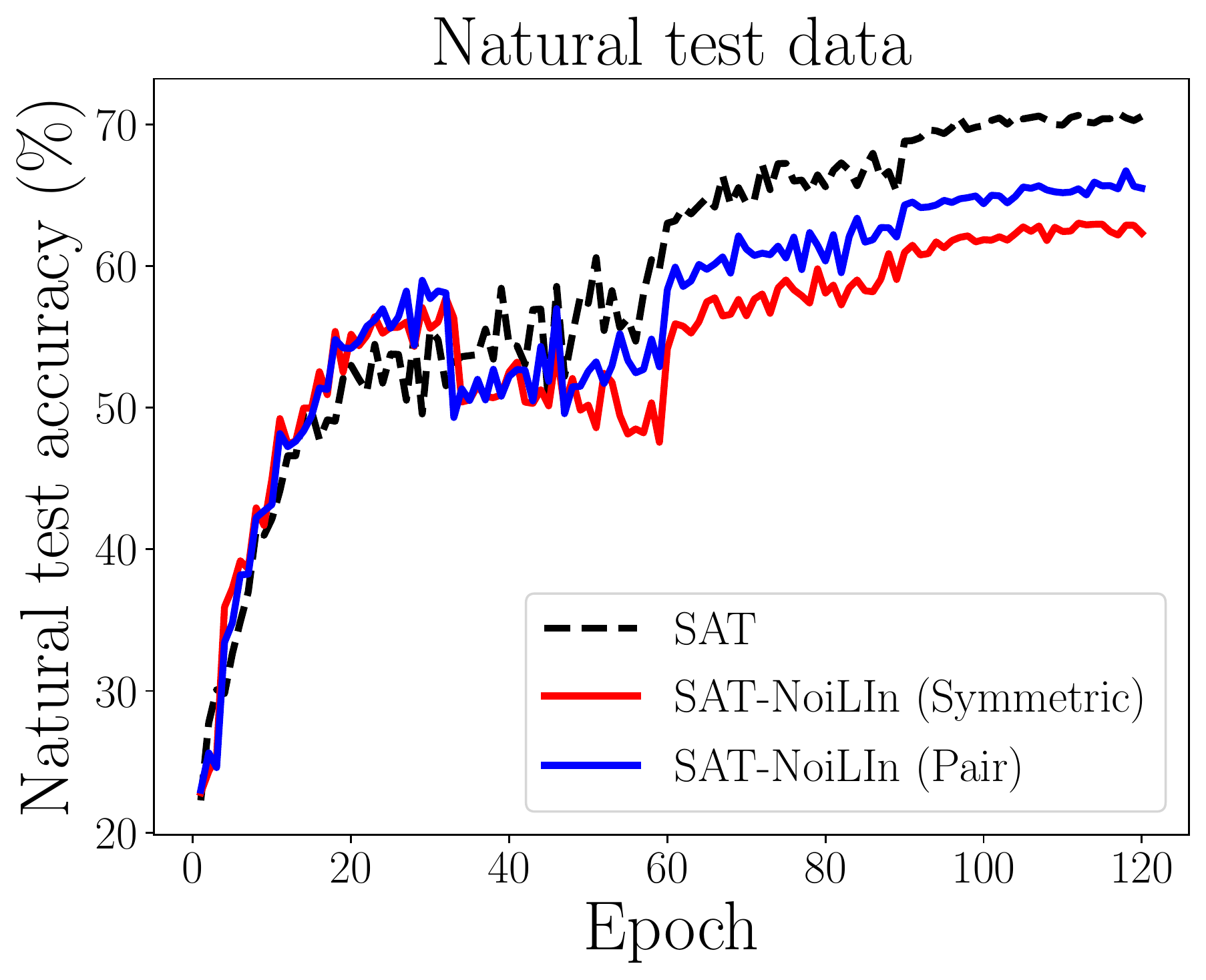}
	\includegraphics[width=0.235\textwidth]{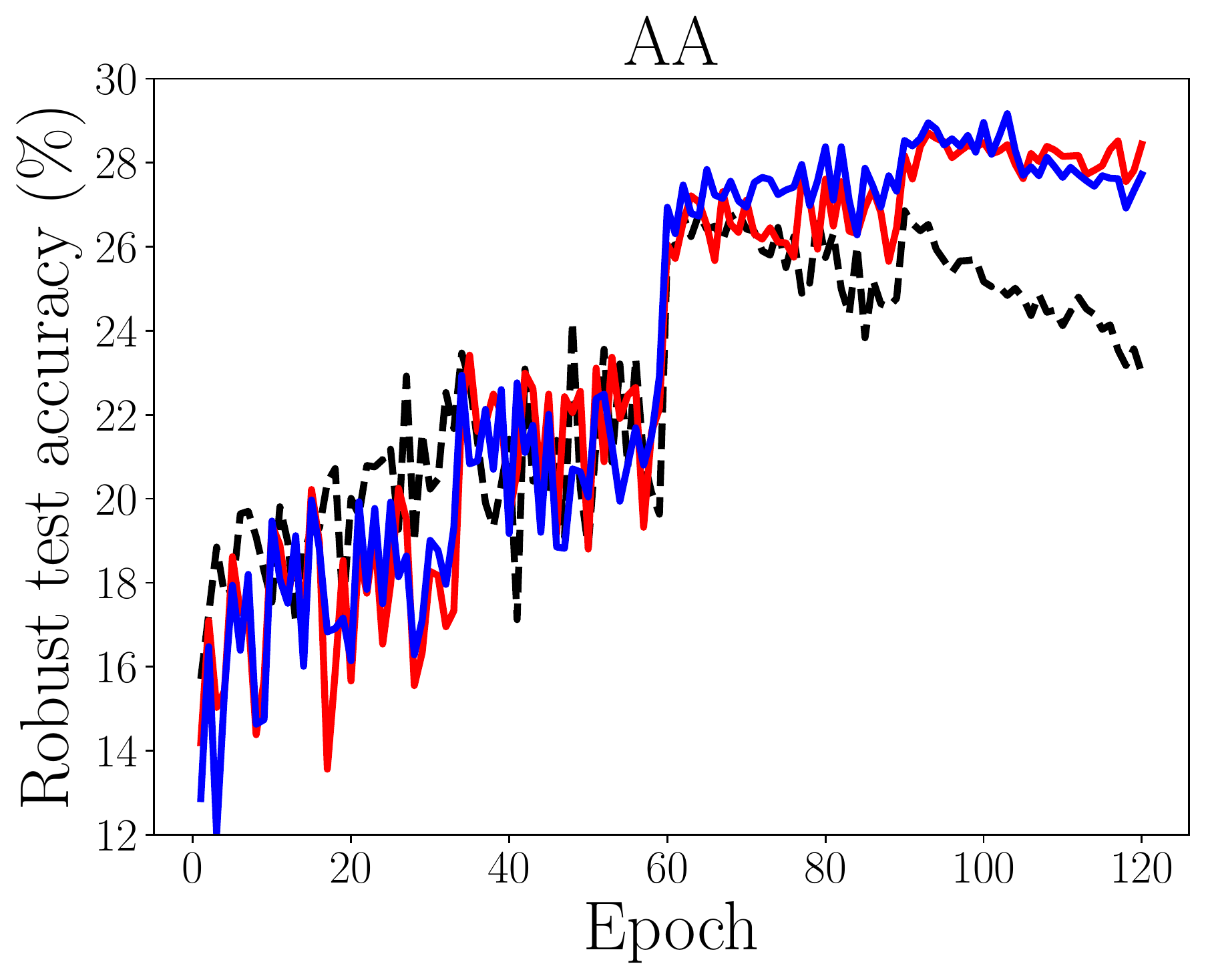}
		\label{fig:AT_NN_eps16}
	}
	\subfigure[NoiLIn enhances robustness further under larger $\epsilon$.]{
	\includegraphics[width=0.235\textwidth]{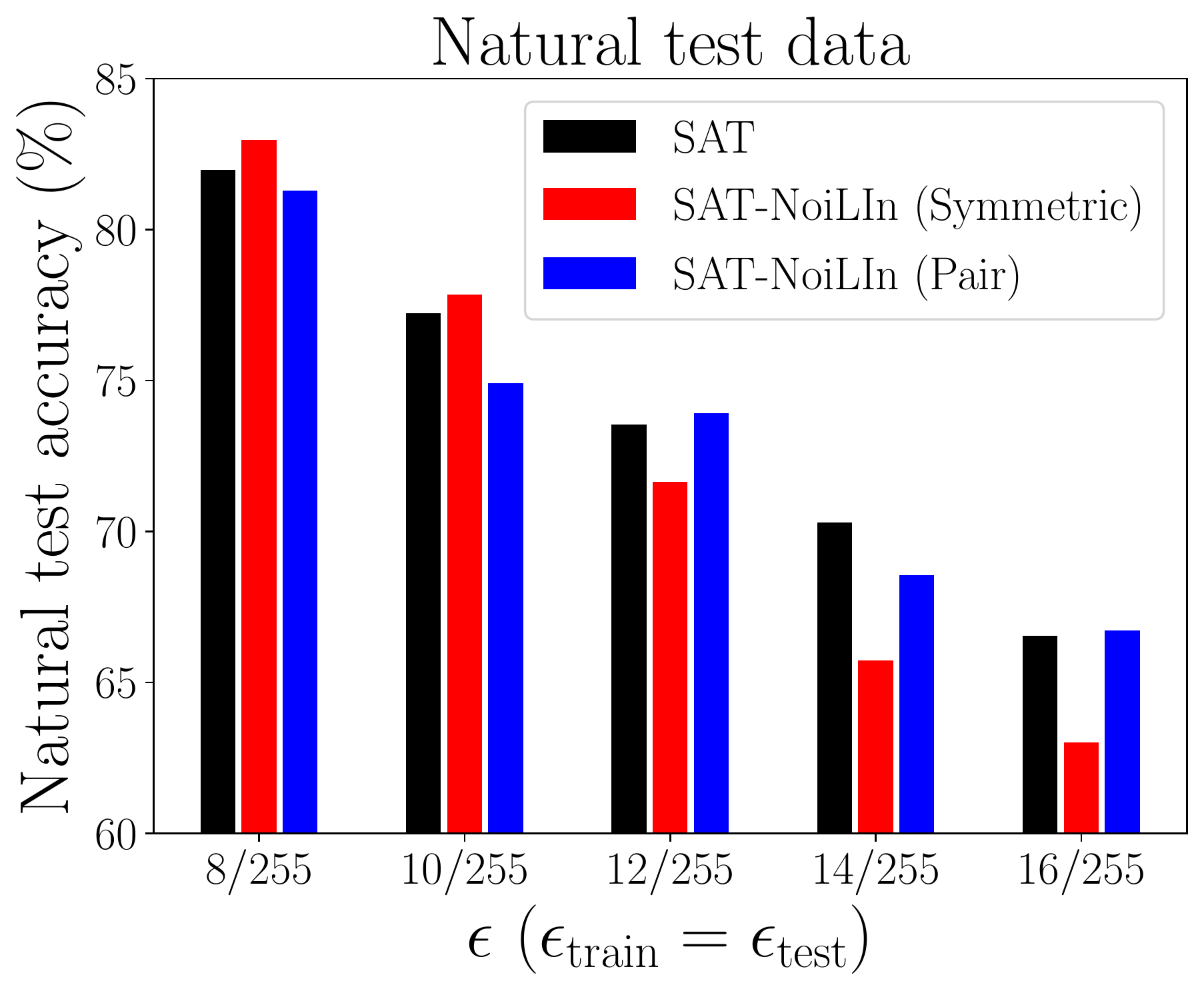}
	\includegraphics[width=0.235\textwidth]{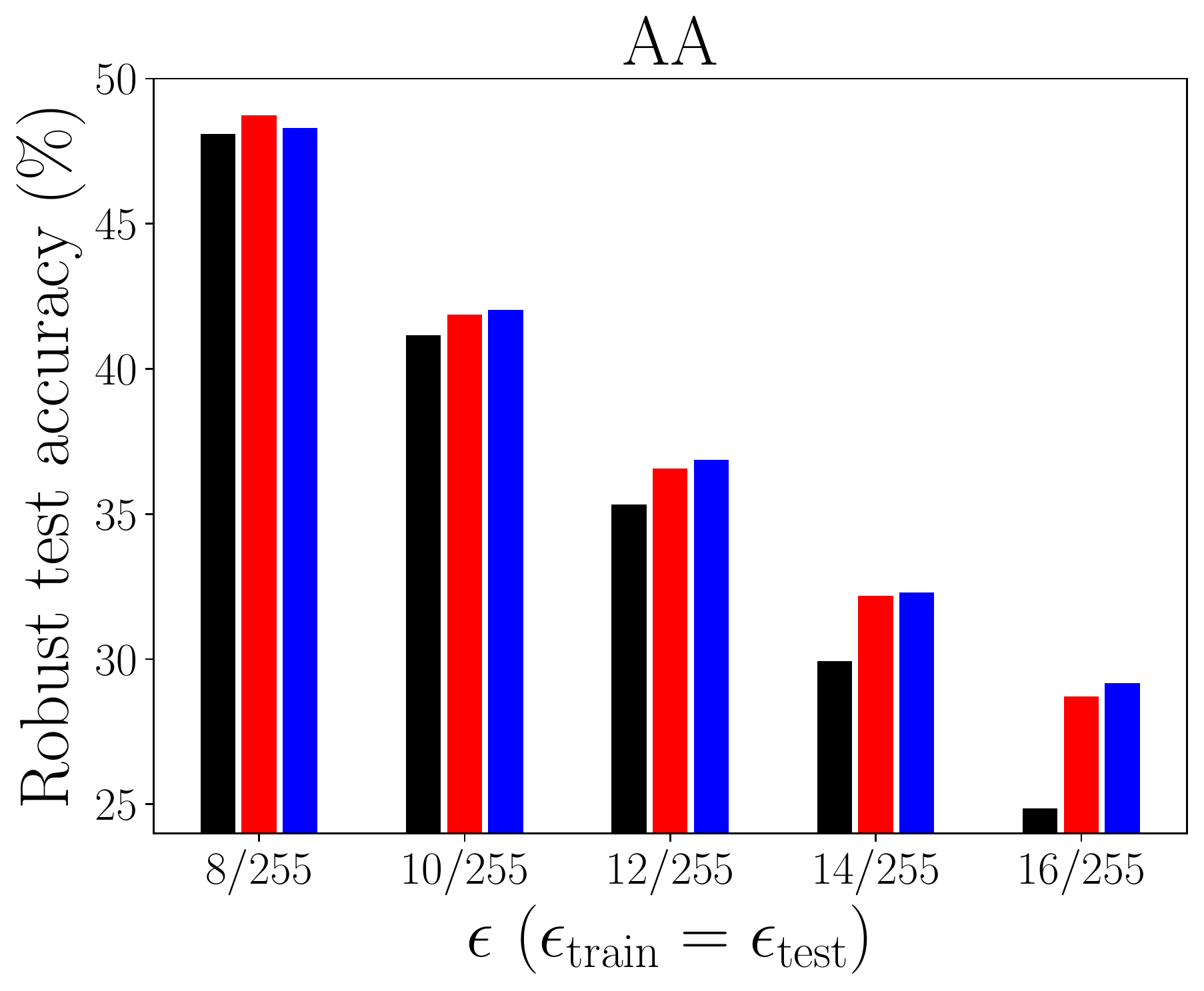}
		\label{fig:AT_NN_diff_eps}
	}
	\vspace{-2mm}
	\caption{Figure~\ref{fig:AT_NN_eps16} shows the learning curves of SAT-NoiLIn under $\epsilon_{\mathrm{train}} = 16/255$. (Robust test accuracy was evaluated on AA attacks with $\epsilon_{\mathrm{test}} = 16/255$). 
	Figure~\ref{fig:AT_NN_diff_eps} shows comparisons between SAT and SAT-NoiLIn under larger $\epsilon_{\mathrm{train}}$. (We reported the test accuracy evaluated at the best checkpoint according to the best robust accuracy. All figures keep $\epsilon_{\mathrm{train}} = \epsilon_{\mathrm{test}}$.) 
	We abbreviate SAT-NoiLIn using symmetric-flipping NLs as ``SAT-NoiLIn (Symmetric)'' and SAT-NoiLIn using pair-flipping NLs as ``SAT-NoiLIn (Pair)''.
	}
	\vspace{-2mm}
\end{figure}

Figure~\ref{fig:AT_NN_eps16} illustrates the learning curves of SAT and SAT-NoiLIn under $\epsilon_{\mathrm{train}} = \epsilon_{\mathrm{test}}=16/255$. Figure~\ref{fig:AT_NN_diff_eps} reports performance of the best checkpoints under different $\epsilon$. The best-checkpoints are chosen according to the best robust accuracy. 
Under larger $\epsilon$, SAT has more severe issue of robust overfitting and the robustness becomes lower. It is because AT fit data's neighborhoods, and the neighborhood becomes exponentially larger w.r.t. data's dimension even if  $\epsilon$ become slightly larger. This will cause larger overlap issues between classes, which requires larger label randomness for the learning.
Fortunately, under larger $\epsilon$, NoiLIn is more effective in relieving robust overfitting and even further improves adversarial robustness.



\vspace{-2mm}

\begin{figure}[h!]
	\vspace{-0mm}
	\centering
	\includegraphics[width=0.26\textwidth]{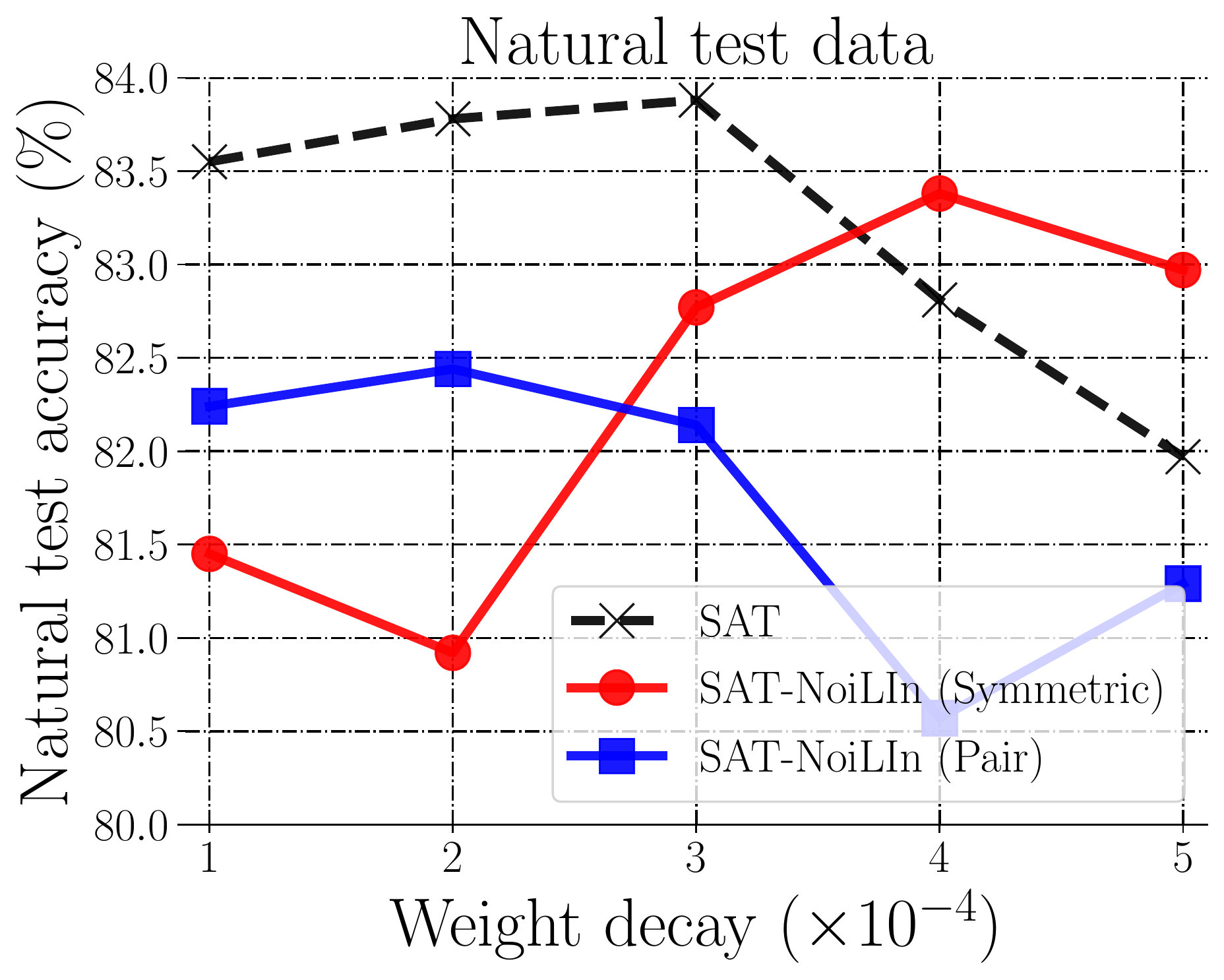}
	\includegraphics[width=0.26\textwidth]{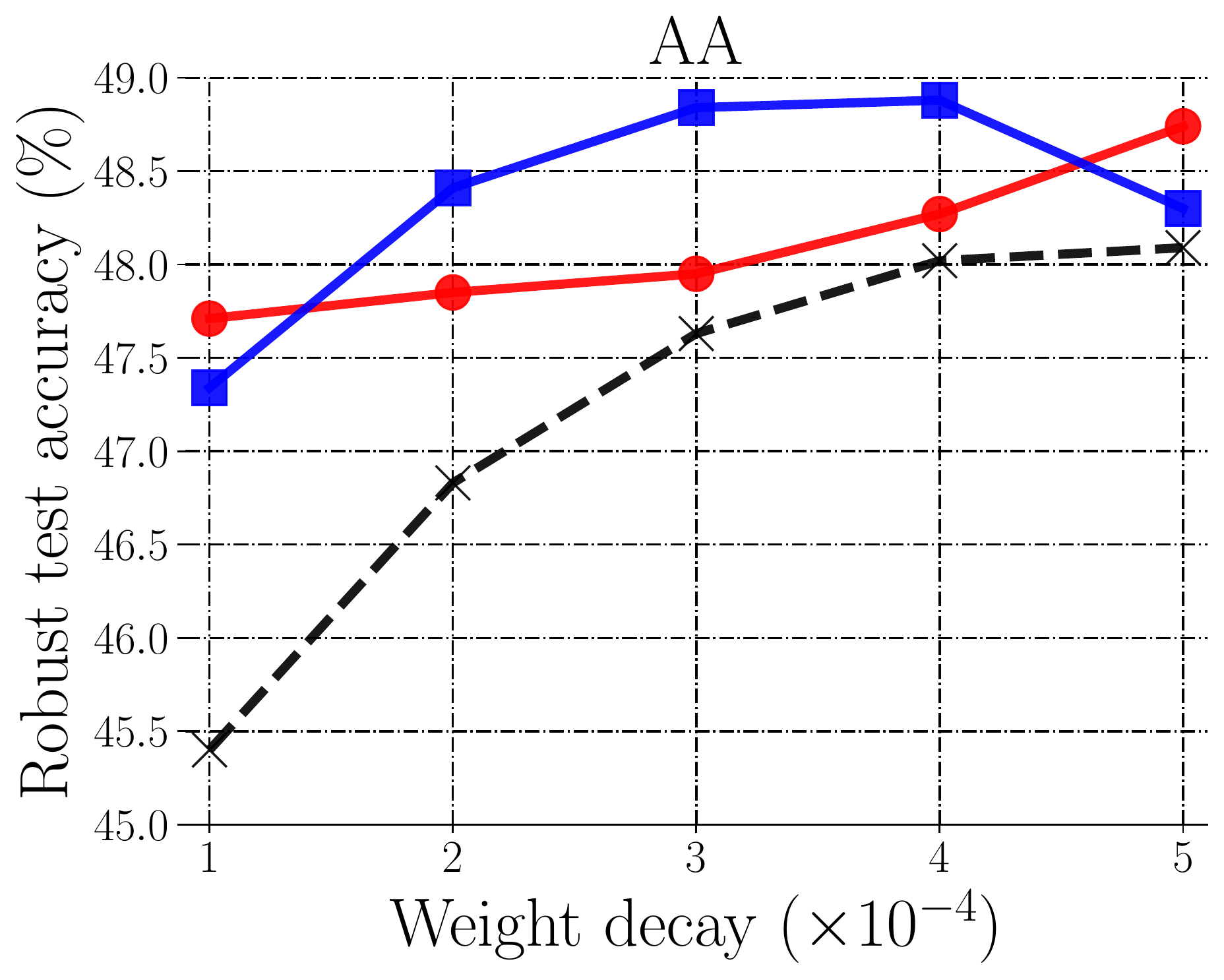}
	\vspace{-2mm}
	\caption{Test accuracy of robust models trained under different values of weight decay.
		}
	\label{fig:AT_NN_diff_wd}
	\vspace{-2mm}
\end{figure}

\paragraph{NoiLIn's sensitivity to weight decay.} 
\cite{pang2021bag} empirically found different values of weight decay can largely affect AT's adversarial robustness. 
In the ablation study, we explore the sensitivity of NoiLIn to weight decay. 
Figure~\ref{fig:AT_NN_diff_wd} shows NoiLIn can reduce AT methods' sensitivity on weight decay. 


In Figure~\ref{fig:AT_NN_diff_wd}, we conduct Algorithm~\ref{alg:AT_NN} under different values of weight decay and compared the SAT-NoiLIn with SAT. We use ResNet-18 and evaluate robustness at the best checkpoint. The settings keep the same as \cite{pang2021bag}.
From the right panel of Figure~\ref{fig:AT_NN_diff_wd}, both red and blue lines are higher and flatter than black lines across different values of weight decay. It signifies that noise injection can enhance adversarial robustness of SAT and meanwhile make SAT less sensitive to varying values of weight decay.

\vspace{-2mm}
\paragraph{NoiLIn's relation with label smoothing (LS).}
When taking the expectation w.r.t. all training epochs, NL injection is similar to employing the technique of label smoothing~\cite{szegedy2016rethinking,muller2019does} into AT (AT-LS)~\citep{cheng2020cat,pang2021bag}. Despite the mathematical similarity, we empirically found that NoiLIn can relieve robust overfitting while LS cannot. 



In Table~\ref{tab:r18_at_LS}, we reported robustness evaluations of AT with LS and AT-NoiLIn using ResNet-18 on CIFAR-10 dataset and set the training settings same as Section~\ref{sec:robust-overfitting}. 
AT-LS uses the $\rho$ level of smoothed label $\Bar{y}$ in outer minimization while using the one-hot label $y$ in inner maximization.
The $\rho$ level of smoothed label $\Bar{y}$ is a $C$-dimensional vector with the $y$-th dimension being $1 - \rho$ and all others being $\frac{\rho}{C-1}$. 
We implement LS with $\rho=0.1$ only in outer minimization (``AT-LS-outer'') and LS with $\rho=0.1$ in both inner maximization and outer minimization (``AT-LS-both''). Further, we even conduct LS in both inner maximization and outer minimization with dynamic LS level $\rho$ (``AT-LS-dynamic''). The schedule of $\rho$ is set exactly same as that of $\eta$ in AT-NoiLIn. We report the test accuracy at the best/last epoch and the gap between them in Table~\ref{tab:r18_at_LS}.
Extensive comparisons between AT-LS-outer under other different $\rho$ and AT-NoiLIn are reported in Appendix~\ref{appendix:LS_diff_level}.

\begin{table}[t!]
	\centering
	\vspace{-0mm}
	\caption{Comparisons between SAT, AT-LS-outer ($\rho=0.1$), AT-LS-both ($\rho=0.1$), AT-LS-adaptive and SAT-NoiLIn (Symmetric). 
		We report the test accuracy of the best checkpoint and that of the last checkpoint as well as the gap between them---``best/last (gap)''.}
	\vspace{-0mm}
	\label{tab:r18_at_LS}
	\resizebox{\columnwidth}{!} { 
		\begin{tabular}{c|ccc}
			\hline
			Defense               & Natural      & C$\&$W$_{\infty}$-100 & AA \\ \hline
			SAT~\cite{Madry_adversarial_training}  & 81.97/84.76 {($+2.79$)}  & 49.53/45.12 ($-4.71$) & 48.09/43.30 ($-4.79$)\\
			AT-LS-outer~\cite{pang2021bag}   & 82.76/85.15 ($+2.39$)  & 50.06/45.40 ($-4.66$) & 48.60/43.44 ($-5.16$) \\
			AT-LS-both   & 82.28/84.34 ($+2.06$)  & 49.79/46.17 ($-3.62$) & 48.44/44.59 ($-3.85$) \\
			AT-LS-adaptive   & \textbf{82.68/85.20 ($+$2.52)}  & 50.03/46.72 ($-3.31$) & 48.68/45.23 ($-3.45$) \\
			SAT-NoiLIn  & {82.97}/83.86 ($+0.89$)  & \textbf{51.21}/\textbf{50.36} \textbf{($-$0.85)} &  \textbf{48.74}/\textbf{47.93} \textbf{($-$0.81)} \\
			\hline
		\end{tabular}
	}
	\vspace{-2mm}
\end{table}

We found none of LS strategies can effectively relieve the robust overfitting. 
Table~\ref{tab:r18_at_LS} shows that compared with variants of AT-LS, AT-NoiLIn can largely relieve robust overfitting, while LS fails to make any effect. 
The reasons may be that AT-LS uses fixed smoothed labels, which is similar to AT using fixed one-hot labels, thereby having lower data diversity during the training process; by contrast, NoiLIn randomly flips labels at each epoch, leading to higher data diversity and thus effectively preventing robust overfitting (see Section~\ref{sec:AT_outer_NL}). 

In addition, compared with LS, NoiLIn saves $(C-1)$ log operations for each data. For each adversarial training data $\xadv$ whose prediction is $p(\xadv)$, the calculation of the cross-entropy loss of AT-LS is $- \sum_{j=1}^{C} \Bar{y}_j\log p_j(\xadv)$ while that of SAT-NoiLIn is $-\log p_{\yadv}(\xadv)$.



\vspace{-2mm}
\paragraph{NoiLIn under different learning rate (LR) schedulers.}
\cite{rice2020overfitting} showed robust overfitting ubiquitously occurs in AT under different LR schedulers. We conducted SAT-NoiLIn under different LR schedulers using ResNet-18 on CIFAR-10 dataset and kept all the training settings, such as the optimizer and hyperparameters of noise rate, exactly same as Section~\ref{sec:robust-overfitting} except the LR scheduler.
\begin{table}[h!]
	\centering
	\vspace{-2mm}
	\caption{Different learning rate schedule}
	\label{tab:r18_at_diff_lr}
	\vspace{0mm}
	\resizebox{\columnwidth}{!} { 
		\begin{tabular}{c|c|ccc}
			\hline
			Defense & LR scheduler & Natural      & C$\&$W$_{\infty}$-100 & AA \\ \hline
			SAT~\cite{Madry_adversarial_training}    & piecewise  & 81.97/84.76 ($+2.79$)  & 49.53/45.12 ($-4.71$) & 48.09/43.30 ($-4.79$)\\ \hline
			\multirow{4}{*}{\begin{tabular}[c]{@{}c@{}} SAT-NoiLIn \\ (Symmetric)\end{tabular}} & piecewise & 82.97/83.86 ($+0.89$) & 51.21/50.36 ($-0.85$) & 48.74/47.93 ($-0.81$)\\
			& multiple decay & 81.74/83.35 {($+1.61$)} & 49.42/48.59 ($-0.83$) & 47.50/46.22 ($-1.28$)\\
			& cosine & \textbf{84.86/85.02 ($+$0.16)}& 50.83/50.59 ($-0.24$) & 48.24/48.08 ($-0.16$)\\
			& cyclic & 83.48/83.48 ($+0.00$) & \textbf{51.41}/\textbf{51.41} \textbf{($+$0.00)} & \textbf{49.05}/\textbf{49.05} \textbf{($+$0.00)} \\ 
			\hline
		\end{tabular}
	}
	\vspace{0mm}
\end{table}
We report the test accuracy of the best checkpoint and that of the last checkpoint as well as the gap between them in Table~\ref{tab:r18_at_diff_lr} and show the learning rate schduler in Figure~\ref{fig:AT_NN_lr_schedule}. We observe the gap between test accuracy of the best checkpoint and that of the last checkpoint largely narrows with automatic NL injection, which clearly indicates NoiLIn can mitigate robust overfitting under all different LR schedulers. Further, we demonstrate the learning curves of SAT-NoiLIn under various LR schedulers in Figure~\ref{fig:AT_NN_lr_schedule} (Appendix~\ref{appendix:noilin_diff_lr}), which again validates NL injection can relieve overfitting.

\begin{figure}[t!]
	\centering
		\includegraphics[width=0.235\textwidth]{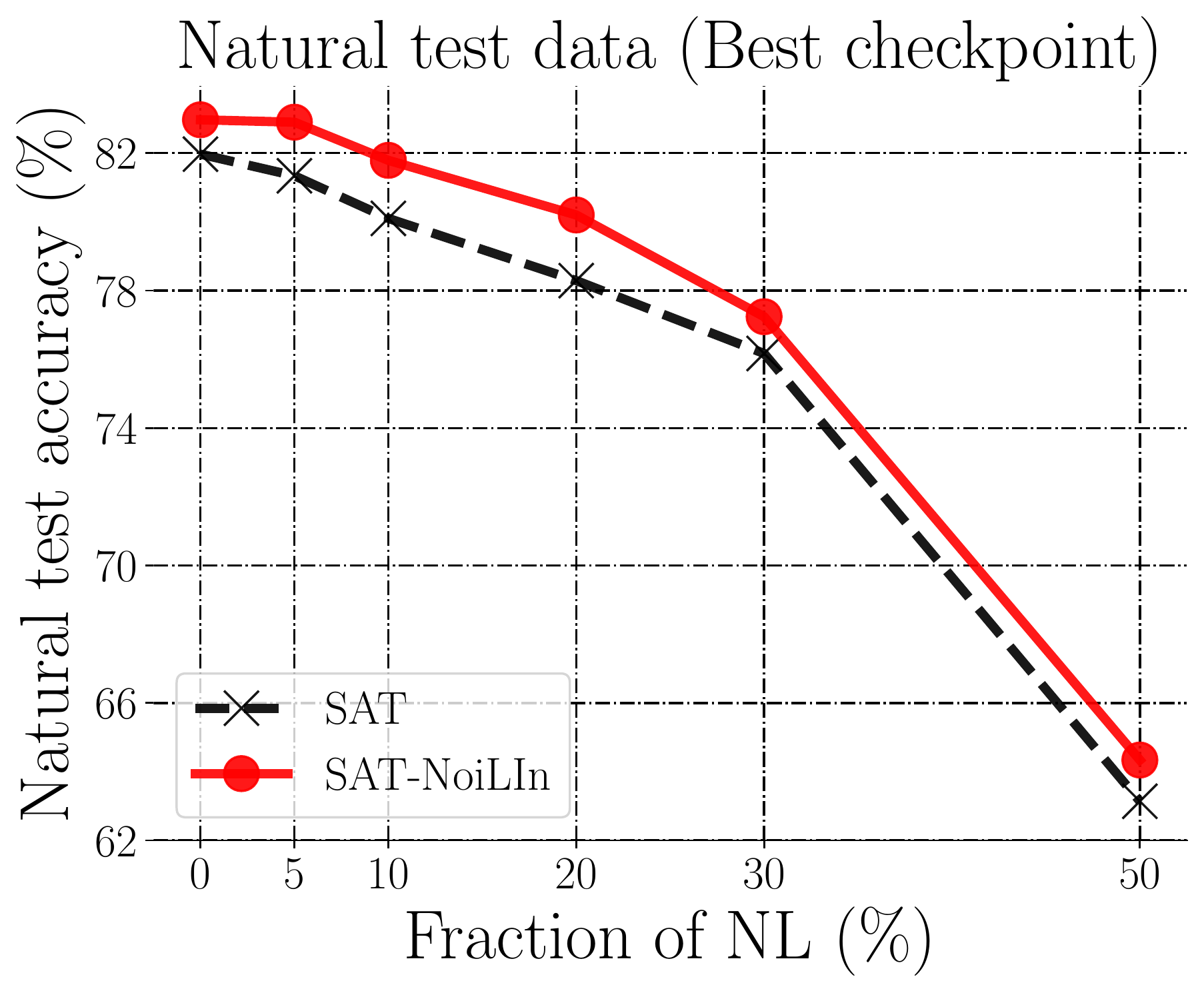}
		\includegraphics[width=0.235\textwidth]{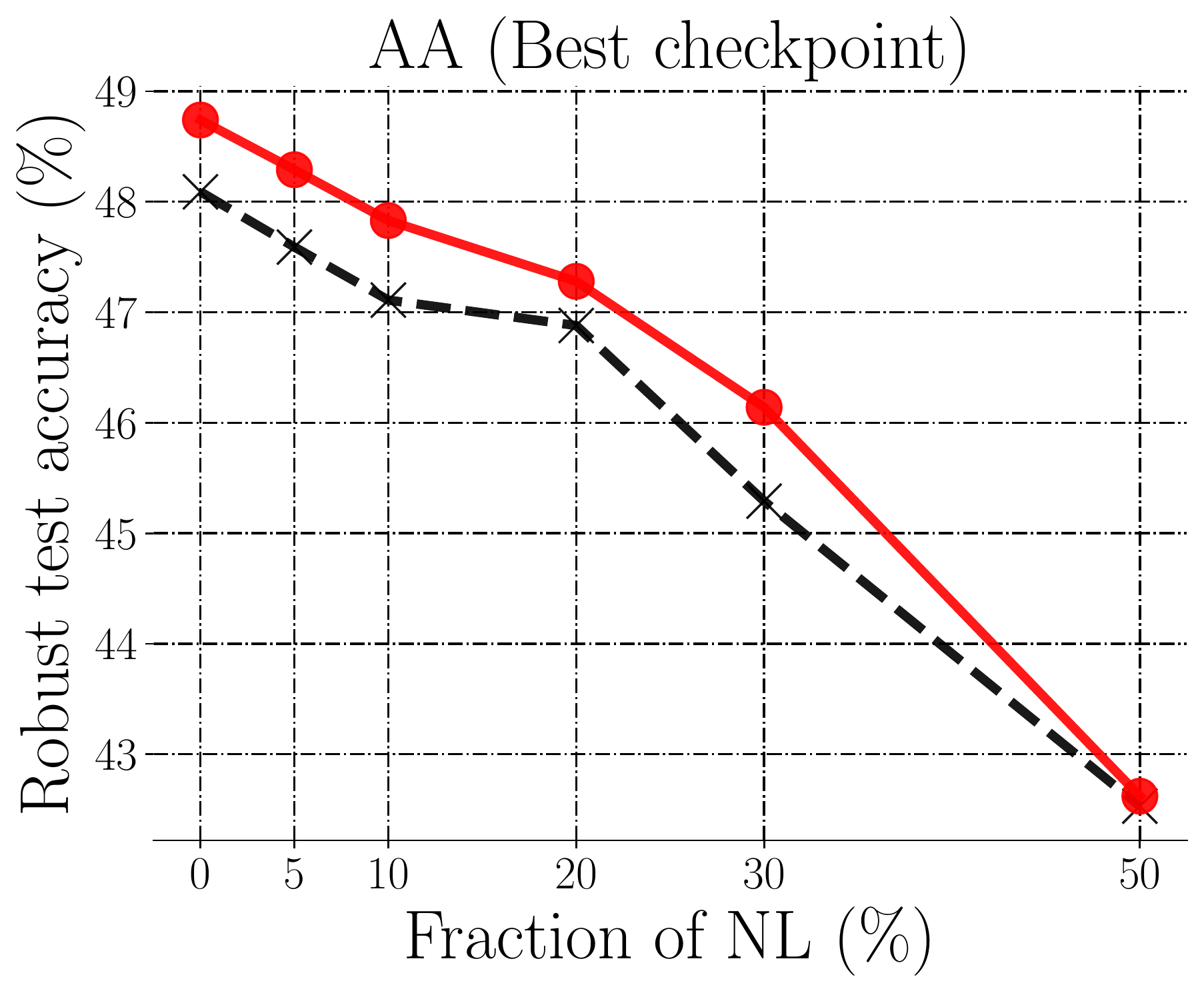}
		\includegraphics[width=0.235\textwidth]{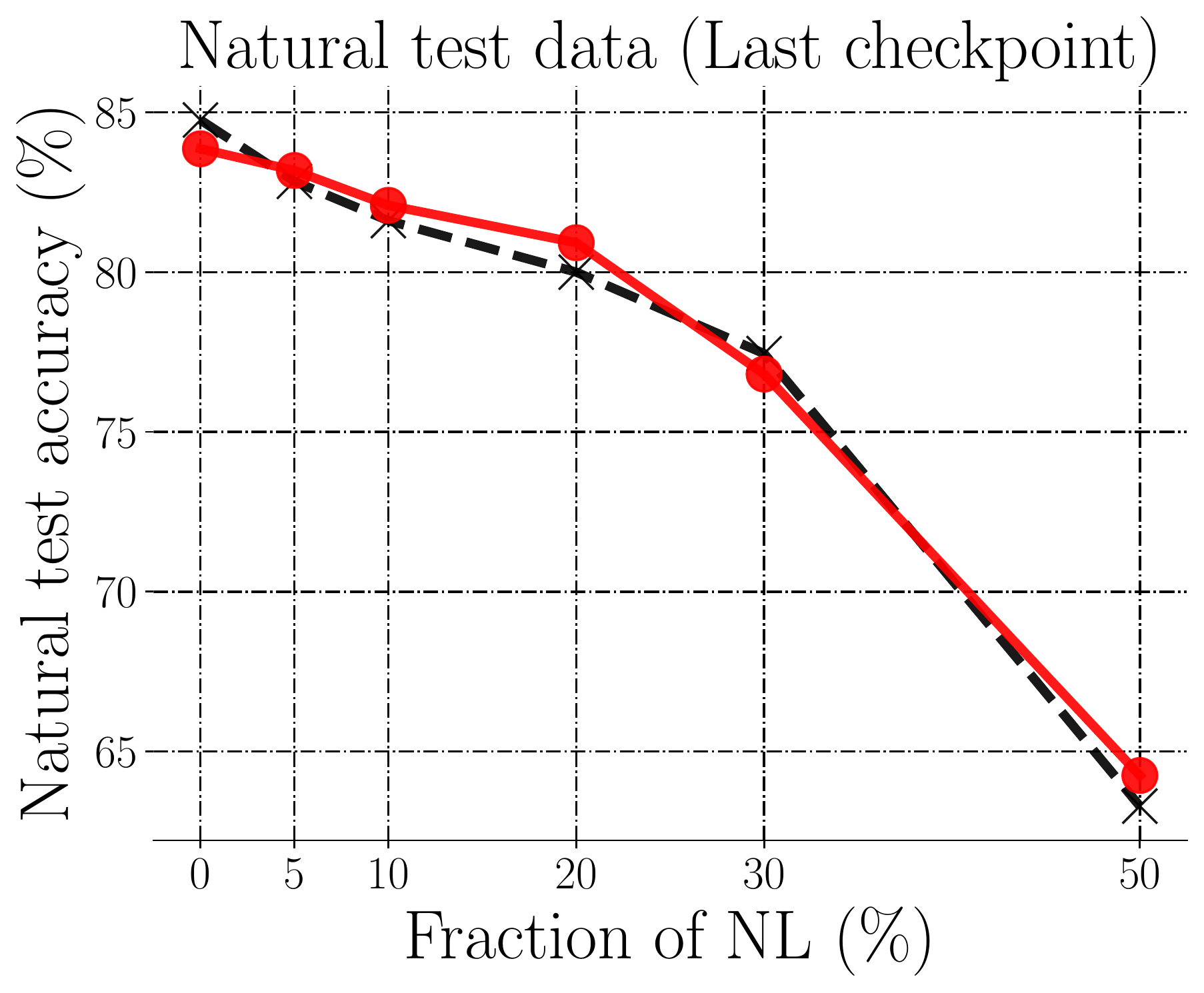}
		\includegraphics[width=0.235\textwidth]{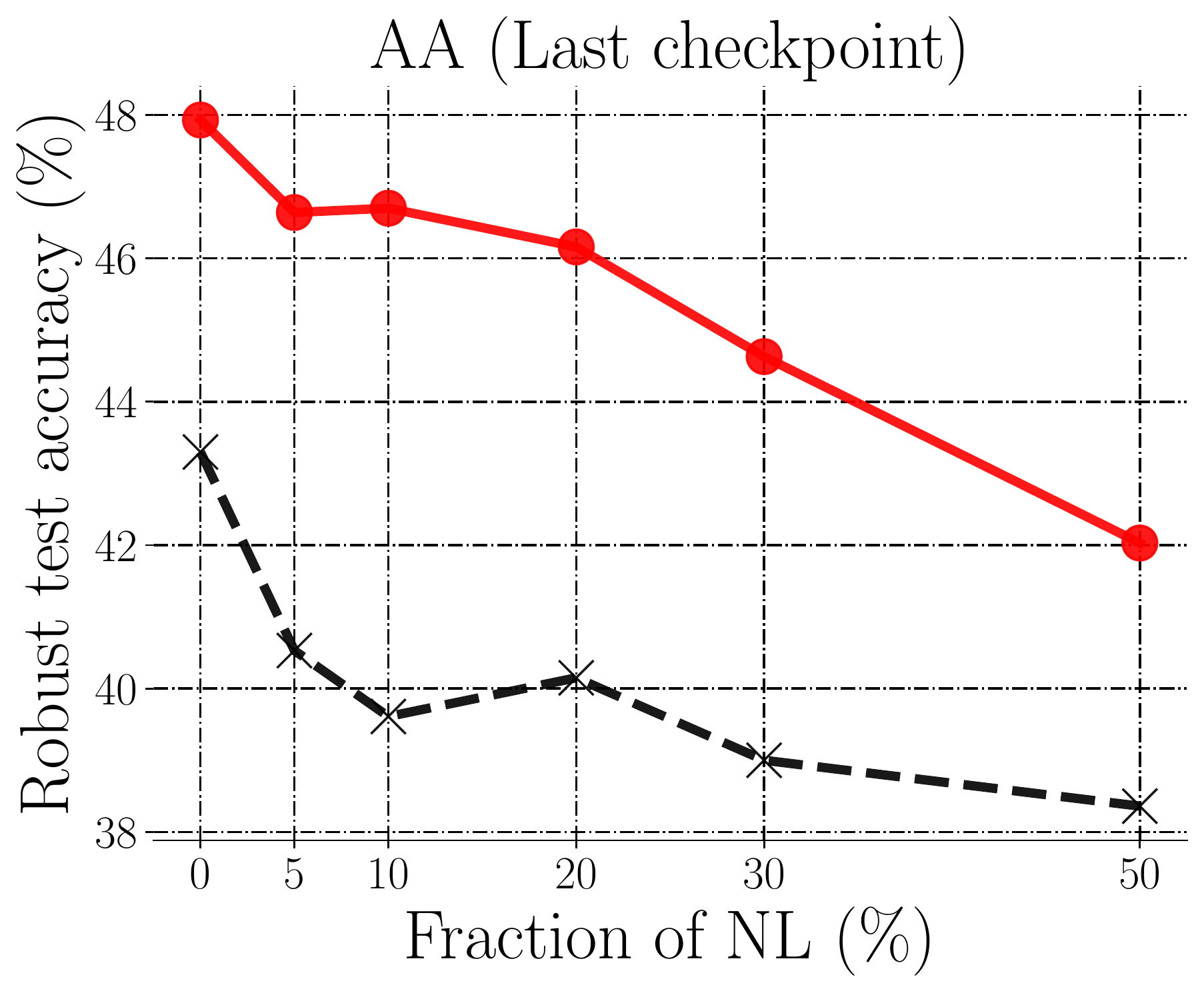}
	\vspace{-2mm}
	\caption{Comparisons between SAT and SAT-NoiLIn on noisy-label training set containing different fractions of NL.}
	\label{fig:diff_noisy_dataset}
	\vspace{-2mm}
\end{figure}

\paragraph{NoiLIn under noisy-label training set.}
Readers may feel curious about what happens when NoiLIn meets noisy-label set because training set is often noisy-labeled~\cite{Tong_xiao_noisy_label}. 
We compare AT-NoiLIn and SAT on noisy CIFAR-10 dataset using ResNet-18. We use symmetric-flipping NL to construct noisy CIFAR-10. The fraction of NL is chosen from $\{0\%, 5\%,10\%,20\%,30\%,50\%\}$. Note that when the fraction of NL is $0\%$, the training set is exactly the clean CIFAR-10 dataset. The training settings of SAT and AT-NoiLIn kept same as Section~\ref{sec:robust-overfitting}. 
In Figure~\ref{fig:diff_noisy_dataset}, we show the performance of SAT (black lines) and AT-NoiLIn (red lines) on noisy-label training set that contains different fractions of NL. 

We find red lines are always above black lines, which indicates that even with noisy-label training set, our NoiLIn is still an effective method to improve the AT methods.  



{\paragraph{NoiLIn with extra unlabeled training data}
~\citet{carmon2019unlabeled} and~\citet{gowal2020uncovering} showed that additional unlabeled data can enhance adversarial robustness. Therefore, we conduct NoiLIn with additional training unlabeled data provided by~\citet{carmon2019unlabeled}. We train WRN-28-10 with the cosine decay learning rate schedule. We set the ratio of labeled-to-unlabelled data per batch to 3:7 following~\citet{gowal2020uncovering}. The total training epoch is 200 and the batch size is 256. Other hyper-parameters for injecting label noise are the same as Section~\ref{sec:exp}. In Table~\ref{tab:wrn_NN_extra}, we compare the natural test accuracy and robust test accuracy (evaluated by AA attack) between NoiLIn and Robust Self-training~\citep{carmon2019unlabeled}. Note that the results of Robust Self-training with extra data~\cite{carmon2019unlabeled} are copied from \href{https://github.com/fra31/auto-attack}{AutoAttack leaderboard}~\citep{croce2020robustbench}. Surprisingly, Table~\ref{tab:wrn_NN_extra} shows that NoiLIn can enhance both natural and robust test accuracy, which validates the superiority of the NoiLIn method.}
\begin{table}[h!]
	\centering
	\vspace{-2mm}
	\caption{Evaluations of NoiLIn with additional unlabeled data on WRN-28-10. 
	}
\vspace{1mm}
	\label{tab:wrn_NN_extra}
		\begin{tabular}{c|cc}
			\hline
			Defense & Natural & AA \\ \hline
			Robust Self-teraining by~\cite{carmon2019unlabeled} &   89.69  & 59.53 \\
			NoiLIn with additional unlabeled data &  \textbf{91.11}   & \textbf{59.91} \\ \hline
		\end{tabular}
		\vspace{-2mm}
\end{table}

\vspace{0mm}
\section{Conclusion}
\label{sec:conclusion}
In this paper, we claim that NL injection can benefit AT.
We have explored the positive effects of NL injection in inner maximization and in outer minimization, respectively. Our observations have motivated us to propose a simple but effective strategy, namely ``NoiLIn'', to combat the issue of robust overfitting and even enhance robustness further (e.g., especially under larger $\epsilon$). 

There are two limitations of our current work. 1) We empirically verified the benefits of NoiLIn, but it is still difficult to explicitly, not intuitively, explain why NL injection improves robustness. 2) We do not know the optimal rate of NL injection at each training epoch, except leveraging validation set. In the future, we will attempt to address these limitations.

\subsection*{Acknowledgment}
JZ was supported by JST Strategic Basic Research Programs, ACT-X, Grant No. JPMJAX21AF and JSPS Grants-in-Aid for Scientific Research (KAKENHI), Early-Career Scientists, Grant No. 22K17955, Japan. 
BH was supported by the RGC Early Career Scheme No. 22200720, NSFC Young Scientists Fund No. 62006202, Guangdong Basic and Applied Basic Research Foundation No. 2022A1515011652, RIKEN Collaborative Research Fund and HKBU CSD Departmental Incentive Grant. 
LTL was partially supported by the Australian Research Council Project DP180103424, DE190101473, IC190100031, DP220102121.
LC was supported by the National Key R$\&$D Program of China No. 2021YFF0900800, NSFC No.91846205, SDNSFC No.ZR2019LZH008, Shandong Provincial Key Research and Development Program (Major Scientific and Technological Innovation Project) No.2021CXGC010108. 
MS was supported by JST AIP Acceleration Research Grant Number JP- MJCR20U3 and the Institute for AI and Beyond, UTokyo

\bibliography{egbib}

\begin{thebibliography}{115}
\providecommand{\natexlab}[1]{#1}
\providecommand{\url}[1]{\texttt{#1}}
\expandafter\ifx\csname urlstyle\endcsname\relax
  \providecommand{\doi}[1]{doi: #1}\else
  \providecommand{\doi}{doi: \begingroup \urlstyle{rm}\Url}\fi

\bibitem[Alayrac et~al.(2019)Alayrac, Uesato, Huang, Fawzi, Stanforth, and
  Kohli]{DeepMind_useto}
Jean{-}Baptiste Alayrac, Jonathan Uesato, Po{-}Sen Huang, Alhussein Fawzi,
  Robert Stanforth, and Pushmeet Kohli.
\newblock Are labels required for improving adversarial robustness?
\newblock In \emph{NeurIPS}, 2019.

\bibitem[Angluin \& Laird(1988)Angluin and Laird]{angluin1988learning}
Dana Angluin and Philip Laird.
\newblock Learning from noisy examples.
\newblock \emph{Machine Learning}, 2\penalty0 (4):\penalty0 343--370, 1988.

\bibitem[Athalye et~al.(2018)Athalye, Carlini, and
  Wagner]{Athalye_ICML_18_Obfuscated_Gradients}
Anish Athalye, Nicholas Carlini, and David~A. Wagner.
\newblock Obfuscated gradients give a false sense of security: Circumventing
  defenses to adversarial examples.
\newblock In \emph{ICML}, 2018.

\bibitem[Balunovic \& Vechev(2020)Balunovic and
  Vechev]{balunovic2020adversarial_certified_robustness}
Mislav Balunovic and Martin Vechev.
\newblock Adversarial training and provable defenses: Bridging the gap.
\newblock In \emph{ICLR}, 2020.

\bibitem[Cai et~al.(2018)Cai, Liu, and Song]{Cai_CAT}
Qi{-}Zhi Cai, Chang Liu, and Dawn Song.
\newblock Curriculum adversarial training.
\newblock In \emph{IJCAI}, 2018.

\bibitem[Carlini \& Wagner(2017{\natexlab{a}})Carlini and
  Wagner]{Carlini017_CW}
Nicholas Carlini and David~A. Wagner.
\newblock Towards evaluating the robustness of neural networks.
\newblock In \emph{Symposium on Security and Privacy (SP)}, 2017{\natexlab{a}}.

\bibitem[Carlini \& Wagner(2017{\natexlab{b}})Carlini and
  Wagner]{DBLP:conf/ccs/Carlini017}
Nicholas Carlini and David~A. Wagner.
\newblock Adversarial examples are not easily detected: Bypassing ten detection
  methods.
\newblock In \emph{Proceedings of the 10th {ACM} Workshop on Artificial
  Intelligence and Security, AISec@CCS}, 2017{\natexlab{b}}.

\bibitem[Carmon et~al.(2019)Carmon, Raghunathan, Schmidt, Liang, and
  Duchi]{carmon2019unlabeled}
Yair Carmon, Aditi Raghunathan, Ludwig Schmidt, Percy Liang, and John~C. Duchi.
\newblock Unlabeled data improves adversarial robustness.
\newblock In \emph{NeurIPS}, 2019.

\bibitem[Chen et~al.(2021{\natexlab{a}})Chen, Chen, Zhou, Qin, Mao, Zhang, and
  Yu]{conf/icassp/ChenC00MZY21}
Kejiang Chen, Yuefeng Chen, Hang Zhou, Chuan Qin, Xiaofeng Mao, Weiming Zhang,
  and Nenghai Yu.
\newblock Adversarial examples detection beyond image space.
\newblock In \emph{{IEEE} International Conference on Acoustics, Speech and
  Signal Processing, {ICASSP}}, 2021{\natexlab{a}}.

\bibitem[Chen et~al.(2020)Chen, Liu, Chang, Cheng, Amini, and
  Wang]{chen_CVPR_pretrain}
Tianlong Chen, Sijia Liu, Shiyu Chang, Yu~Cheng, Lisa Amini, and Zhangyang
  Wang.
\newblock Adversarial robustness: From self-supervised pre-training to
  fine-tuning.
\newblock In \emph{CVPR}, 2020.

\bibitem[Chen et~al.(2021{\natexlab{b}})Chen, Zhang, Liu, Chang, and
  Wang]{chen2021robust}
Tianlong Chen, Zhenyu Zhang, Sijia Liu, Shiyu Chang, and Zhangyang Wang.
\newblock Robust overfitting may be mitigated by properly learned smoothening.
\newblock In \emph{ICLR}, 2021{\natexlab{b}}.

\bibitem[Cheng et~al.(2020)Cheng, Lei, Chen, Dhillon, and Hsieh]{cheng2020cat}
Minhao Cheng, Qi~Lei, Pin-Yu Chen, Inderjit Dhillon, and Cho-Jui Hsieh.
\newblock Cat: Customized adversarial training for improved robustness.
\newblock \emph{arXiv:2002.06789}, 2020.

\bibitem[Cisse et~al.(2017)Cisse, Bojanowski, Grave, Dauphin, and
  Usunier]{cisse2017parseval}
Moustapha Cisse, Piotr Bojanowski, Edouard Grave, Yann Dauphin, and Nicolas
  Usunier.
\newblock Parseval networks: Improving robustness to adversarial examples.
\newblock In \emph{ICML}, 2017.

\bibitem[Cohen et~al.(2020)Cohen, Sapiro, and Giryes]{DBLP:conf/cvpr/CohenSG20}
Gilad Cohen, Guillermo Sapiro, and Raja Giryes.
\newblock Detecting adversarial samples using influence functions and nearest
  neighbors.
\newblock In \emph{CVPR}, 2020.

\bibitem[Cohen et~al.(2019)Cohen, Rosenfeld, and
  Kolter]{Jeremy_cohen_certified_robustness_random_smoothing}
Jeremy~M. Cohen, Elan Rosenfeld, and J.~Zico Kolter.
\newblock Certified adversarial robustness via randomized smoothing.
\newblock In \emph{ICML}, 2019.

\bibitem[Croce \& Hein(2020)Croce and Hein]{croce2020reliable}
Francesco Croce and Matthias Hein.
\newblock Reliable evaluation of adversarial robustness with an ensemble of
  diverse parameter-free attacks.
\newblock In \emph{ICML}, 2020.

\bibitem[Croce et~al.(2020)Croce, Andriushchenko, Sehwag, Debenedetti,
  Flammarion, Chiang, Mittal, and Hein]{croce2020robustbench}
Francesco Croce, Maksym Andriushchenko, Vikash Sehwag, Edoardo Debenedetti,
  Nicolas Flammarion, Mung Chiang, Prateek Mittal, and Matthias Hein.
\newblock Robustbench: a standardized adversarial robustness benchmark.
\newblock \emph{arXiv preprint arXiv:2010.09670}, 2020.

\bibitem[Deng et~al.(2009)Deng, Dong, Socher, Li, Li, and
  Fei-Fei]{deng2009imagenet}
Jia Deng, Wei Dong, Richard Socher, Li-Jia Li, Kai Li, and Li~Fei-Fei.
\newblock Imagenet: A large-scale hierarchical image database.
\newblock In \emph{2009 IEEE conference on computer vision and pattern
  recognition}, pp.\  248--255. Ieee, 2009.

\bibitem[Ding et~al.(2020)Ding, Sharma, Lui, and Huang]{ding2020mma}
Gavin~Weiguang Ding, Yash Sharma, Kry Yik~Chau Lui, and Ruitong Huang.
\newblock Mma training: Direct input space margin maximization through
  adversarial training.
\newblock In \emph{ICLR}, 2020.

\bibitem[Dong et~al.(2022)Dong, Xu, Yang, Pang, Deng, Su, and
  Zhu]{dong2022exploring}
Yinpeng Dong, Ke~Xu, Xiao Yang, Tianyu Pang, Zhijie Deng, Hang Su, and Jun Zhu.
\newblock Exploring memorization in adversarial training.
\newblock In \emph{ICLR}, 2022.

\bibitem[Donhauser et~al.(2021)Donhauser, Tifrea, Aerni, Heckel, and
  Yang]{donhauser2021interpolation}
Konstantin Donhauser, Alexandru Tifrea, Michael Aerni, Reinhard Heckel, and
  Fanny Yang.
\newblock Interpolation can hurt robust generalization even when there is no
  noise.
\newblock In \emph{NeurIPS}, 2021.

\bibitem[Du et~al.(2021)Du, Zhang, Han, Liu, Rong, Niu, Huang, and
  Sugiyama]{pmlr-v139-du21f}
Xuefeng Du, Jingfeng Zhang, Bo~Han, Tongliang Liu, Yu~Rong, Gang Niu, Junzhou
  Huang, and Masashi Sugiyama.
\newblock Learning diverse-structured networks for adversarial robustness.
\newblock In \emph{ICML}, 2021.

\bibitem[Engstrom et~al.(2018)Engstrom, Ilyas, and
  Athalye]{engstrom2018evaluating}
Logan Engstrom, Andrew Ilyas, and Anish Athalye.
\newblock Evaluating and understanding the robustness of adversarial logit
  pairing.
\newblock \emph{arXiv preprint arXiv:1807.10272}, 2018.

\bibitem[Farnia et~al.(2019)Farnia, Zhang, and Tse]{farnia2018generalizable}
Farzan Farnia, Jesse~M. Zhang, and David Tse.
\newblock Generalizable adversarial training via spectral normalization.
\newblock In \emph{ICLR}, 2019.

\bibitem[Gong et~al.(2021)Gong, Ren, Ye, and Liu]{Gong_2021_CVPR}
Chengyue Gong, Tongzheng Ren, Mao Ye, and Qiang Liu.
\newblock Maxup: Lightweight adversarial training with data augmentation
  improves neural network training.
\newblock In \emph{CVPR}, 2021.

\bibitem[Goodfellow et~al.(2015)Goodfellow, Shlens, and
  Szegedy]{Goodfellow14_Adversarial_examples}
Ian~J. Goodfellow, Jonathon Shlens, and Christian Szegedy.
\newblock Explaining and harnessing adversarial examples.
\newblock In \emph{ICLR}, 2015.

\bibitem[Gowal et~al.(2021{\natexlab{a}})Gowal, Qin, Uesato, Mann, and
  Kohli]{gowal2020uncovering}
Sven Gowal, Chongli Qin, Jonathan Uesato, Timothy Mann, and Pushmeet Kohli.
\newblock Uncovering the limits of adversarial training against norm-bounded
  adversarial examples.
\newblock \emph{arXiv:2010.03593}, 2021{\natexlab{a}}.

\bibitem[Gowal et~al.(2021{\natexlab{b}})Gowal, Rebuffi, Wiles, Stimberg,
  Calian, and Mann]{gowal2021improving}
Sven Gowal, Sylvestre-Alvise Rebuffi, Olivia Wiles, Florian Stimberg,
  Dan~Andrei Calian, and Timothy Mann.
\newblock Improving robustness using generated data.
\newblock In \emph{NeurIPS}, 2021{\natexlab{b}}.

\bibitem[Guo et~al.(2020)Guo, Yang, Xu, Liu, and Lin]{guo2020meets}
Minghao Guo, Yuzhe Yang, Rui Xu, Ziwei Liu, and Dahua Lin.
\newblock When nas meets robustness: In search of robust architectures against
  adversarial attacks.
\newblock In \emph{CVPR}, 2020.

\bibitem[Han et~al.(2018)Han, Yao, Yu, Niu, Xu, Hu, Tsang, and
  Sugiyama]{han2018co}
Bo~Han, Quanming Yao, Xingrui Yu, Gang Niu, Miao Xu, Weihua Hu, Ivor Tsang, and
  Masashi Sugiyama.
\newblock Co-teaching: Robust training of deep neural networks with extremely
  noisy labels.
\newblock In \emph{NeurIPS}, 2018.

\bibitem[He et~al.(2016)He, Zhang, Ren, and Sun]{he2016deep}
Kaiming He, Xiangyu Zhang, Shaoqing Ren, and Jian Sun.
\newblock Deep residual learning for image recognition.
\newblock In \emph{CVPR}, 2016.

\bibitem[Hein \& Andriushchenko(2017)Hein and Andriushchenko]{hein2017formal}
Matthias Hein and Maksym Andriushchenko.
\newblock Formal guarantees on the robustness of a classifier against
  adversarial manipulation.
\newblock In \emph{NeurIPS}, 2017.

\bibitem[Huang et~al.(2021)Huang, Wang, Erfani, Gu, Bailey, and
  Ma]{huang2021exploring}
Hanxun Huang, Yisen Wang, Sarah~Monazam Erfani, Quanquan Gu, James Bailey, and
  Xingjun Ma.
\newblock Exploring architectural ingredients of adversarially robust deep
  neural networks.
\newblock In \emph{NeurIPS}, 2021.

\bibitem[Ilyas et~al.(2019)Ilyas, Santurkar, Tsipras, Engstrom, Tran, and
  Madry]{robust_features_nips2019_madry}
Andrew Ilyas, Shibani Santurkar, Dimitris Tsipras, Logan Engstrom, Brandon
  Tran, and Aleksander Madry.
\newblock Adversarial examples are not bugs, they are features.
\newblock In \emph{NeurIPS}, 2019.

\bibitem[Jiang et~al.(2020)Jiang, Chen, Chen, and Wang]{Jiang_NIPS_pretrain}
Ziyu Jiang, Tianlong Chen, Ting Chen, and Zhangyang Wang.
\newblock Robust pre-training by adversarial contrastive learning.
\newblock In \emph{NeurIPS}, 2020.

\bibitem[Khachaturov et~al.(2021)Khachaturov, Shumailov, Zhao, Papernot, and
  Anderson]{khachaturov2021markpainting}
David Khachaturov, Ilia Shumailov, Yiren Zhao, Nicolas Papernot, and Ross
  Anderson.
\newblock Markpainting: Adversarial machine learning meets inpainting.
\newblock In \emph{ICML}, 2021.

\bibitem[Krizhevsky(2009)]{krizhevsky2009learning_cifar10}
Alex Krizhevsky.
\newblock Learning multiple layers of features from tiny images.
\newblock Technical report, 2009.

\bibitem[Laine \& Aila(2017)Laine and Aila]{laine2016temporal}
Samuli Laine and Timo Aila.
\newblock Temporal ensembling for semi-supervised learning.
\newblock In \emph{ICLR}, 2017.

\bibitem[L{\'{e}}cuyer et~al.(2019)L{\'{e}}cuyer, Atlidakis, Geambasu, Hsu, and
  Jana]{Lecuyer_certified_robustness_with_DP}
Mathias L{\'{e}}cuyer, Vaggelis Atlidakis, Roxana Geambasu, Daniel Hsu, and
  Suman Jana.
\newblock Certified robustness to adversarial examples with differential
  privacy.
\newblock In \emph{Symposium on Security and Privacy (SP)}, 2019.

\bibitem[Lee et~al.(2018)Lee, Lee, Lee, and Shin]{DBLP:conf/nips/LeeLLS18}
Kimin Lee, Kibok Lee, Honglak Lee, and Jinwoo Shin.
\newblock A simple unified framework for detecting out-of-distribution samples
  and adversarial attacks.
\newblock In \emph{NeurIPS}, 2018.

\bibitem[Lee et~al.(2020)Lee, Lee, and Yoon]{Lee_2020_CVPR}
Saehyung Lee, Hyungyu Lee, and Sungroh Yoon.
\newblock Adversarial vertex mixup: Toward better adversarially robust
  generalization.
\newblock In \emph{CVPR}, 2020.

\bibitem[Li \& Li(2017)Li and Li]{DBLP:conf/iccv/LiL17}
Xin Li and Fuxin Li.
\newblock Adversarial examples detection in deep networks with convolutional
  filter statistics.
\newblock In \emph{ICCV}, 2017.

\bibitem[Li et~al.(2021)Li, Yang, Wang, and Xu]{li2021neural}
Yanxi Li, Zhaohui Yang, Yunhe Wang, and Chang Xu.
\newblock Neural architecture dilation for adversarial robustness.
\newblock In \emph{NeurIPS}, 2021.

\bibitem[Liu et~al.(2019)Liu, Zhang, Zhang, Hou, Liu, Zha, and
  Yu]{DBLP:conf/cvpr/LiuZZHLZY19}
Jiayang Liu, Weiming Zhang, Yiwei Zhang, Dongdong Hou, Yujia Liu, Hongyue Zha,
  and Nenghai Yu.
\newblock Detection based defense against adversarial examples from the
  steganalysis point of view.
\newblock In \emph{CVPR}, 2019.

\bibitem[Ma et~al.(2018)Ma, Li, Wang, Erfani, Wijewickrema, Schoenebeck, Song,
  Houle, and Bailey]{DBLP:conf/iclr/Ma0WEWSSHB18}
Xingjun Ma, Bo~Li, Yisen Wang, Sarah~M. Erfani, Sudanthi N.~R. Wijewickrema,
  Grant Schoenebeck, Dawn Song, Michael~E. Houle, and James Bailey.
\newblock Characterizing adversarial subspaces using local intrinsic
  dimensionality.
\newblock In \emph{ICLR}, 2018.

\bibitem[Madry et~al.(2018)Madry, Makelov, Schmidt, Tsipras, and
  Vladu]{Madry_adversarial_training}
Aleksander Madry, Aleksandar Makelov, Ludwig Schmidt, Dimitris Tsipras, and
  Adrian Vladu.
\newblock Towards deep learning models resistant to adversarial attacks.
\newblock In \emph{ICLR}, 2018.

\bibitem[Metzen et~al.(2017)Metzen, Genewein, Fischer, and
  Bischoff]{DBLP:conf/iclr/MetzenGFB17}
Jan~Hendrik Metzen, Tim Genewein, Volker Fischer, and Bastian Bischoff.
\newblock On detecting adversarial perturbations.
\newblock In \emph{ICLR}, 2017.

\bibitem[Mirman et~al.(2018)Mirman, Gehr, and
  Vechev]{icml/MirmanGV18_internal_bound_propagation}
Matthew Mirman, Timon Gehr, and Martin~T. Vechev.
\newblock Differentiable abstract interpretation for provably robust neural
  networks.
\newblock In \emph{ICML}, 2018.

\bibitem[Moosavi-Dezfooli et~al.(2019)Moosavi-Dezfooli, Fawzi, Uesato, and
  Frossard]{moosavi2019robustness_curvature}
Seyed-Mohsen Moosavi-Dezfooli, Alhussein Fawzi, Jonathan Uesato, and Pascal
  Frossard.
\newblock Robustness via curvature regularization, and vice versa.
\newblock In \emph{CVPR}, 2019.

\bibitem[Mosbach et~al.(2018)Mosbach, Andriushchenko, Trost, Hein, and
  Klakow]{mosbach2018logit}
Marius Mosbach, Maksym Andriushchenko, Thomas Trost, Matthias Hein, and
  Dietrich Klakow.
\newblock Logit pairing methods can fool gradient-based attacks.
\newblock \emph{arXiv preprint arXiv:1810.12042}, 2018.

\bibitem[M{\"u}ller et~al.(2019)M{\"u}ller, Kornblith, and
  Hinton]{muller2019does}
Rafael M{\"u}ller, Simon Kornblith, and Geoffrey~E Hinton.
\newblock When does label smoothing help?
\newblock \emph{Advances in neural information processing systems}, 32, 2019.

\bibitem[Najafi et~al.(2019)Najafi, Maeda, Koyama, and
  Miyato]{najafi2019robustness}
Amir Najafi, Shin{-}ichi Maeda, Masanori Koyama, and Takeru Miyato.
\newblock Robustness to adversarial perturbations in learning from incomplete
  data.
\newblock In \emph{NeurIPS}, 2019.

\bibitem[Natarajan et~al.(2013)Natarajan, Dhillon, Ravikumar, and
  Tewari]{natarajan2013learning}
Nagarajan Natarajan, Inderjit~S Dhillon, Pradeep~K Ravikumar, and Ambuj Tewari.
\newblock Learning with noisy labels.
\newblock In \emph{NeurIPS}, 2013.

\bibitem[Pang et~al.(2018)Pang, Du, Dong, and Zhu]{DBLP:conf/nips/PangDDZ18}
Tianyu Pang, Chao Du, Yinpeng Dong, and Jun Zhu.
\newblock Towards robust detection of adversarial examples.
\newblock In \emph{NeurIPS}, 2018.

\bibitem[Pang et~al.(2019)Pang, Xu, Du, Chen, and
  Zhu]{Pang_ICML_19_AT_Ensemble}
Tianyu Pang, Kun Xu, Chao Du, Ning Chen, and Jun Zhu.
\newblock Improving adversarial robustness via promoting ensemble diversity.
\newblock In \emph{ICML}, 2019.

\bibitem[Pang et~al.(2021)Pang, Yang, Dong, Su, and Zhu]{pang2021bag}
Tianyu Pang, Xiao Yang, Yinpeng Dong, Hang Su, and Jun Zhu.
\newblock Bag of tricks for adversarial training.
\newblock \emph{ICLR}, 2021.

\bibitem[Qin et~al.(2019)Qin, Martens, Gowal, Krishnan, Dvijotham, Fawzi, De,
  Stanforth, and Kohli]{local_linearilization}
Chongli Qin, James Martens, Sven Gowal, Dilip Krishnan, Krishnamurthy
  Dvijotham, Alhussein Fawzi, Soham De, Robert Stanforth, and Pushmeet Kohli.
\newblock Adversarial robustness through local linearization.
\newblock In \emph{NeurIPS}, 2019.

\bibitem[Qin et~al.(2020)Qin, Frosst, Sabour, Raffel, Cottrell, and
  Hinton]{DBLP:conf/iclr/QinFSRCH20}
Yao Qin, Nicholas Frosst, Sara Sabour, Colin Raffel, Garrison~W. Cottrell, and
  Geoffrey~E. Hinton.
\newblock Detecting and diagnosing adversarial images with class-conditional
  capsule reconstructions.
\newblock In \emph{ICLR}, 2020.

\bibitem[Raghunathan et~al.(2020)Raghunathan, Xie, Yang, Duchi, and
  Liang]{raghunathan2020understanding}
Aditi Raghunathan, Sang~Michael Xie, Fanny Yang, John Duchi, and Percy Liang.
\newblock Understanding and mitigating the tradeoff between robustness and
  accuracy.
\newblock In \emph{ICML}, 2020.

\bibitem[Rice et~al.(2020)Rice, Wong, and Kolter]{rice2020overfitting}
Leslie Rice, Eric Wong, and J~Zico Kolter.
\newblock Overfitting in adversarially robust deep learning.
\newblock In \emph{ICML}, 2020.

\bibitem[Ross \& Doshi-Velez(2018)Ross and Doshi-Velez]{ross2018improving}
Andrew~Slavin Ross and Finale Doshi-Velez.
\newblock Improving the adversarial robustness and interpretability of deep
  neural networks by regularizing their input gradients.
\newblock In \emph{AAAI}, 2018.

\bibitem[Roth et~al.(2019)Roth, Kilcher, and Hofmann]{DBLP:conf/icml/RothKH19}
Kevin Roth, Yannic Kilcher, and Thomas Hofmann.
\newblock The odds are odd: {A} statistical test for detecting adversarial
  examples.
\newblock In \emph{ICML}, 2019.

\bibitem[Sanyal et~al.(2021)Sanyal, Dokania, Kanade, and
  Torr]{sanyal2020benign}
Amartya Sanyal, Puneet~K. Dokania, Varun Kanade, and Philip Torr.
\newblock How benign is benign overfitting ?
\newblock In \emph{ICLR}, 2021.

\bibitem[Schmidt et~al.(2018)Schmidt, Santurkar, Tsipras, Talwar, and
  Madry]{schmidt2018adversarial_more_data}
Ludwig Schmidt, Shibani Santurkar, Dimitris Tsipras, Kunal Talwar, and
  Aleksander Madry.
\newblock Adversarially robust generalization requires more data.
\newblock In \emph{NeurIPS}, 2018.

\bibitem[Sehwag et~al.(2020)Sehwag, Wang, Mittal, and Jana]{sehwag2020hydra}
Vikash Sehwag, Shiqi Wang, Prateek Mittal, and Suman Jana.
\newblock Hydra: Pruning adversarially robust neural networks.
\newblock In \emph{NeurIPS}, 2020.

\bibitem[Shafahi et~al.(2019{\natexlab{a}})Shafahi, Ghiasi, Huang, and
  Goldstein]{shafahi2019label}
Ali Shafahi, Amin Ghiasi, Furong Huang, and Tom Goldstein.
\newblock Label smoothing and logit squeezing: a replacement for adversarial
  training?
\newblock \emph{arXiv preprint arXiv:1910.11585}, 2019{\natexlab{a}}.

\bibitem[Shafahi et~al.(2019{\natexlab{b}})Shafahi, Najibi, Ghiasi, Xu,
  Dickerson, Studer, Davis, Taylor, and
  Goldstein]{Ali_NIPS19_adversarial_training_for_free}
Ali Shafahi, Mahyar Najibi, Mohammad~Amin Ghiasi, Zheng Xu, John Dickerson,
  Christoph Studer, Larry~S Davis, Gavin Taylor, and Tom Goldstein.
\newblock Adversarial training for free!
\newblock In \emph{NeurIPS}, 2019{\natexlab{b}}.

\bibitem[Sheikholeslami et~al.(2021)Sheikholeslami, Lotfi, and
  Kolter]{DBLP:conf/iclr/SheikholeslamiL21}
Fatemeh Sheikholeslami, Ali Lotfi, and J.~Zico Kolter.
\newblock Provably robust classification of adversarial examples with
  detection.
\newblock In \emph{ICLR}, 2021.

\bibitem[Singla \& Feizi(2020)Singla and Feizi]{icml_20_second_order}
Sahil Singla and Soheil Feizi.
\newblock Second-order provable defenses against adversarial attacks.
\newblock In \emph{ICML}, 2020.

\bibitem[Smith \& Gal(2018)Smith and Gal]{DBLP:conf/uai/SmithG18}
Lewis Smith and Yarin Gal.
\newblock Understanding measures of uncertainty for adversarial example
  detection.
\newblock In \emph{UAI}, 2018.

\bibitem[Song et~al.(2019)Song, He, Wang, and
  Hopcroft]{song_iclr2019_domain_adaptation}
Chuanbiao Song, Kun He, Liwei Wang, and John~E. Hopcroft.
\newblock Improving the generalization of adversarial training with domain
  adaptation.
\newblock In \emph{ICLR}, 2019.

\bibitem[Song et~al.(2020)Song, He, Lin, Wang, and Hopcroft]{song_iclr20_RBS}
Chuanbiao Song, Kun He, Jiadong Lin, Liwei Wang, and John~E. Hopcroft.
\newblock Robust local features for improving the generalization of adversarial
  training.
\newblock In \emph{ICLR}, 2020.

\bibitem[Sperl et~al.(2020)Sperl, Kao, Chen, Lei, and
  B{\"{o}}ttinger]{DBLP:conf/eurosp/SperlKCLB20}
Philip Sperl, Ching{-}Yu Kao, Peng Chen, Xiao Lei, and Konstantin
  B{\"{o}}ttinger.
\newblock {DLA:} dense-layer-analysis for adversarial example detection.
\newblock In \emph{{IEEE} European Symposium on Security and Privacy,
  EuroS{\&}P}, 2020.

\bibitem[Sriramanan et~al.(2020)Sriramanan, Addepalli, Baburaj, and
  R.]{sriramanan2020guided}
Gaurang Sriramanan, Sravanti Addepalli, Arya Baburaj, and Venkatesh~Babu R.
\newblock Guided adversarial attack for evaluating and enhancing adversarial
  defenses.
\newblock In \emph{NeurIPS}, 2020.

\bibitem[Stutz et~al.(2019)Stutz, Hein, and Schiele]{stutz2019disentangling}
David Stutz, Matthias Hein, and Bernt Schiele.
\newblock Disentangling adversarial robustness and generalization.
\newblock In \emph{CVPR}, 2019.

\bibitem[Szegedy et~al.(2014)Szegedy, Zaremba, Sutskever, Bruna, Erhan,
  Goodfellow, and Fergus]{szegedy}
Christian Szegedy, Wojciech Zaremba, Ilya Sutskever, Joan Bruna, Dumitru Erhan,
  Ian Goodfellow, and Rob Fergus.
\newblock Intriguing properties of neural networks.
\newblock In \emph{ICLR}, 2014.

\bibitem[Szegedy et~al.(2016)Szegedy, Vanhoucke, Ioffe, Shlens, and
  Wojna]{szegedy2016rethinking}
Christian Szegedy, Vincent Vanhoucke, Sergey Ioffe, Jon Shlens, and Zbigniew
  Wojna.
\newblock Rethinking the inception architecture for computer vision.
\newblock In \emph{Proceedings of the IEEE conference on computer vision and
  pattern recognition}, pp.\  2818--2826, 2016.

\bibitem[Tian et~al.(2021)Tian, Zhou, Li, and Duan]{DBLP:conf/aaai/Tian0LD21}
Jinyu Tian, Jiantao Zhou, Yuanman Li, and Jia Duan.
\newblock Detecting adversarial examples from sensitivity inconsistency of
  spatial-transform domain.
\newblock In \emph{AAAI}, 2021.

\bibitem[Tian et~al.(2018)Tian, Yang, and Cai]{DBLP:conf/aaai/TianYC18}
Shixin Tian, Guolei Yang, and Ying Cai.
\newblock Detecting adversarial examples through image transformation.
\newblock In \emph{AAAI}, 2018.

\bibitem[Tram{\`{e}}r et~al.(2018)Tram{\`{e}}r, Kurakin, Papernot, Goodfellow,
  Boneh, and McDaniel]{Tramer_iclr_18}
Florian Tram{\`{e}}r, Alexey Kurakin, Nicolas Papernot, Ian~J. Goodfellow, Dan
  Boneh, and Patrick~D. McDaniel.
\newblock Ensemble adversarial training: Attacks and defenses.
\newblock In \emph{ICLR}, 2018.

\bibitem[Tsipras et~al.(2019)Tsipras, Santurkar, Engstrom, Turner, and
  Madry]{tsipras19_robustness_at_odd}
Dimitris Tsipras, Shibani Santurkar, Logan Engstrom, Alexander Turner, and
  Aleksander Madry.
\newblock Robustness may be at odds with accuracy.
\newblock In \emph{ICLR}, 2019.

\bibitem[Tsuzuku et~al.(2018)Tsuzuku, Sato, and
  Sugiyama]{Tsuzuku_Lipschitz_margin_training_scalable_certification}
Yusuke Tsuzuku, Issei Sato, and Masashi Sugiyama.
\newblock Lipschitz-{M}argin training: Scalable certification of perturbation
  invariance for deep neural networks.
\newblock In \emph{NeurIPS}, 2018.

\bibitem[Van~Rooyen et~al.(2015)Van~Rooyen, Menon, and
  Williamson]{van2015learning}
Brendan Van~Rooyen, Aditya~Krishna Menon, and Robert~C Williamson.
\newblock Learning with symmetric label noise: The importance of being
  unhinged.
\newblock In \emph{NeurIPS}, 2015.

\bibitem[Vivek \& Babu(2020)Vivek and Babu]{Babu_cvpr_2020}
B.~S. Vivek and R.~Venkatesh Babu.
\newblock Single-step adversarial training with dropout scheduling.
\newblock In \emph{CVPR}, 2020.

\bibitem[Wang et~al.(2019)Wang, Ma, Bailey, Yi, Zhou, and
  Gu]{Wang_Xingjun_MA_FOSC_DAT}
Yisen Wang, Xingjun Ma, James Bailey, Jinfeng Yi, Bowen Zhou, and Quanquan Gu.
\newblock On the convergence and robustness of adversarial training.
\newblock In \emph{ICML}, 2019.

\bibitem[Wang et~al.(2020)Wang, Zou, Yi, Bailey, Ma, and
  Gu]{wang2020improving_MART}
Yisen Wang, Difan Zou, Jinfeng Yi, James Bailey, Xingjun Ma, and Quanquan Gu.
\newblock Improving adversarial robustness requires revisiting misclassified
  examples.
\newblock In \emph{ICLR}, 2020.

\bibitem[Weng et~al.(2018)Weng, Zhang, Chen, Yi, Su, Gao, Hsieh, and
  Daniel]{Weng_Evaluating_robustness_extreme_value_approach}
Tsui{-}Wei Weng, Huan Zhang, Pin{-}Yu Chen, Jinfeng Yi, Dong Su, Yupeng Gao,
  Cho{-}Jui Hsieh, and Luca Daniel.
\newblock Evaluating the robustness of neural networks: An extreme value theory
  approach.
\newblock In \emph{ICLR}, 2018.

\bibitem[Wong \& Kolter(2018)Wong and
  Kolter]{Eric_Wong_provable_defence_convex_polytope}
Eric Wong and J.~Zico Kolter.
\newblock Provable defenses against adversarial examples via the convex outer
  adversarial polytope.
\newblock In \emph{ICML}, 2018.

\bibitem[Wong et~al.(2020)Wong, Rice, and Kolter]{wong2020fast_zico_kolter}
Eric Wong, Leslie Rice, and J.~Zico Kolter.
\newblock Fast is better than free: Revisiting adversarial training.
\newblock In \emph{ICLR}, 2020.

\bibitem[Wu et~al.(2021{\natexlab{a}})Wu, Chen, Cai, He, and Gu]{wu2020wider}
Boxi Wu, Jinghui Chen, Deng Cai, Xiaofei He, and Quanquan Gu.
\newblock Do wider neural networks really help adversarial robustness?
\newblock In \emph{NeurIPS}, 2021{\natexlab{a}}.

\bibitem[Wu et~al.(2020)Wu, Xia, and Wang]{wu2020adversarial}
Dongxian Wu, Shu-Tao Xia, and Yisen Wang.
\newblock Adversarial weight perturbation helps robust generalization.
\newblock \emph{NeurIPS}, 2020.

\bibitem[Wu et~al.(2021{\natexlab{b}})Wu, Arora, Wu, and
  Yang]{DBLP:conf/aaai/WuAWY21}
Yuhang Wu, Sunpreet~S. Arora, Yanhong Wu, and Hao Yang.
\newblock Beating attackers at their own games: Adversarial example detection
  using adversarial gradient directions.
\newblock In \emph{AAAI}, 2021{\natexlab{b}}.

\bibitem[Xiao et~al.(2019)Xiao, Tjeng, Shafiullah, and
  Madry]{DBLP:conf/iclr/XiaoTSM19}
Kai~Yuanqing Xiao, Vincent Tjeng, Nur Muhammad~(Mahi) Shafiullah, and
  Aleksander Madry.
\newblock Training for faster adversarial robustness verification via inducing
  relu stability.
\newblock In \emph{ICLR}, 2019.

\bibitem[Xiao et~al.(2015)Xiao, Xia, Yang, Huang, and
  Wang]{Tong_xiao_noisy_label}
Tong Xiao, Tian Xia, Yi~Yang, Chang Huang, and Xiaogang Wang.
\newblock Learning from massive noisy labeled data for image classification.
\newblock In \emph{CVPR}, 2015.

\bibitem[Xie \& Yuille(2020)Xie and Yuille]{DBLP:conf/iclr/XieY20}
Cihang Xie and Alan~L. Yuille.
\newblock Intriguing properties of adversarial training at scale.
\newblock In \emph{ICLR}, 2020.

\bibitem[Xie et~al.(2020)Xie, Tan, Gong, Yuille, and Le]{xie2020smooth}
Cihang Xie, Mingxing Tan, Boqing Gong, Alan Yuille, and Quoc~V Le.
\newblock Smooth adversarial training.
\newblock \emph{arXiv preprint arXiv:2006.14536}, 2020.

\bibitem[Xie et~al.(2016)Xie, Wang, Wei, Wang, and Tian]{xie2016disturblabel}
Lingxi Xie, Jingdong Wang, Zhen Wei, Meng Wang, and Qi~Tian.
\newblock Disturblabel: Regularizing cnn on the loss layer.
\newblock In \emph{CVPR}, 2016.

\bibitem[Xu et~al.(2021)Xu, Liu, Li, Jain, and Tang]{xu2021robust}
Han Xu, Xiaorui Liu, Yaxin Li, Anil Jain, and Jiliang Tang.
\newblock To be robust or to be fair: Towards fairness in adversarial training.
\newblock In \emph{ICML}, 2021.

\bibitem[Yan et~al.(2021)Yan, Zhang, Niu, Feng, Tan, and
  Sugiyama]{pmlr-v139-yan21e}
Hanshu Yan, Jingfeng Zhang, Gang Niu, Jiashi Feng, Vincent Tan, and Masashi
  Sugiyama.
\newblock Cifs: Improving adversarial robustness of cnns via channel-wise
  importance-based feature selection.
\newblock In \emph{ICML}, 2021.

\bibitem[Yan et~al.(2018)Yan, Guo, and Zhang]{yan2018deep}
Ziang Yan, Yiwen Guo, and Changshui Zhang.
\newblock Deep defense: Training dnns with improved adversarial robustness.
\newblock In \emph{NeurIPS}, 2018.

\bibitem[Yang et~al.(2020{\natexlab{a}})Yang, Zhang, Dong, Inkawhich, Gardner,
  Touchet, Wilkes, Berry, and Li]{yang2020dverge}
Huanrui Yang, Jingyang Zhang, Hongliang Dong, Nathan Inkawhich, Andrew Gardner,
  Andrew Touchet, Wesley Wilkes, Heath Berry, and Hai Li.
\newblock Dverge: Diversifying vulnerabilities for enhanced robust generation
  of ensembles.
\newblock In \emph{NeurIPS}, 2020{\natexlab{a}}.

\bibitem[Yang et~al.(2020{\natexlab{b}})Yang, Chen, Hsieh, Wang, and
  Jordan]{DBLP:conf/aaai/YangCHWJ20}
Puyudi Yang, Jianbo Chen, Cho{-}Jui Hsieh, Jane{-}Ling Wang, and Michael~I.
  Jordan.
\newblock {ML-LOO:} detecting adversarial examples with feature attribution.
\newblock In \emph{AAAI}, 2020{\natexlab{b}}.

\bibitem[Yang et~al.(2020{\natexlab{c}})Yang, Rashtchian, Zhang, Salakhutdinov,
  and Chaudhuri]{yang2020closer}
Yao{-}Yuan Yang, Cyrus Rashtchian, Hongyang Zhang, Russ~R. Salakhutdinov, and
  Kamalika Chaudhuri.
\newblock A closer look at accuracy vs. robustness.
\newblock In \emph{NeurIPS}, 2020{\natexlab{c}}.

\bibitem[Yi et~al.(2021)Yi, Hou, Sun, Shang, Jiang, Liu, and
  Ma]{yi2021improved}
Mingyang Yi, Lu~Hou, Jiacheng Sun, Lifeng Shang, Xin Jiang, Qun Liu, and
  Zhi-Ming Ma.
\newblock Improved ood generalization via adversarial training and
  pre-training.
\newblock In \emph{ICML}, 2021.

\bibitem[Yin et~al.(2019)Yin, Ramchandran, and Bartlett]{Dong_Yin_adv_gen}
Dong Yin, Kannan Ramchandran, and Peter~L. Bartlett.
\newblock Rademacher complexity for adversarially robust generalization.
\newblock In \emph{ICML}, 2019.

\bibitem[Yin \& Rohde(2020)Yin and Rohde]{DBLP:conf/iclr/YinKR20}
Xuwang Yin and Soheil Kolouri Gustavo~K. Rohde.
\newblock {GAT:} generative adversarial training for adversarial example
  detection and robust classification.
\newblock In \emph{ICLR}, 2020.

\bibitem[Zagoruyko \& Komodakis(2016)Zagoruyko and Komodakis]{zagoruyko2016WRN}
Sergey Zagoruyko and Nikos Komodakis.
\newblock Wide residual networks.
\newblock \emph{arXiv:1605.07146}, 2016.

\bibitem[Zhang et~al.(2021{\natexlab{a}})Zhang, Cai, Lu, He, and
  Wang]{zhang2021towards}
Bohang Zhang, Tianle Cai, Zhou Lu, Di~He, and Liwei Wang.
\newblock Towards certifying l-infinity robustness using neural networks with
  l-inf-dist neurons.
\newblock In \emph{ICML}, 2021{\natexlab{a}}.

\bibitem[Zhang et~al.(2019{\natexlab{a}})Zhang, Zhang, Lu, Zhu, and
  Dong]{Lu_yiping_NIPS19_yopo}
Dinghuai Zhang, Tianyuan Zhang, Yiping Lu, Zhanxing Zhu, and Bin Dong.
\newblock You only propagate once: Accelerating adversarial training via
  maximal principle.
\newblock In \emph{NeurIPS}, 2019{\natexlab{a}}.

\bibitem[Zhang et~al.(2019{\natexlab{b}})Zhang, Yu, Jiao, Xing, Ghaoui, and
  Jordan]{Zhang_trades}
Hongyang Zhang, Yaodong Yu, Jiantao Jiao, Eric~P. Xing, Laurent~El Ghaoui, and
  Michael~I. Jordan.
\newblock Theoretically principled trade-off between robustness and accuracy.
\newblock In \emph{ICML}, 2019{\natexlab{b}}.

\bibitem[Zhang et~al.(2020{\natexlab{a}})Zhang, Chen, Xiao, Gowal, Stanforth,
  Li, Boning, and Hsieh]{zhang2020towards_certifiable}
Huan Zhang, Hongge Chen, Chaowei Xiao, Sven Gowal, Robert Stanforth, Bo~Li,
  Duane Boning, and Cho-Jui Hsieh.
\newblock Towards stable and efficient training of verifiably robust neural
  networks.
\newblock In \emph{ICLR}, 2020{\natexlab{a}}.

\bibitem[Zhang et~al.(2020{\natexlab{b}})Zhang, Xu, Han, Niu, Cui, Sugiyama,
  and Kankanhalli]{zhang2020fat}
Jingfeng Zhang, Xilie Xu, Bo~Han, Gang Niu, Lizhen Cui, Masashi Sugiyama, and
  Mohan Kankanhalli.
\newblock Attacks which do not kill training make adversarial learning
  stronger.
\newblock In \emph{ICML}, 2020{\natexlab{b}}.

\bibitem[Zhang et~al.(2021{\natexlab{b}})Zhang, Zhu, Niu, Han, Sugiyama, and
  Kankanhalli]{zhang2021geometryaware}
Jingfeng Zhang, Jianing Zhu, Gang Niu, Bo~Han, Masashi Sugiyama, and Mohan
  Kankanhalli.
\newblock Geometry-aware instance-reweighted adversarial training.
\newblock In \emph{ICLR}, 2021{\natexlab{b}}.

\bibitem[Zhang \& Zhu(2019)Zhang and Zhu]{zhang2019interpreting}
Tianyuan Zhang and Zhanxing Zhu.
\newblock Interpreting adversarially trained convolutional neural networks.
\newblock In \emph{ICML}, 2019.

\bibitem[Zhu et~al.(2021)Zhu, Zhang, Han, Liu, Niu, Yang, Kankanhalli, and
  Sugiyama]{zhu2021understanding}
Jianing Zhu, Jingfeng Zhang, Bo~Han, Tongliang Liu, Gang Niu, Hongxia Yang,
  Mohan Kankanhalli, and Masashi Sugiyama.
\newblock Understanding the interaction of adversarial training with noisy
  labels.
\newblock \emph{arXiv:2102.03482}, 2021.

\end{thebibliography}
\bibliographystyle{tmlr}

\clearpage
\appendix

\section{NL Injection in Inner Maximization}
\label{appendix:inner_NL}
\begin{figure}[tp!]
	\centering
	\vspace{-0mm}
	\includegraphics[width=0.244\textwidth]{figures/inner_min/r18_eps8_NC/symmetric/N_C_symmetric_eps8_natural_test_acc.pdf}
	\includegraphics[width=0.244\textwidth]{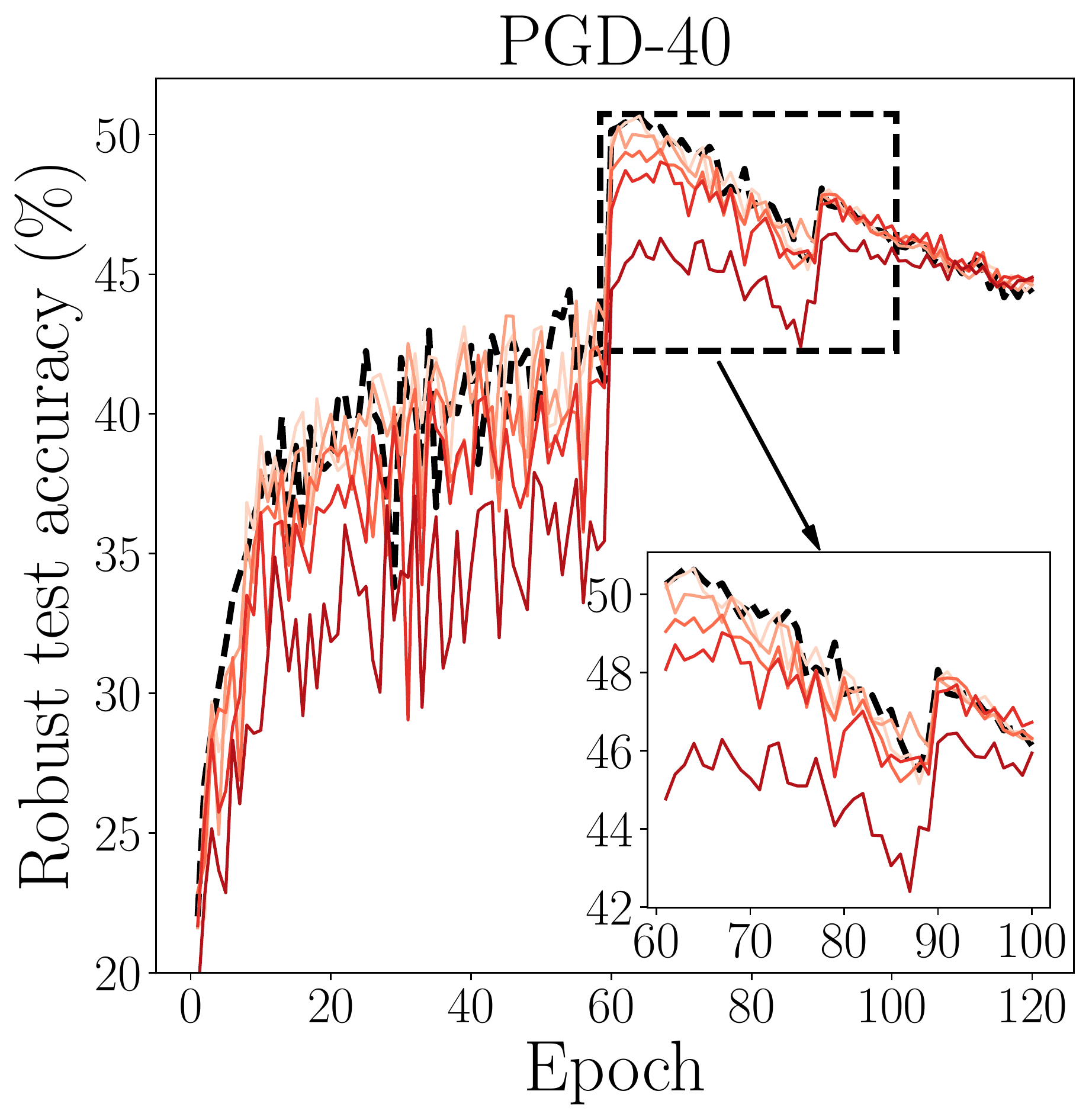}
	\includegraphics[width=0.244\textwidth]{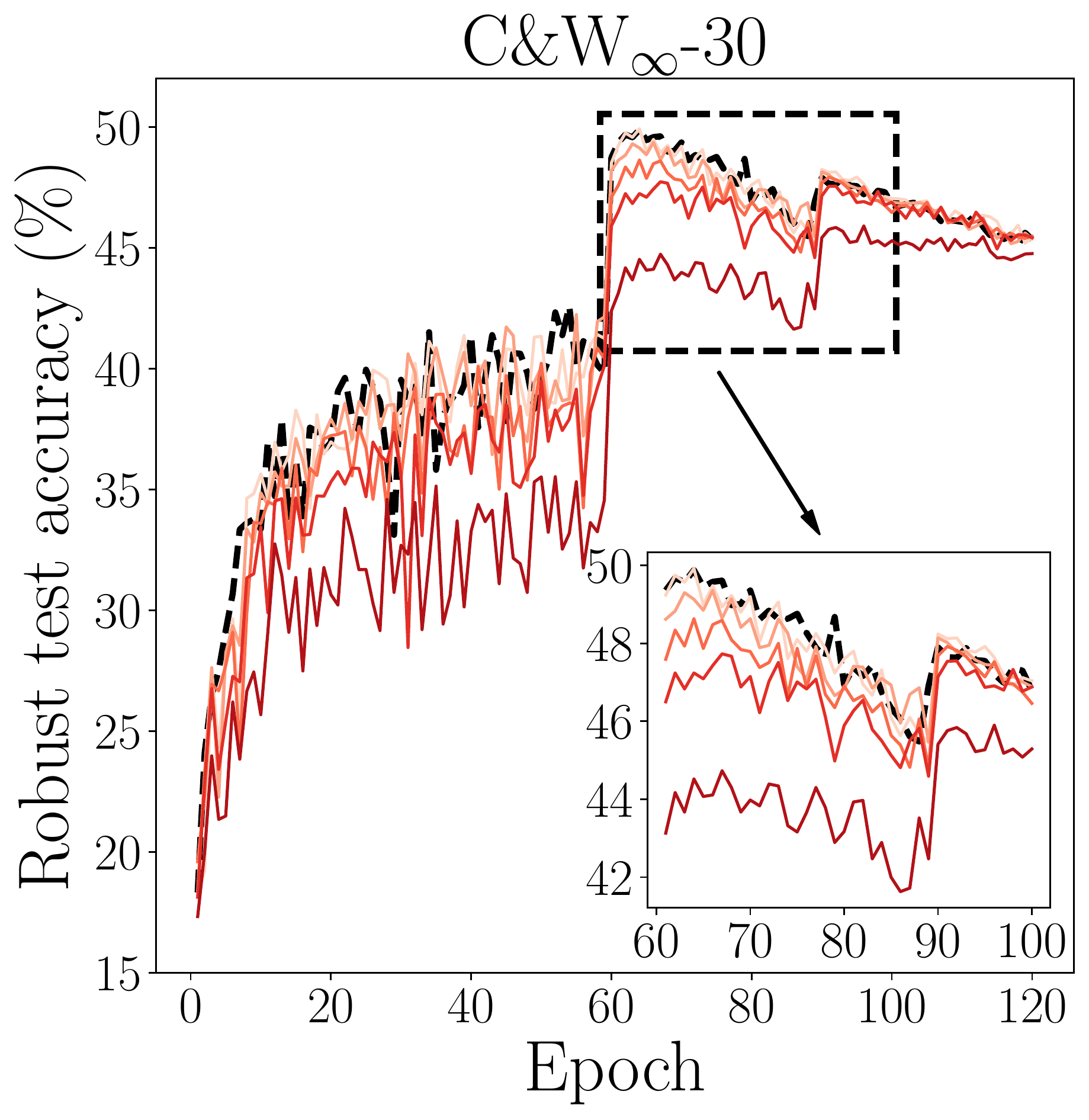}
	\includegraphics[width=0.244\textwidth]{figures/inner_min/r18_eps8_NC/symmetric/N_C_symmetric_eps8_aa_test_acc.pdf}
	\vspace{-0mm}
	\caption{The learning curves of injecting various levels of symmetric-flipping NL in inner maximization. The number in the legend represents noise rate $\eta$.
	}
	\label{Append_fig:inner_NC_symmetric}
	\centering
	\vspace{-0mm}
\end{figure}

\begin{figure}[t!]
	\vspace{-0mm}
	\centering
	\includegraphics[width=0.244\textwidth]{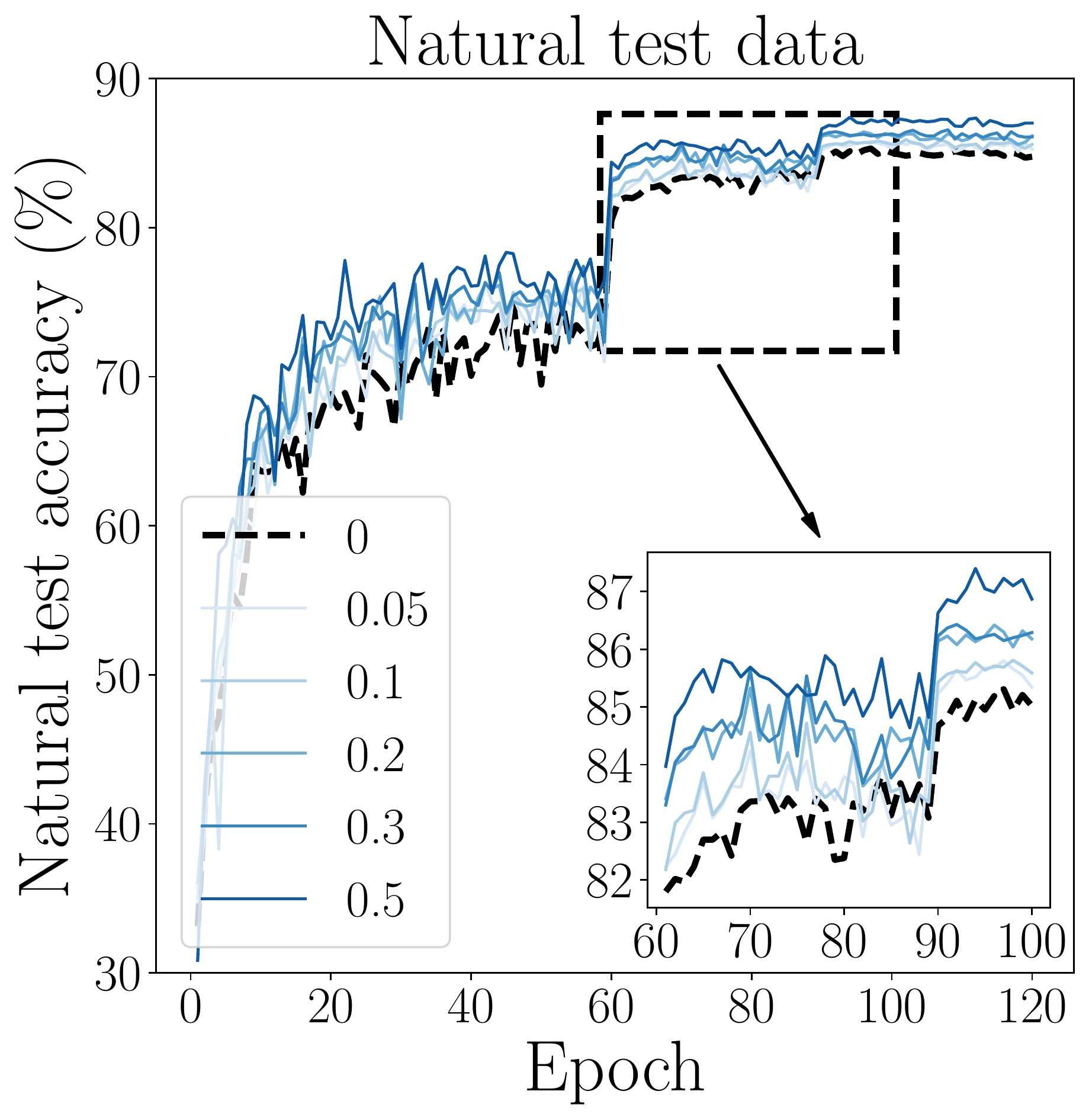}
	\includegraphics[width=0.244\textwidth]{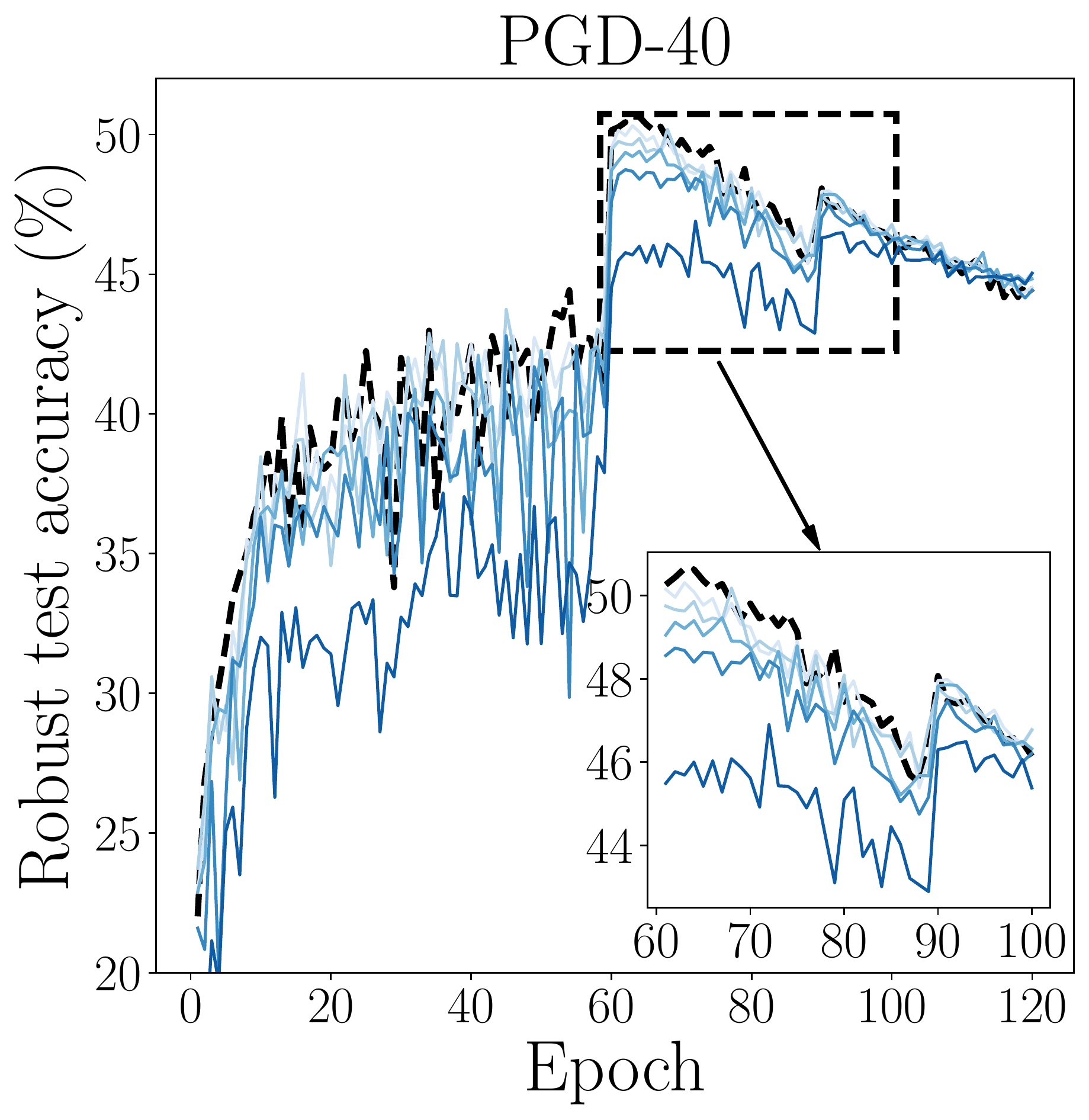} 
	\includegraphics[width=0.244\textwidth]{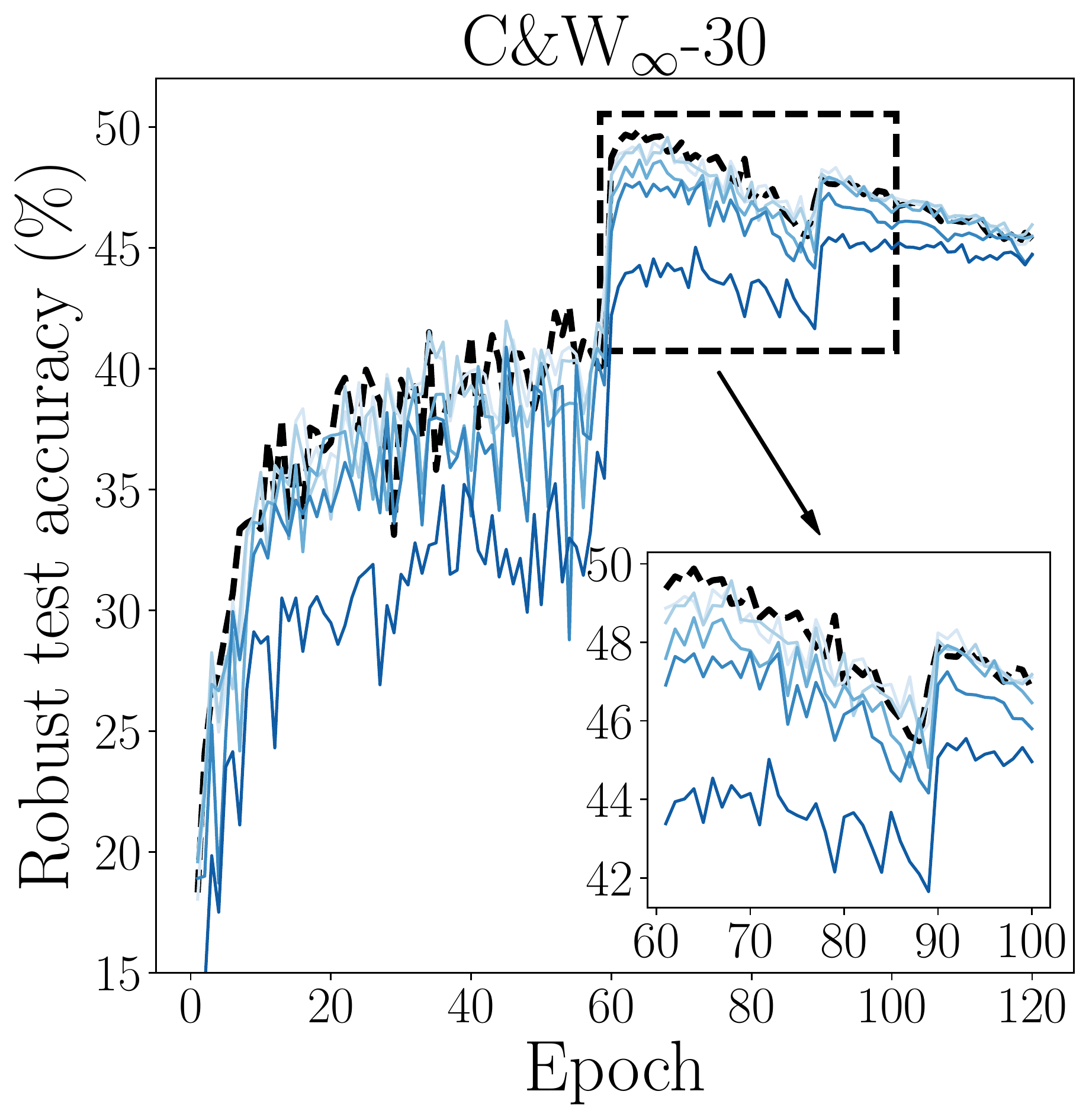}
	\includegraphics[width=0.244\textwidth]{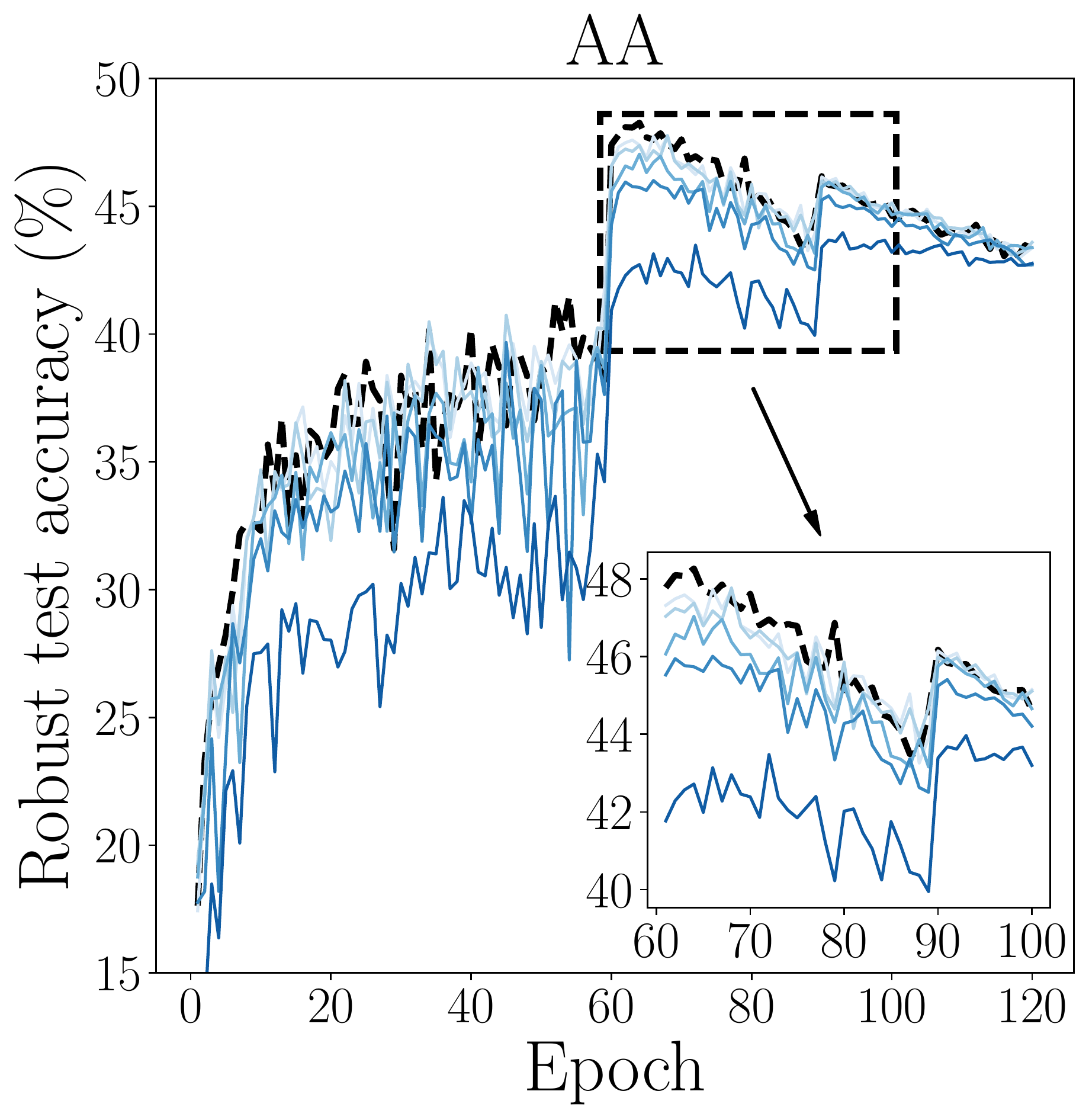}
	\vspace{-0mm}
	\caption{The learning curves of injecting various levels of pair-flipping NL in inner maximization. The number in the legend represents noise rate $\eta$.}
	\label{Append_fig:inner_NC_pairflip}
	\vspace{-0mm}
\end{figure}

Figure~\ref{Append_fig:inner_NC_symmetric} is symmetric-flipping NL injection in inner maximization; Figure~\ref{Append_fig:inner_NC_pairflip} is pair-flipping NL injection in inner maximization. 
We evaluated the robust models based on the four evaluation metrics, i.e., natural test accuracy on natural test data, robust test accuracy on adversarial data generated by PGD-40~\cite{carmon2019unlabeled}, \CW-30 ($L_{\infty}$ version of C$\&$W~\cite{Carlini017_CW} optimized by PGD-30), and AutoAttack~\cite{croce2020reliable} (AA), respectively. The adversarial test data are bounded by $L_{\infty}$ perturbations with $\epsilon_{\mathrm{test}}=8/255$.
Figures~\ref{Append_fig:inner_NC_symmetric} and~\ref{Append_fig:inner_NC_pairflip} consistently show NL injection in inner maximization improves AT's generalization and degrades AT's robustness. 
\begin{figure}[t!]
	\centering
	\includegraphics[width=0.28\textwidth]{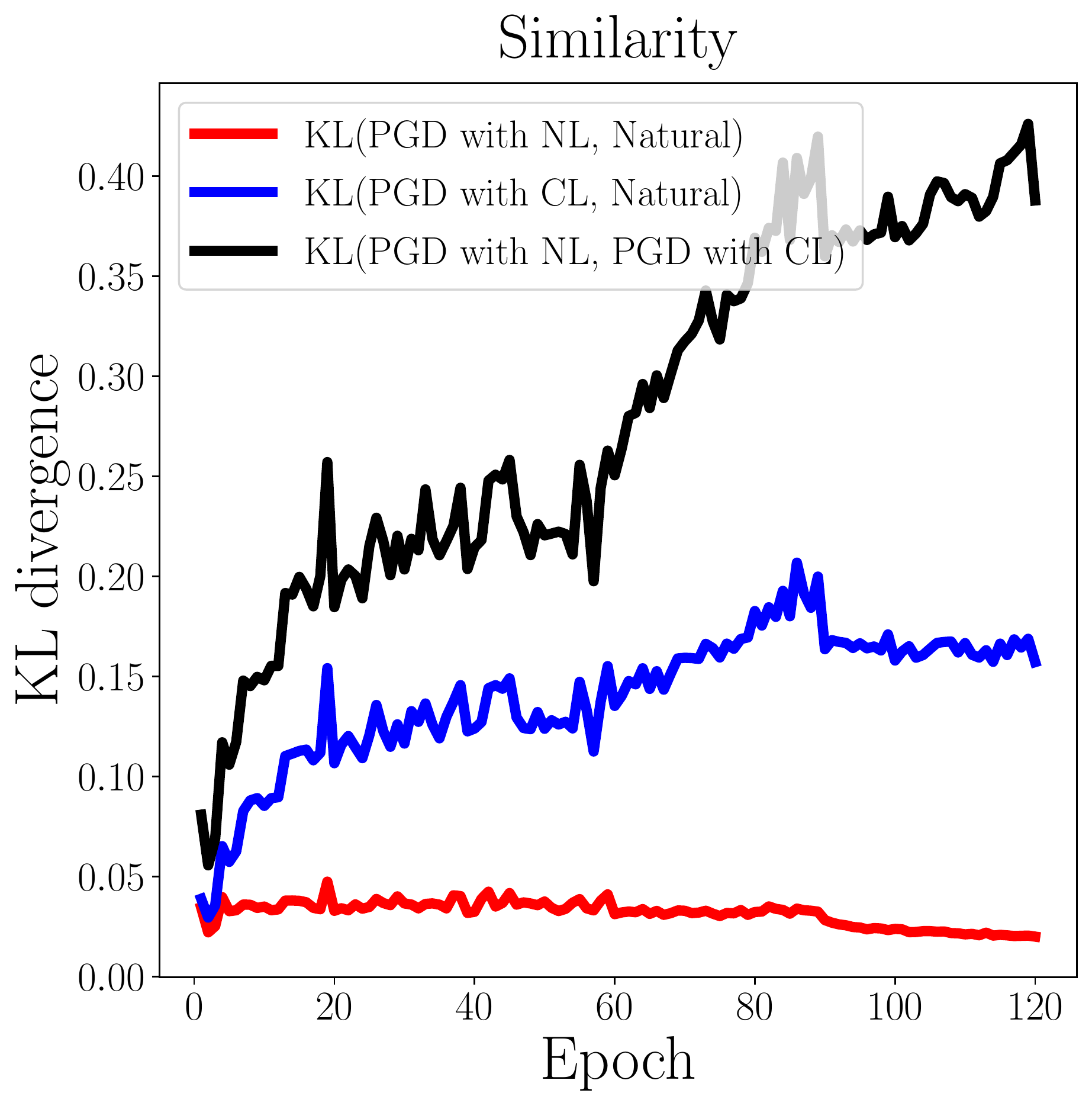}
	\includegraphics[width=0.28\textwidth]{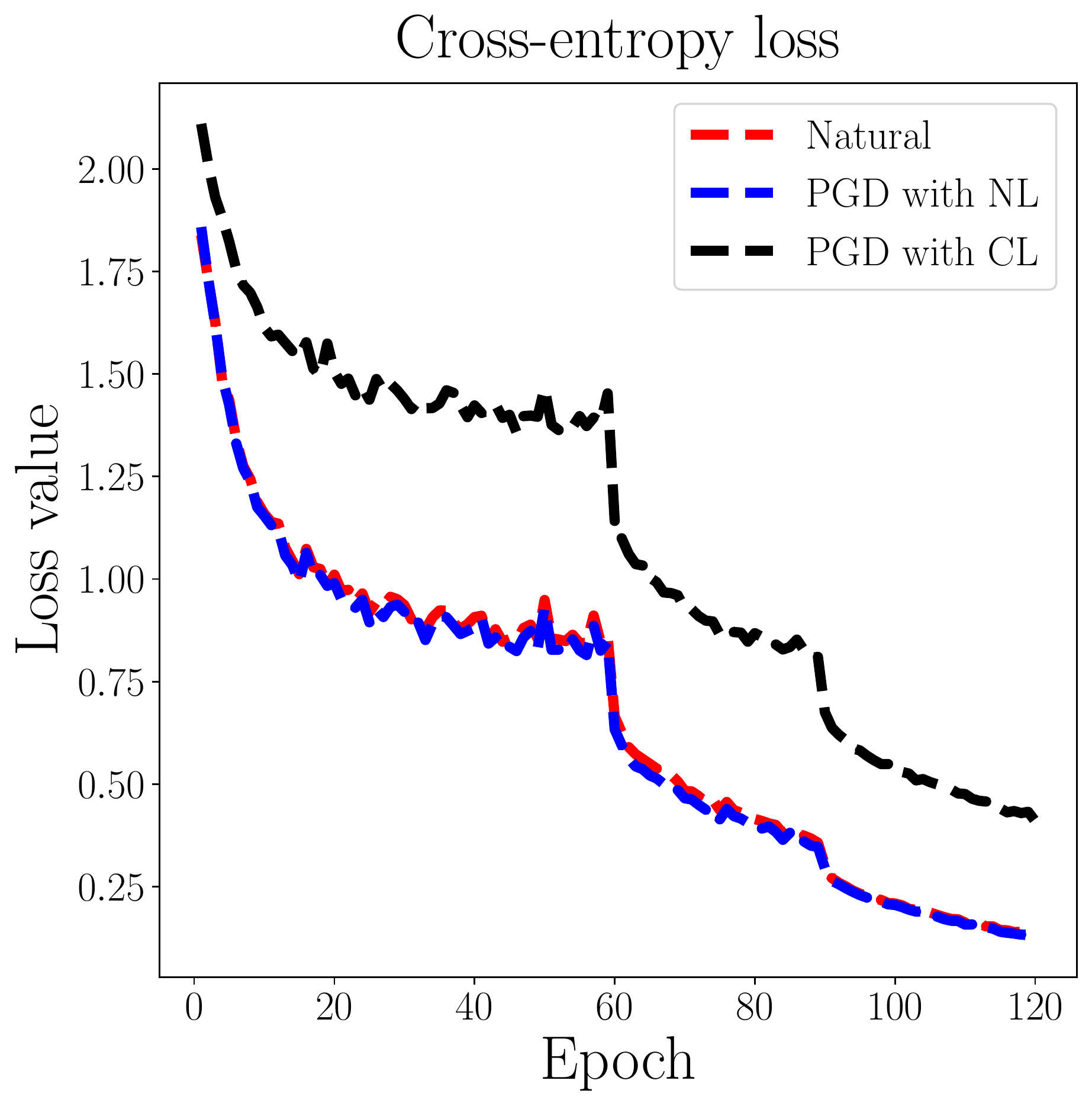}
	\vspace{-0mm}
	\caption{The similarity among PGD with CL (adversarial data generated with correct label), PGD with symmetric-flipping NL (adversarial data generated with noisy label), and Natural data on CIFAR-10 dataset.}
	\label{fig:NL_inner_cifar}
	\vspace{-0mm}
\end{figure}
To further justify the above phenomenon in Figure~\ref{fig:NL_inner_toy_example}, in Figure~\ref{fig:NL_inner_cifar}, we used a real-world dataset---CIFAR-10---to reveal the similarity among ``PGD with NL'' (PGD adversarial data with wrong labels), ``PGD with CL'' (PGD adversarial data with correct labels), and ``Natural'' (natural data). 
We conducted a standard adversarial training for 120 epochs and saved the model's checkpoint at every epoch. 
At every checkpoint, we randomly selected 1000 training data and generated ``PGD with NL'' and ``PGD with CL'' of these data.
We used model's Kullback–Leibler (KL) loss~\cite{Zhang_trades} as the similarity metric (the smaller value means larger similarity). In the left panel of Figure~\ref{fig:NL_inner_cifar}, we compared similarity among ``PGD with NL'', ``PGD with CL'', and ``Natural''. Besides, in the right panel of Figure~\ref{fig:NL_inner_cifar}, we used the correct labels to calculate the cross-entropy loss of ``PGD with NL'', ``PGD with CL'', and ``Natural'', respectively. 

The left panel of Figure~\ref{fig:NL_inner_cifar} shows that the value of KL(PGD with NL, Natural) (red solid line) is apparently lower than that of KL(PGD with CL, Natural) (blue solid line). Besides, the right panel of Figure~\ref{fig:NL_inner_cifar} shows that the cross-entropy loss of ``PGD with NL'' is almost identical to that of ``Natural''. 
Compared with ``PGD with CL'', ``PGD with NL'' is more similar to ``Natural'' (natural data). 
Therefore, label-flipped adversarial data are similar to natural data. 

\section{NL Injection in Outer Minimization}
\label{appendix:outer_NL}
\begin{figure}
	\centering
	\includegraphics[width=0.244\textwidth]{figures/outer_max/r18_eps8_CN/C_N_symmetric_eps8_natural_test_acc.pdf}
	\includegraphics[width=0.244\textwidth]{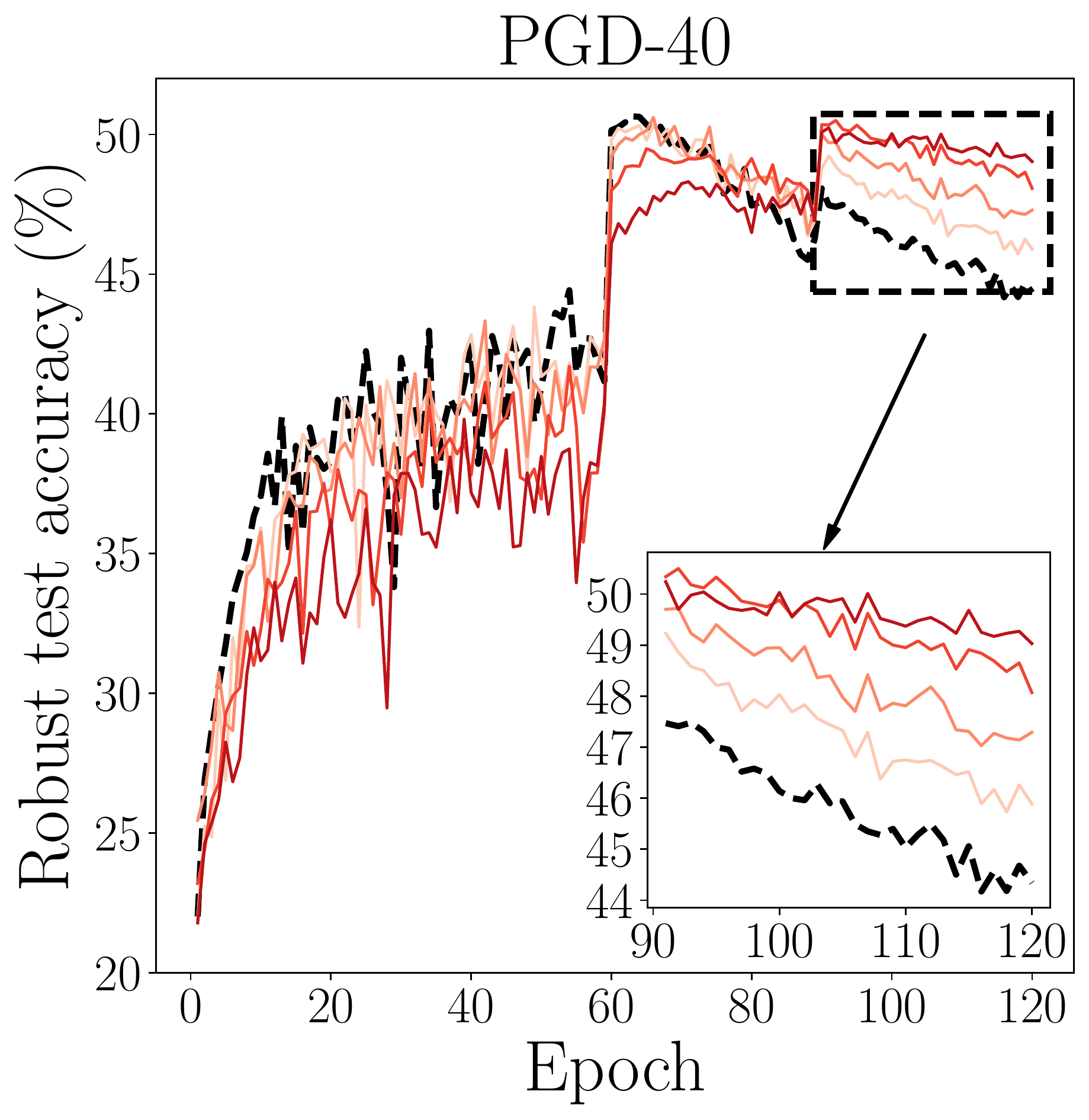}
	\includegraphics[width=0.244\textwidth]{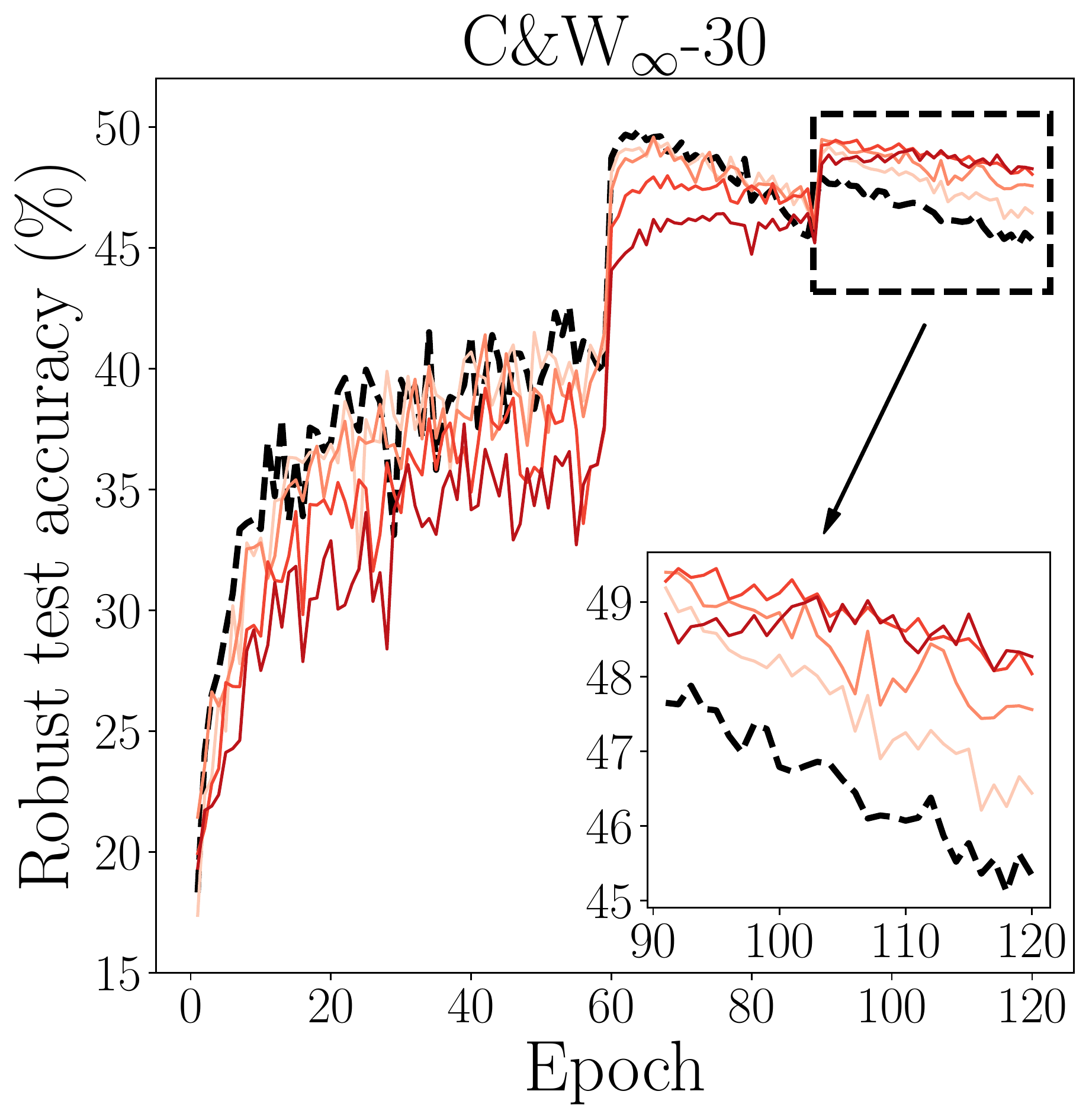}
	\includegraphics[width=0.244\textwidth]{figures/outer_max/r18_eps8_CN/C_N_symmetric_eps8_aa_test_acc.pdf}
	\caption{The learning curves of injecting various levels of symmetric-flipping NL in outer minimization. The number in the legend represents noise rate $\eta$}
	\label{Append_fig:symmetric-flipping_outer_minimization}
\end{figure}

\begin{figure}
	\centering
	\includegraphics[width=0.244\textwidth]{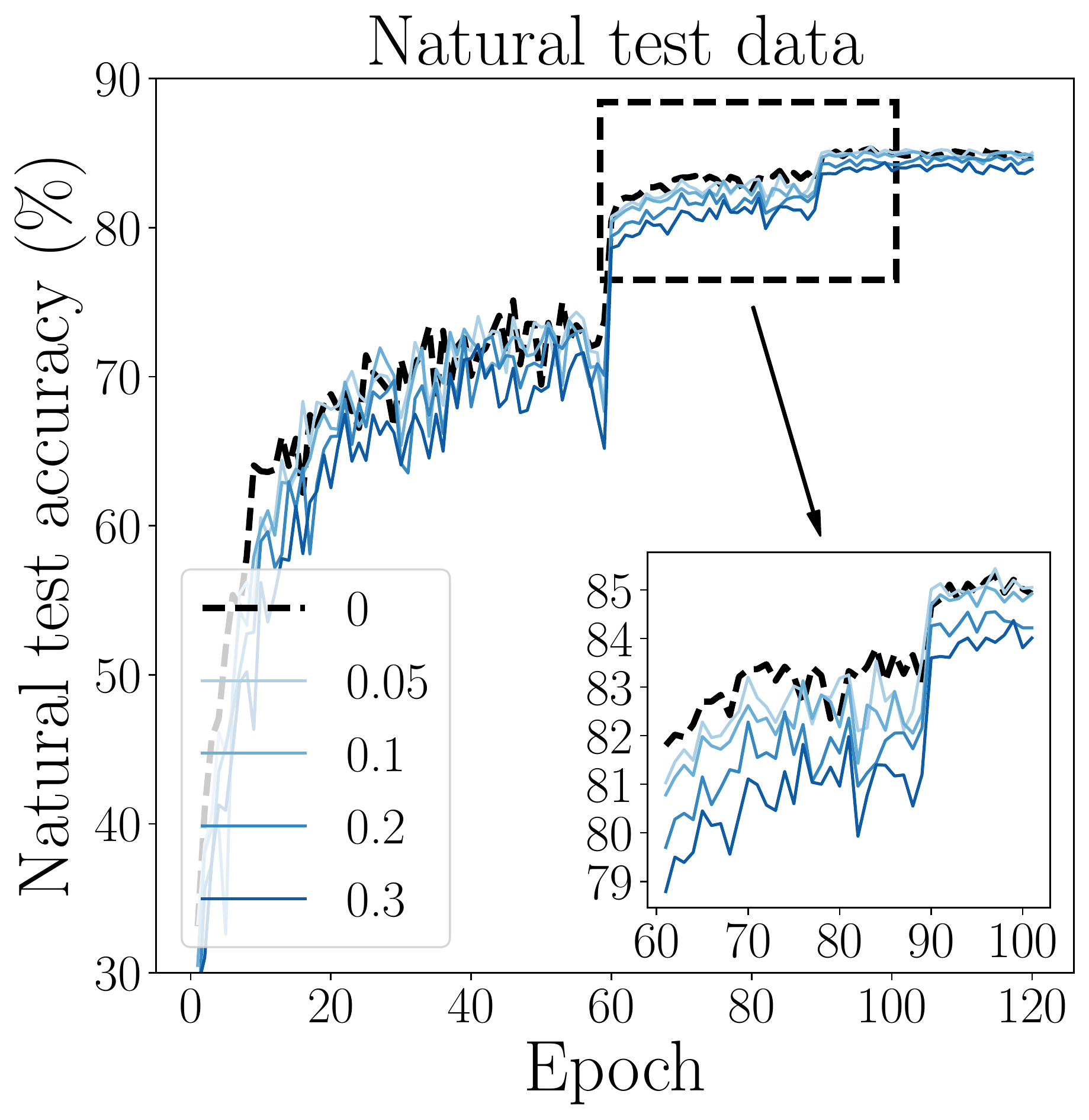}
	\includegraphics[width=0.244\textwidth]{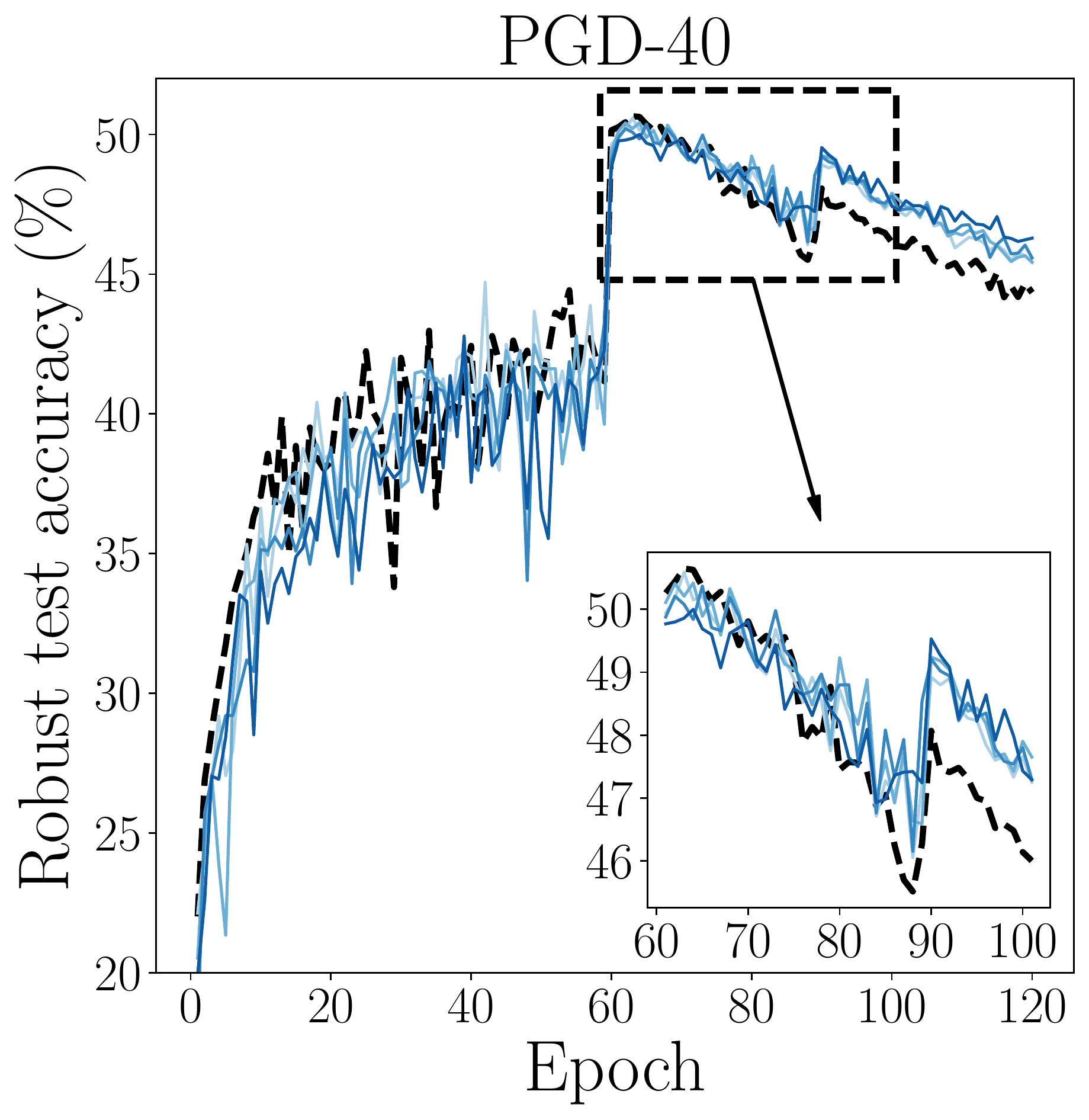}
	\includegraphics[width=0.244\textwidth]{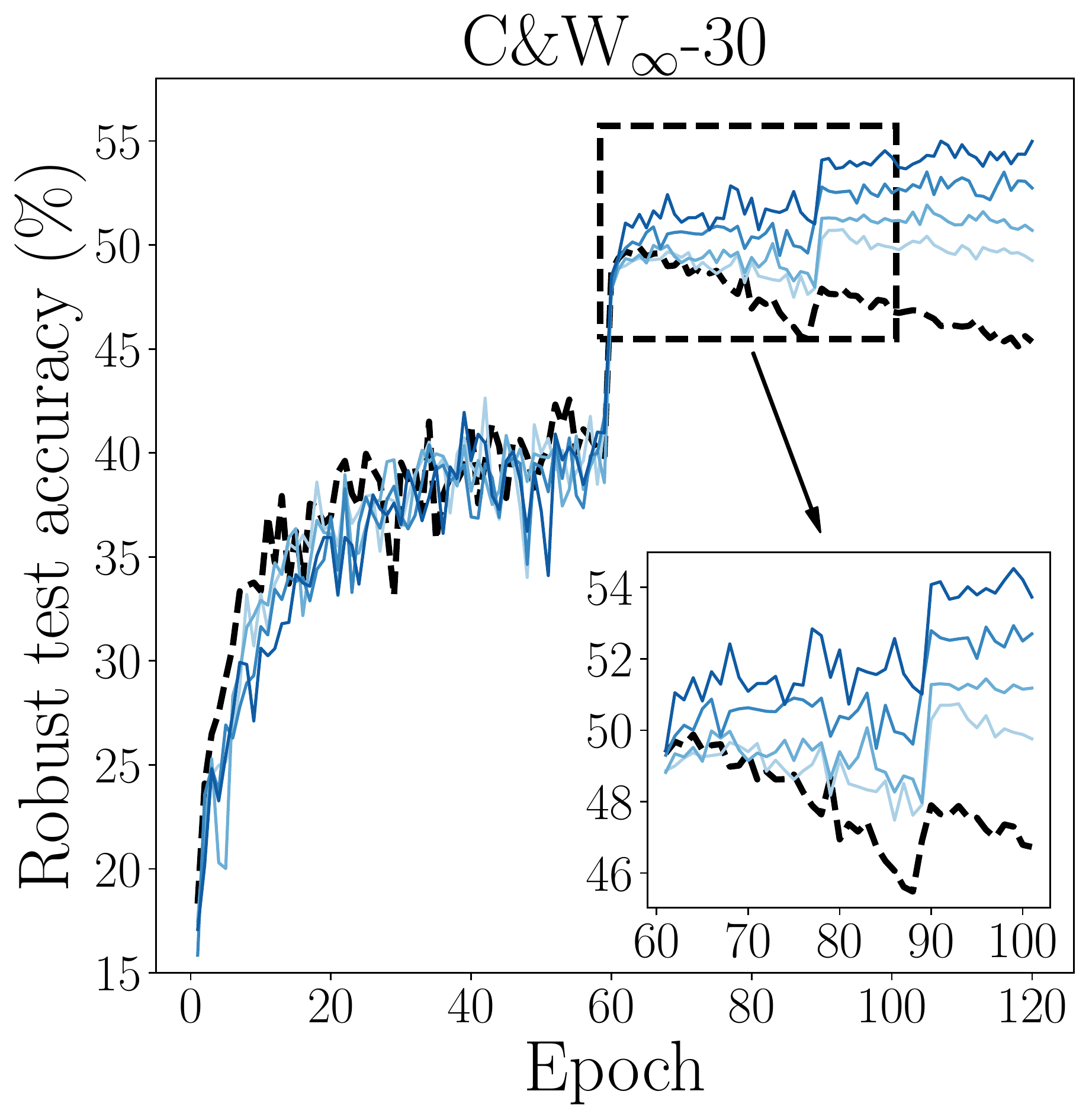}
	\includegraphics[width=0.244\textwidth]{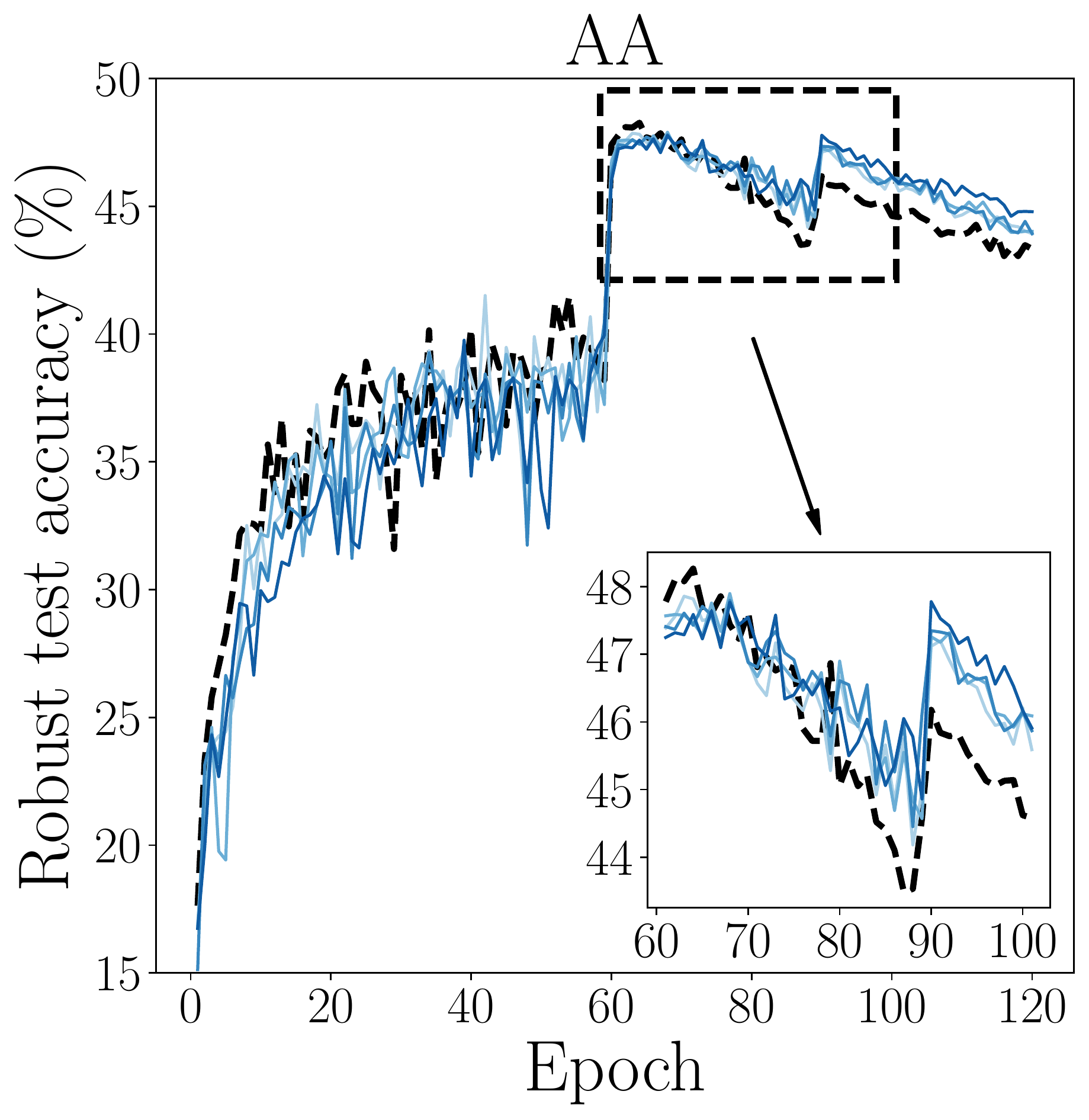}
	\caption{The learning curves of injecting various levels of pair-flipping NL in outer minimization. The number in the legend represents noise rate $\eta$}
	\label{Append_fig:pair-flipping_outer_minimization}
\end{figure}

Figure~\ref{Append_fig:symmetric-flipping_outer_minimization} is symmetric-flipping NL injection in outer minimization; Figure~\ref{Append_fig:pair-flipping_outer_minimization} is pair-flipping NL injection in outer minimization. The evaluation metrics keep the same as Appendix~\ref{appendix:inner_NL}. 

\begin{figure}[tp!]
	\vspace{-0mm}
	\centering
	\includegraphics[width=0.428\textwidth]{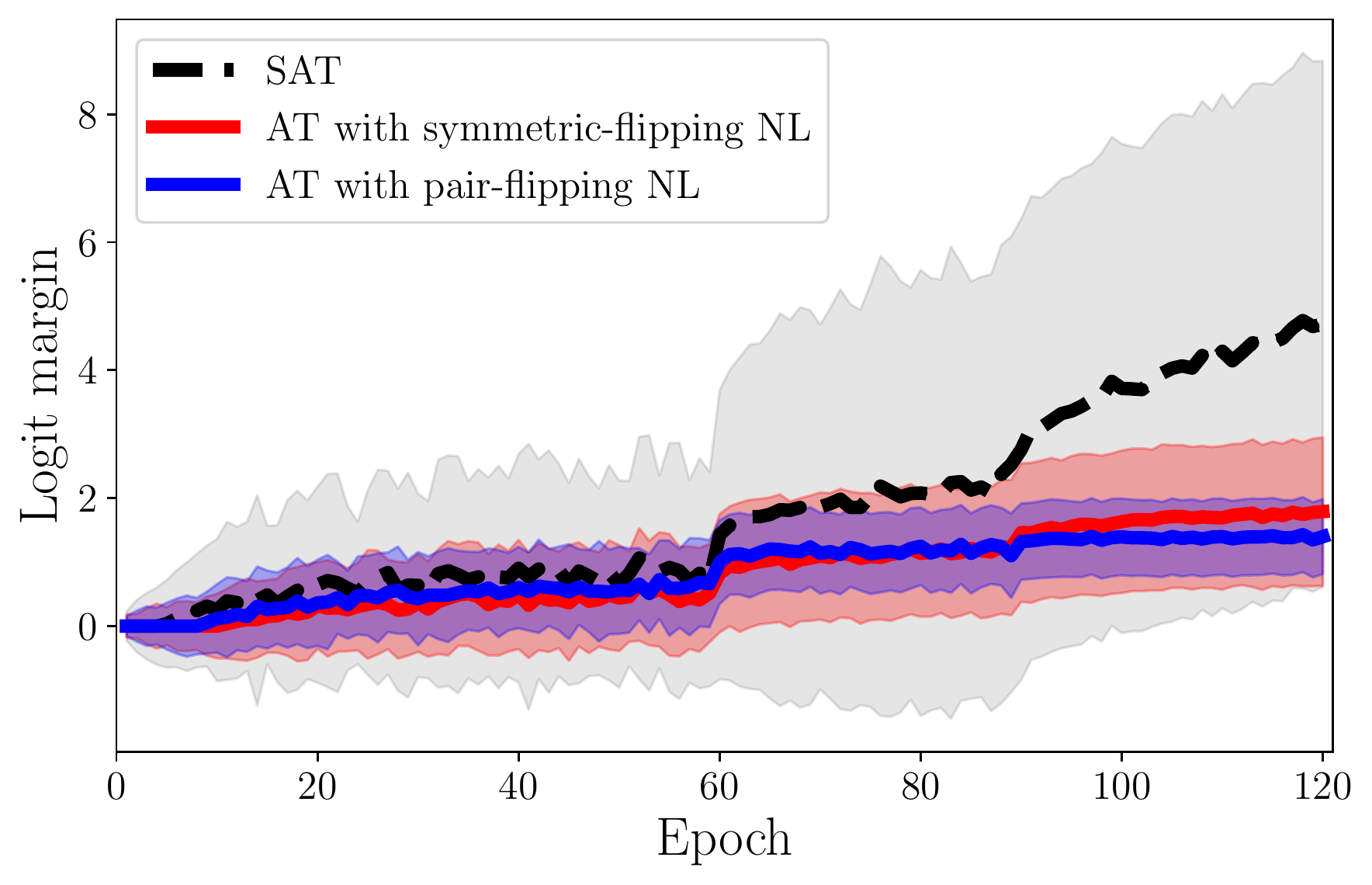}
	\caption{We compare the median \textit{logit margin} of natural test data and its standard deviation over the training process of SAT (black dashed line), that of AT with symmetric-flipping NL in outer minimization (red line), and that of AT with pair-flipping NL in outer minimization (blue line). 
		We find NL in outer minimization reduces logit margins, thus obfuscating gradients of the C$\&$W attack~\cite{Carlini017_CW}.}
	\label{fig:outer_NL_margin}
\end{figure}

From the third panel of Figure~\ref{Append_fig:pair-flipping_outer_minimization}, we observe that NL in outer minimization can significantly improve the model's robustness against the C$\&$W attack, especially injecting pair-flipping label noise. 

To figure out the reason for this interesting phenomenon, at every training epoch, we illustrated the median \textit{logit margin} $\big(f_{\theta}^{y}(x) - \max_{j \neq y}f_{\theta}^{j}(x)\big)$ over all natural test data and its standard deviation in Figure~\ref{fig:outer_NL_margin}. We compare SAT (black dashed line) with AT with NL (the red line for injecting symmetric noise and the blue line for injecting pair flipping noise). 
As shown in Figure~\ref{fig:outer_NL_margin}, injecting NL in outer minimization leads to a small logit margin, especially injecting pair-flipping NL (the blue line).

Note that the implementation of the C$\&$W attack follows 
\begin{equation}
	\label{Eq:cw_inner_maximization}
	\xadv = {\arg\max}_{\xadv \in \epsball[\bx]} \big(\max( \max_{j \neq y}f^{j}_{\theta}(\xadv) - f^{y}_{\theta}(\xadv) - \kappa, 0)\big),
\end{equation} 
where $\kappa$ is a positive constant value that aids the optimization. 
When the logit margin is small and close to 0, the starting point of optimization of the inner loss of the C$\&$W attack (i.e., $(\max( \max_{j \neq y}f^{j}_{\theta}(\xadv) - f^{y}_{\theta}(\xadv) - \kappa, 0))$ in Eq.~\eqref{Eq:cw_inner_maximization}) is also small and close to 0. This will incur the gradient vanishing problem and hurdle the optimization for finding the C$\&$W-adversarial data. Therefore, NL in outer minimization obfuscates gradients of C$\&$W attacks by reducing logit margins. 

\section{Injecting NL in Both Inner Maximization and Outer Minimization}
\label{appendix:NN_fixed}

\begin{figure}[t!]
	\centering
	\includegraphics[width=0.244\textwidth]{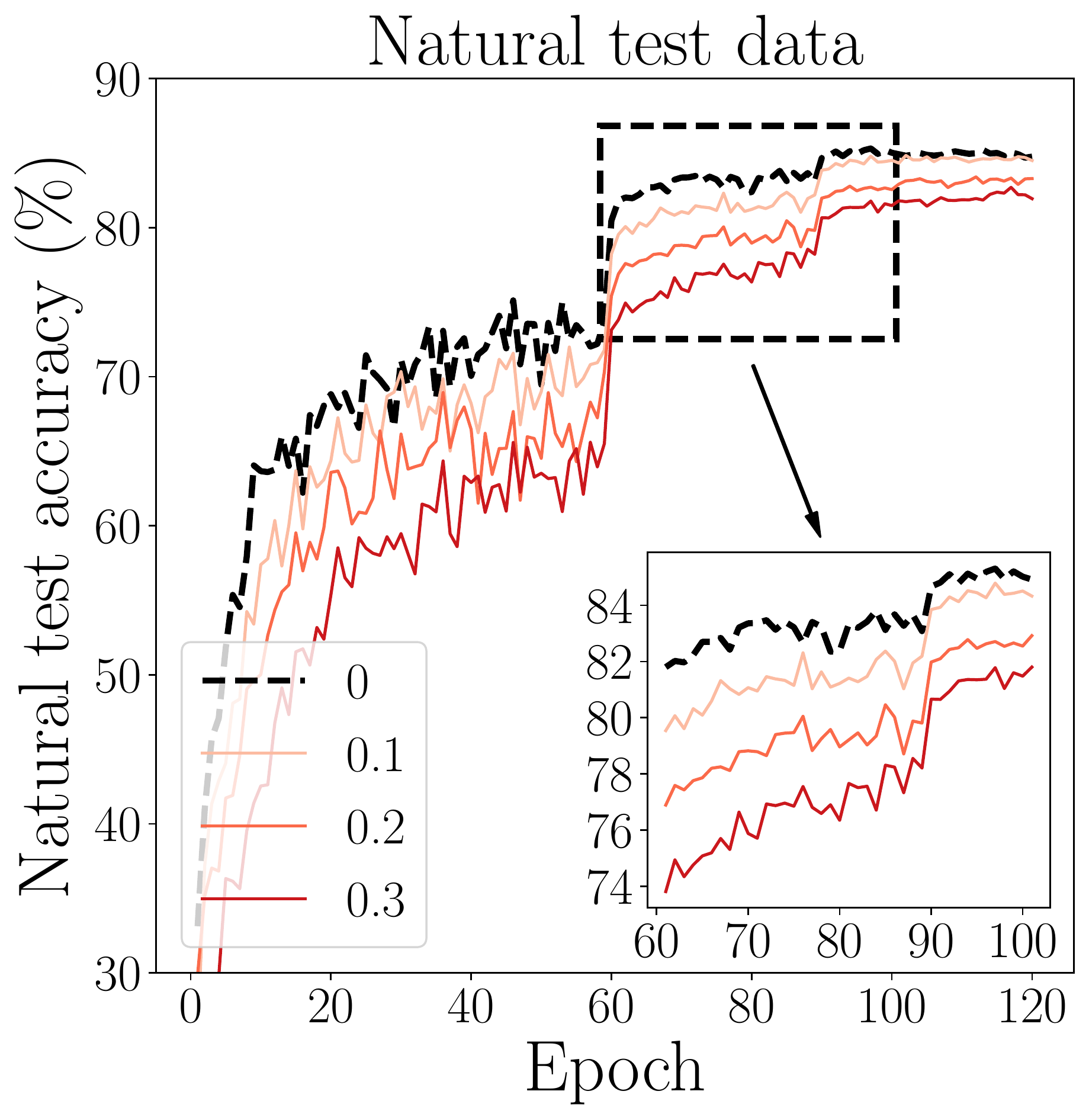}
	\includegraphics[width=0.244\textwidth]{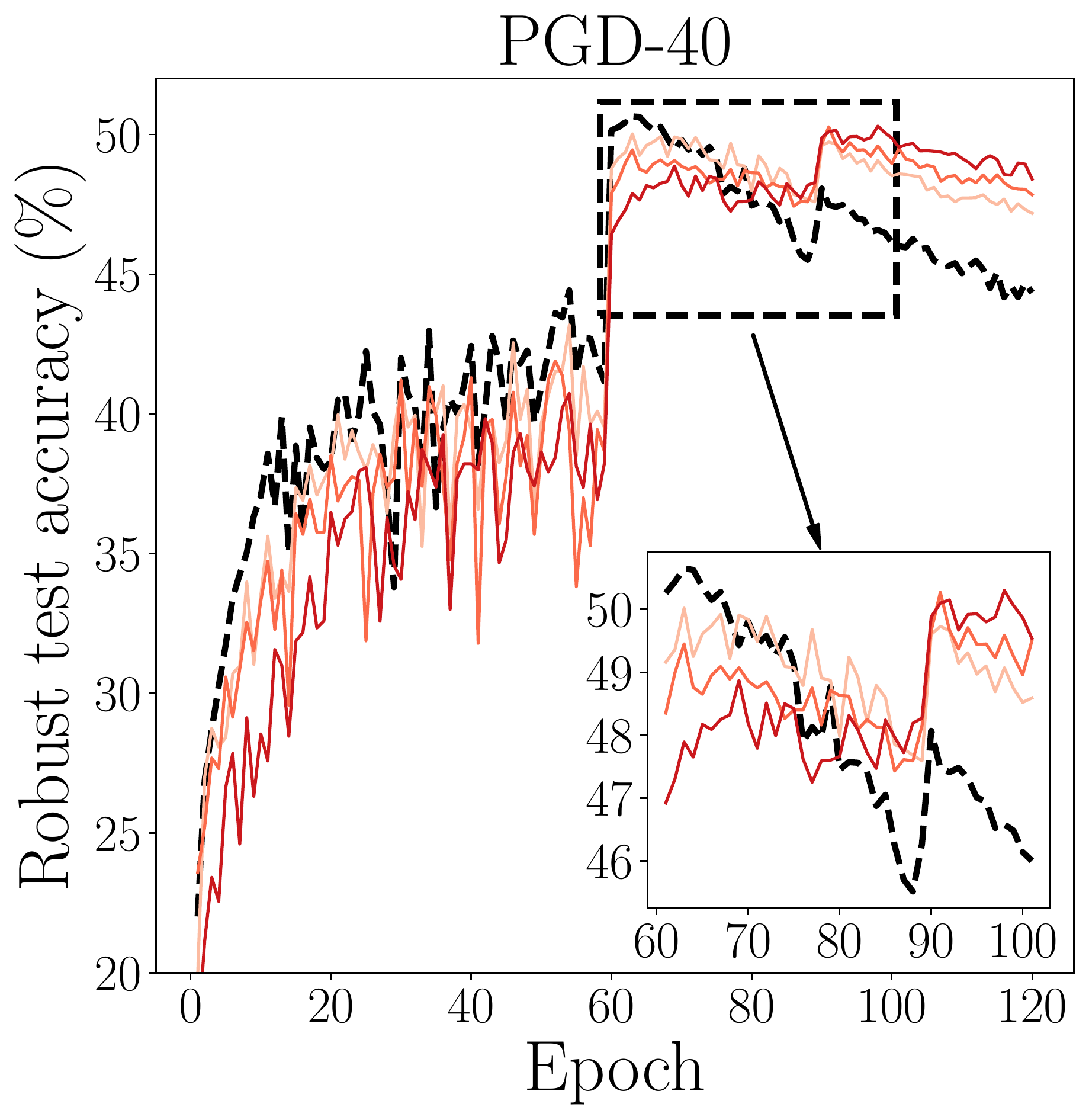}
	\includegraphics[width=0.244\textwidth]{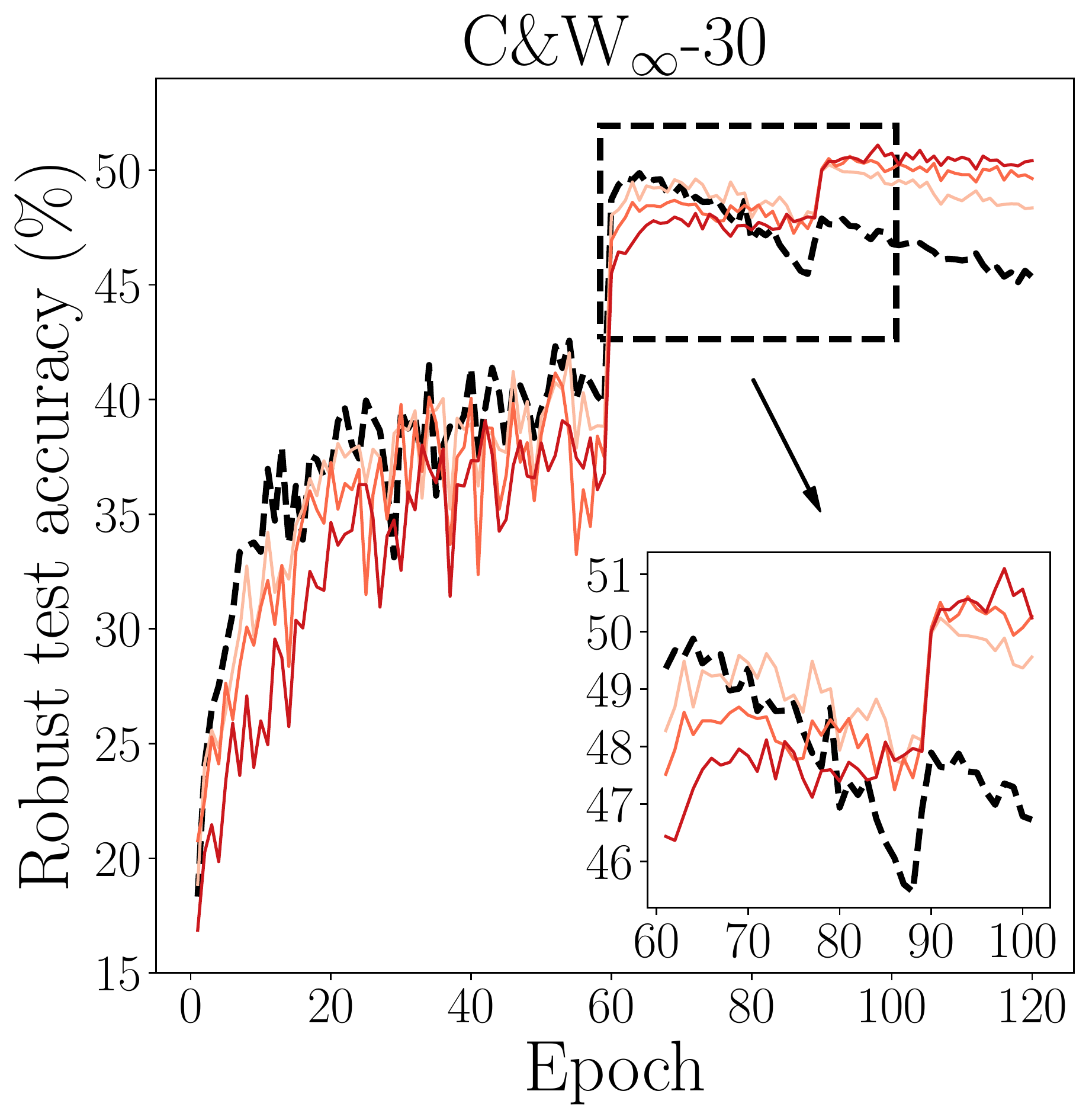}
	\includegraphics[width=0.244\textwidth]{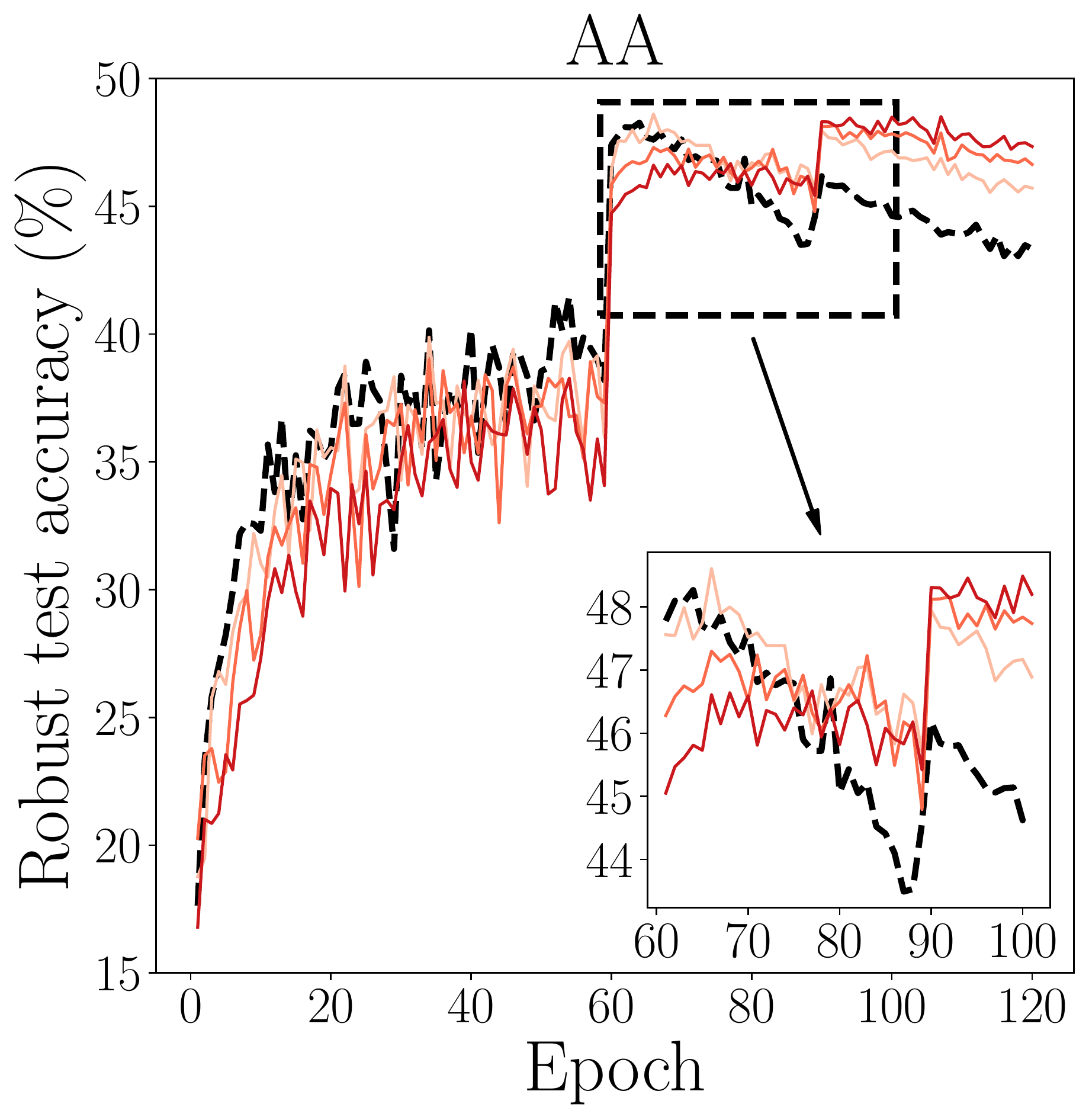}
	\caption{The learning curves of injecting symmetric-flipping NL into both inner maximization and outer minimization. The number in the legend represents noise rate $\eta$.}
	\label{fig:AT_NN_symmetric}
\end{figure}
\begin{figure}[t!]
	\centering
	\includegraphics[width=0.244\textwidth]{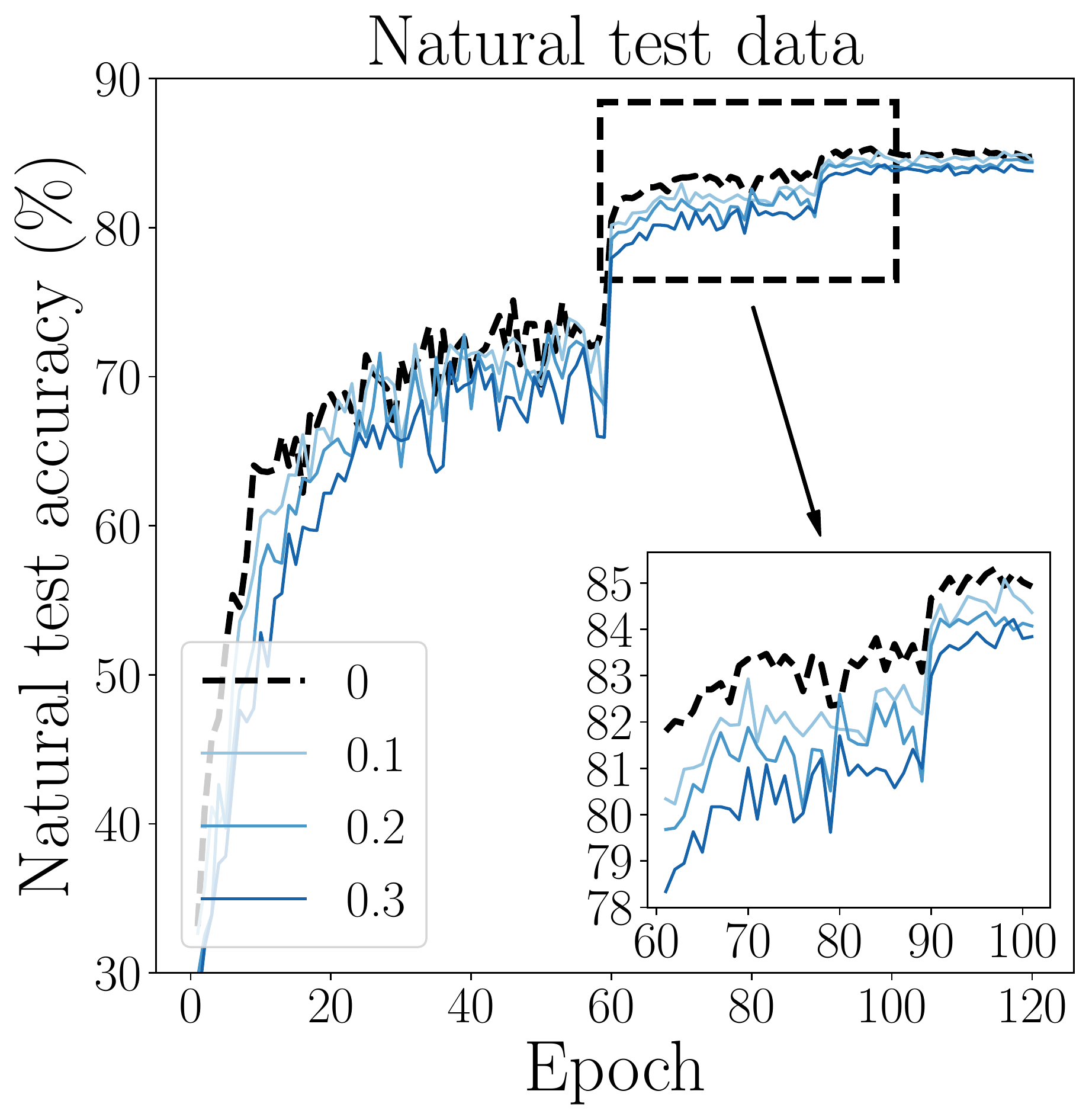}
	\includegraphics[width=0.244\textwidth]{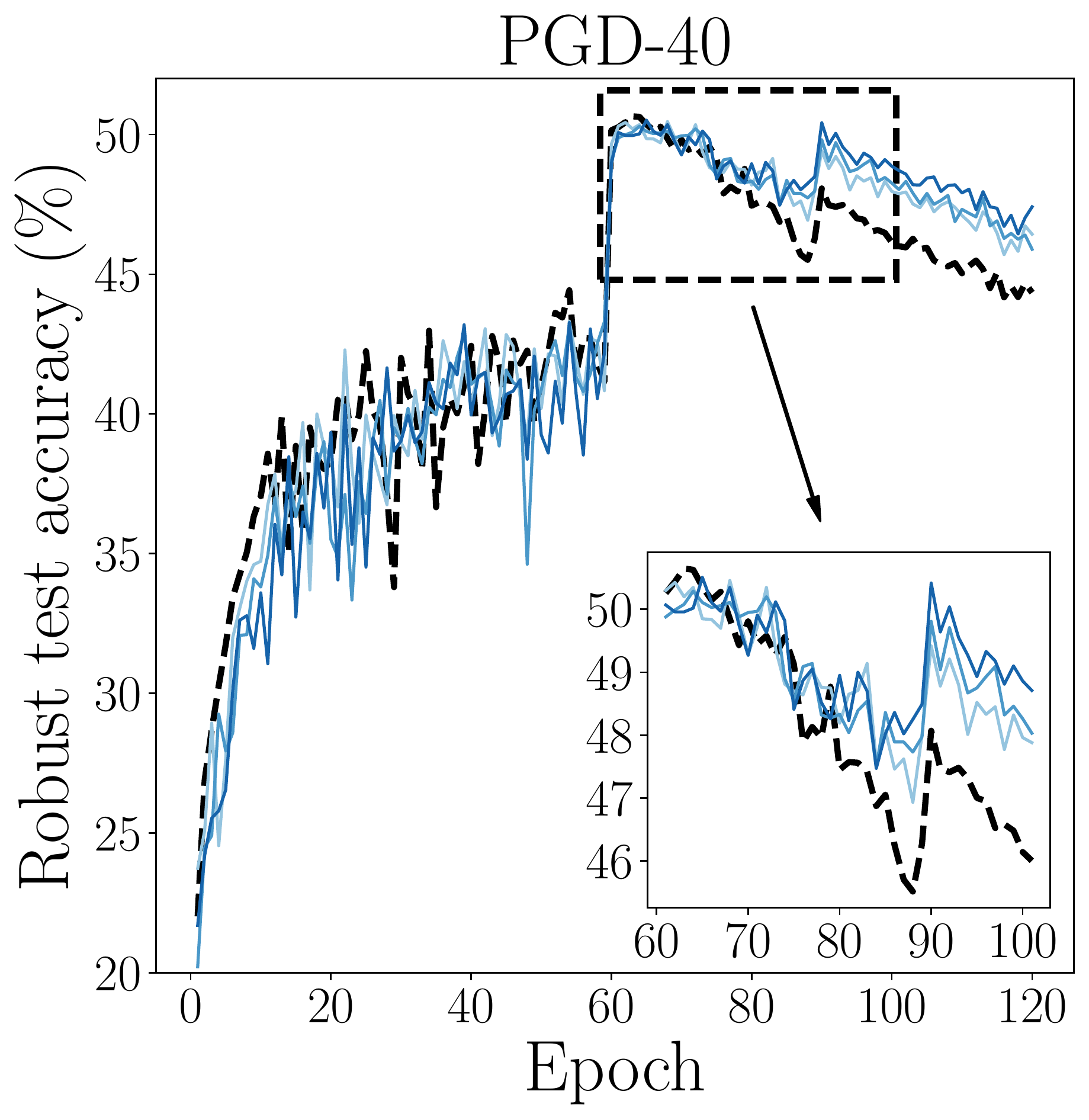}
	\includegraphics[width=0.244\textwidth]{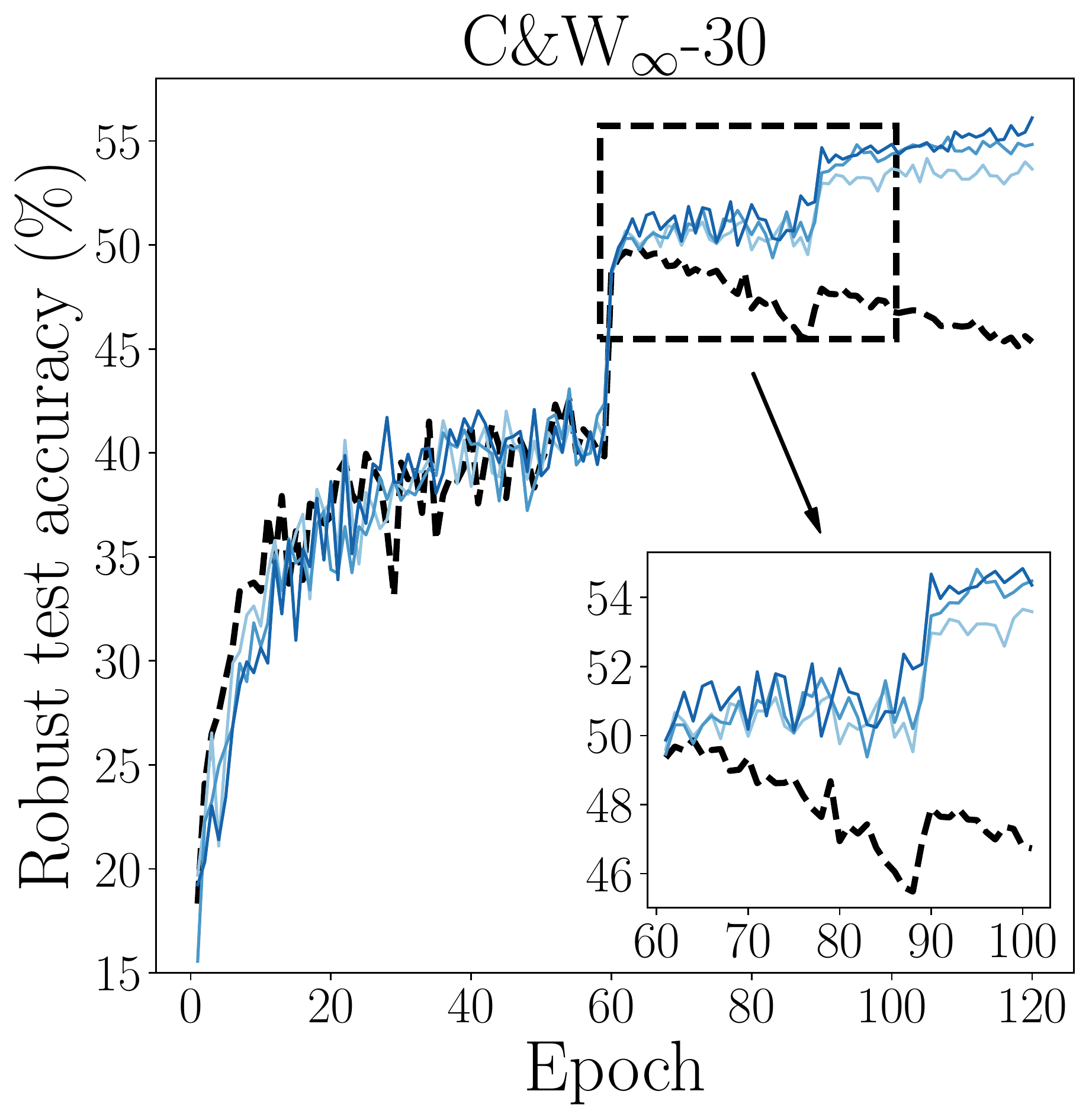}
	\includegraphics[width=0.244\textwidth]{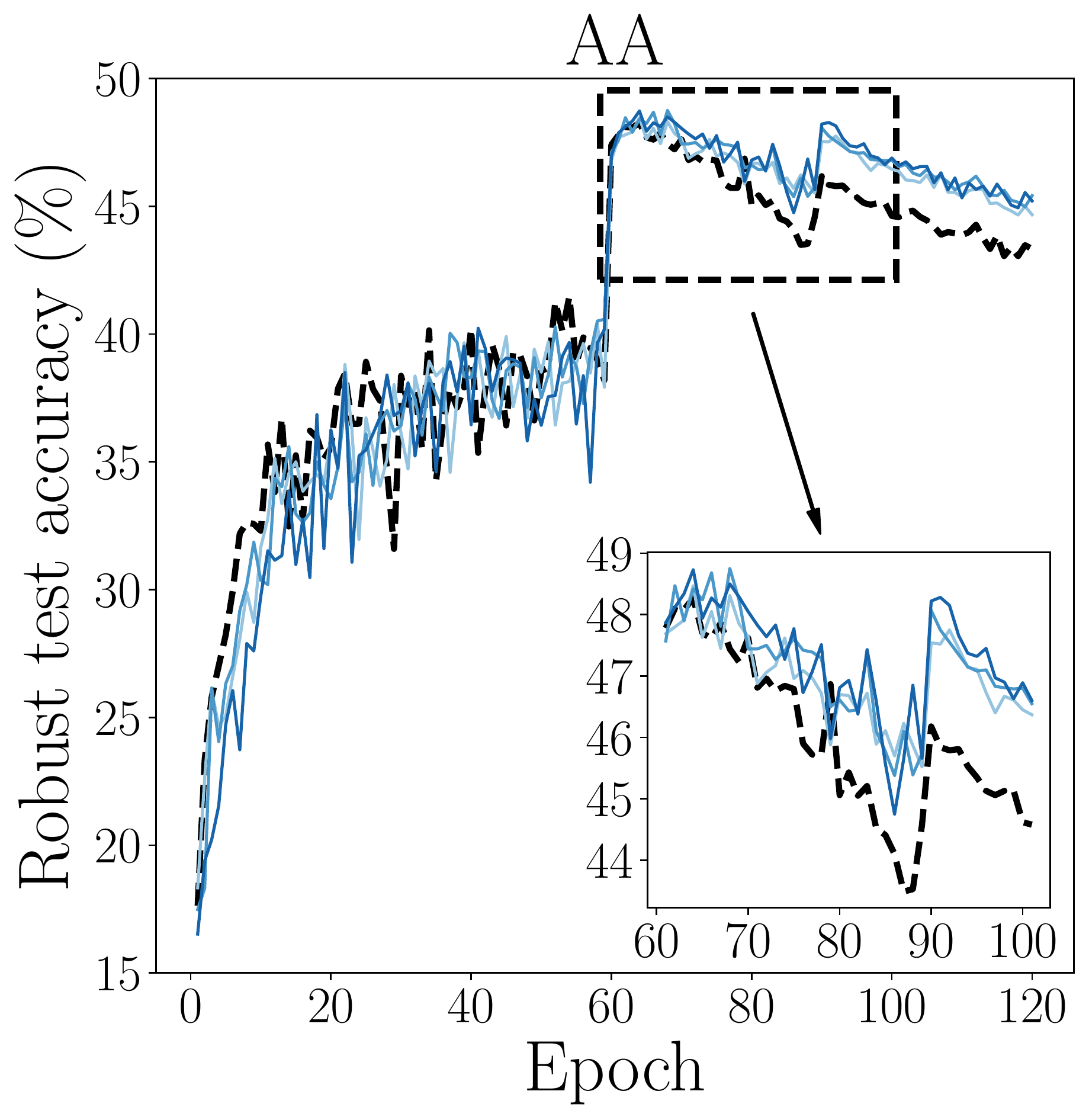}
	\caption{The learning curves of injecting pair-flipping NL into both inner maximization and outer minimization. The number in the legend represents noise rate $\eta$.}
	\label{fig:AT_NN_pairflip}
\end{figure}

\paragraph{Observation (iii)} We report the learning curves of ResNet-18 trained by AT with NL injection in both inner maximization and outer minimization on CIFAR-10 dataset in Figure~\ref{fig:AT_NN_symmetric} (symmetric-flipping NL) and Figure~\ref{fig:AT_NN_pairflip} (pair-flipping NL). The noise rate $\eta$ is chosen from $\{ 0.1, 0.2, 0.3 \}$. The detailed training settings (e.g., the optimizer and learning schedule) are the same as Section~\ref{sec:NL_inner_experiment}.

Figures~\ref{fig:AT_NN_symmetric} and~\ref{fig:AT_NN_pairflip} show that, with the increasing of $\eta$ (the color gradually becomes deeper), the natural test accuracy (the leftmost panels) on natural test data decreases; while the robust test accuracy (the right three panels) on adversarial data evaluated at the last checkpoint and at the best checkpoint increases simultaneously. For example, when $\eta$ equals to 0.1 (blue lines with different shades) in Figure~\ref{fig:AT_NN_symmetric}, the natural test accuracy at Epoch 120 is comparable with SAT (black dashed line), and the robust test accuracy at Epoch 120 is obviously above the black dashed line. Such observation manifests that injecting NL in both inner maximization and outer minimization can alleviate robust overfitting and even improve the best-checkpoint robustness. 

\begin{figure}[t!]
	\centering
	\includegraphics[width=0.244\textwidth]{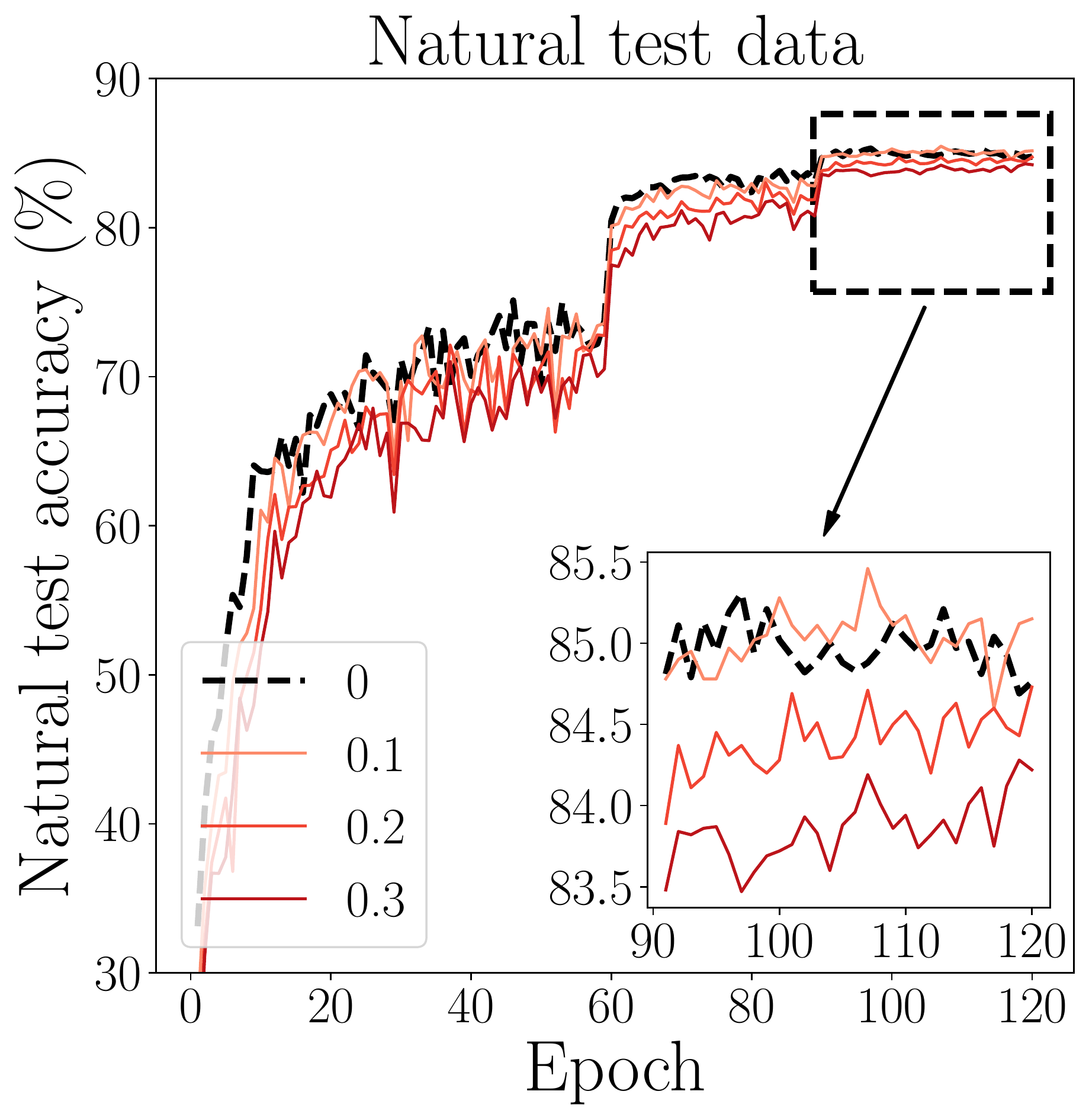}
	\includegraphics[width=0.244\textwidth]{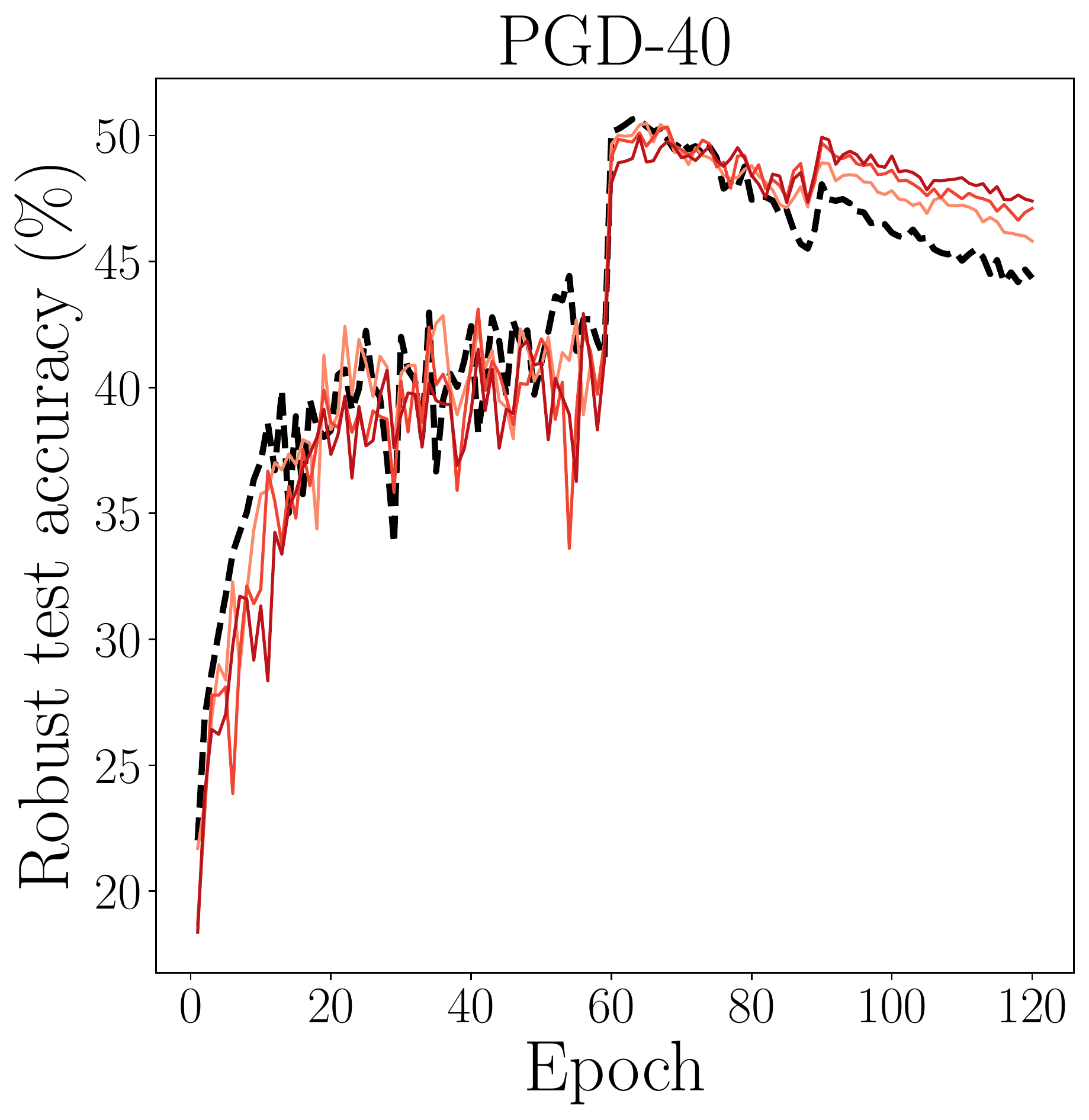}
	\includegraphics[width=0.244\textwidth]{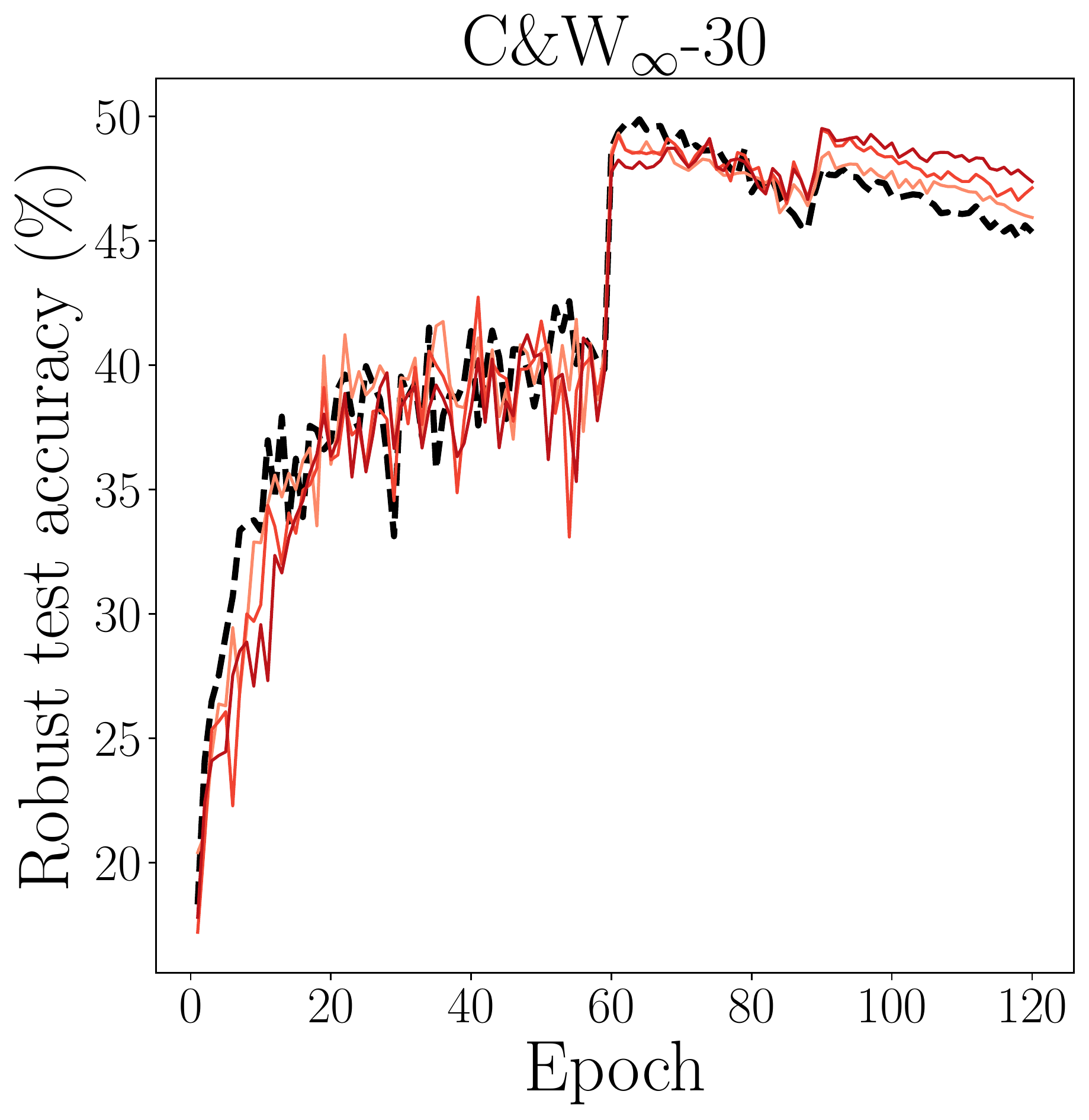}
	\includegraphics[width=0.244\textwidth]{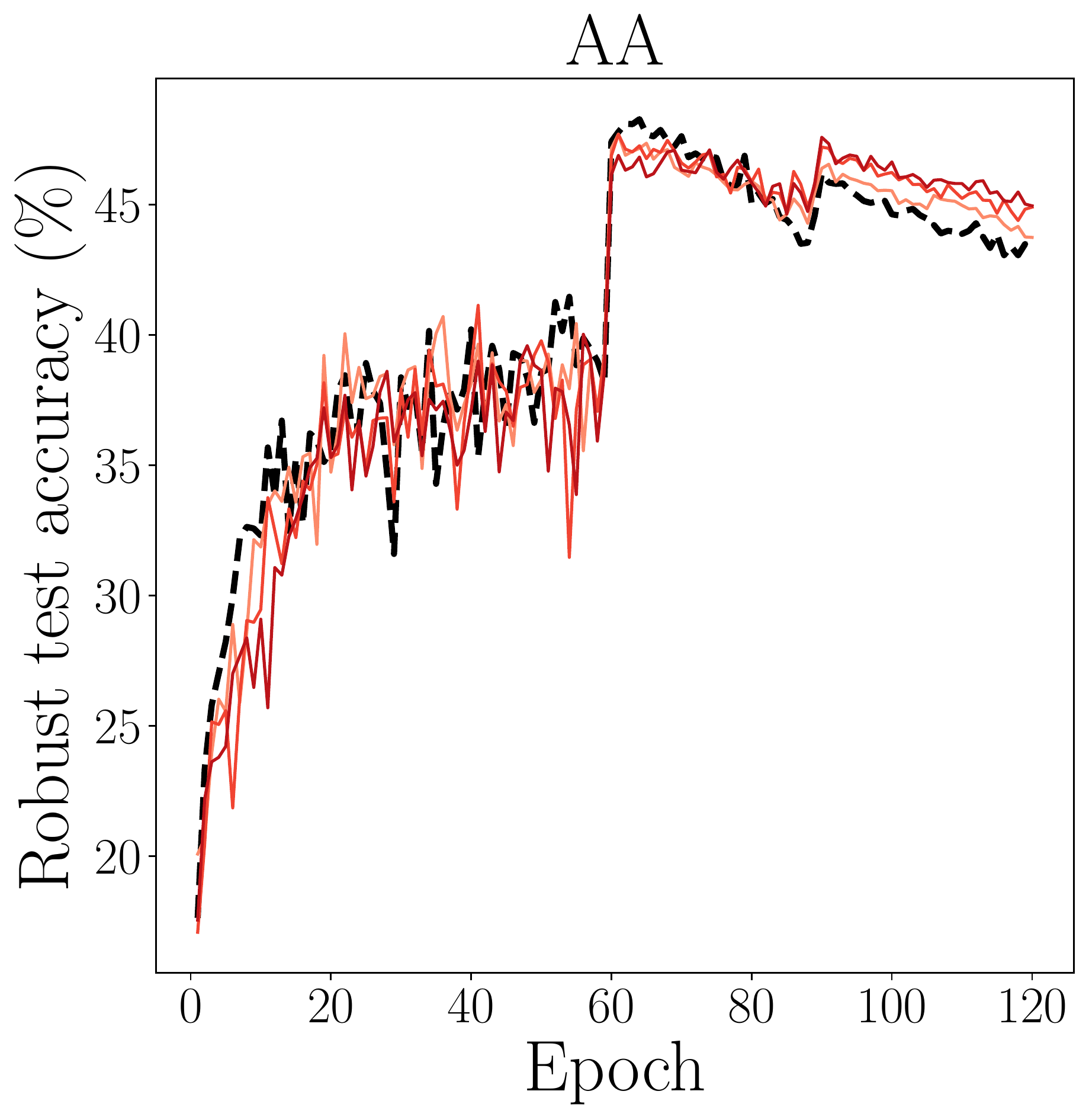}
	\caption{The learning curves of ResNet-18 trained by AT with different NL injection in inner maximization and outer minimization. The number in the legend represents noise rate $\eta$.}
	\label{fig:AT_NN_symmetric_diff}
\end{figure}

\paragraph{Observation (iv)}
Figure~\ref{fig:AT_NN_symmetric_diff} demonstrates the learning curves of ResNet-18 trained by AT with different NL injection in inner maximization and outer minimization. In details, at every minibatch, we choose a portion of data to flip labels in inner maximization and another portion of data to flip labels in outer minimization. In Figure~\ref{fig:AT_NN_symmetric_diff}, NL is generated by symmetric flipping. The noise rate $\eta$ keeps same in both inner maximization and outer minimization, and $\eta \in \{0.1, 0.2, 0.3 \}$. The detailed training settings (e.g., the optimizer and learning schedule) are the same as Section~\ref{sec:NL_inner_experiment}. Figure~\ref{fig:AT_NN_symmetric_diff} shows that adversarial robustness is slightly degraded, which indicates that label-flipped adversarial data no longer serve the regularization purpose and are no longer adversarial as well to enhance robustness.

\section{Extensive Experimental Details}
\label{appendix:NN_pair}

\paragraph{Computing resources.} 
All experiments were conducted on a machine with Intel Xeon Gold 5218 CPU, 250GB RAM and six NVIDIA Tesla V100 SXM2 GPU, and three machines with Intel Xeon Silver 4214 CPU, 128GB RAM and four GeForce RTX 2080 Ti GPU.

\paragraph{Selection criterion for the best checkpoint.}
We selected the best checkpoint based on robust test accuracy on AA~\cite{croce2020reliable} adversarial data when using ResNet-18. When using Wide ResNet, we use PGD-10 adversarial data for the consideration of speed. 
Note that the best checkpoints appear in Figure~\ref{fig:AT_NN_diff_wd}, Figure~\ref{fig:AT_NN_diff_eps}, Figure~\ref{fig:diff_noisy_dataset}, and all tables.

\begin{figure*}[t!]
	\centering
	\vspace{-0mm}
	\includegraphics[width=0.32\textwidth]{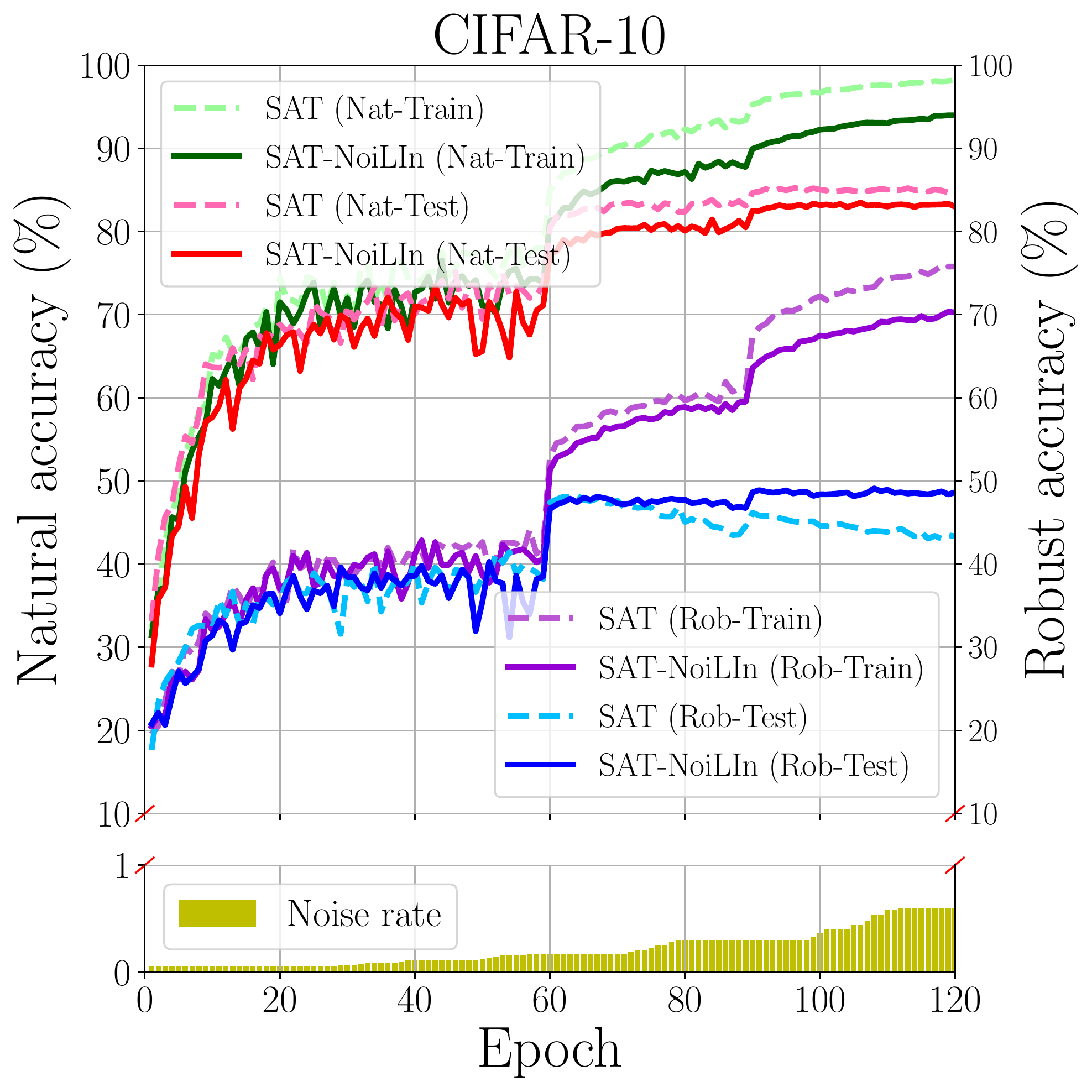}
	\includegraphics[width=0.32\textwidth]{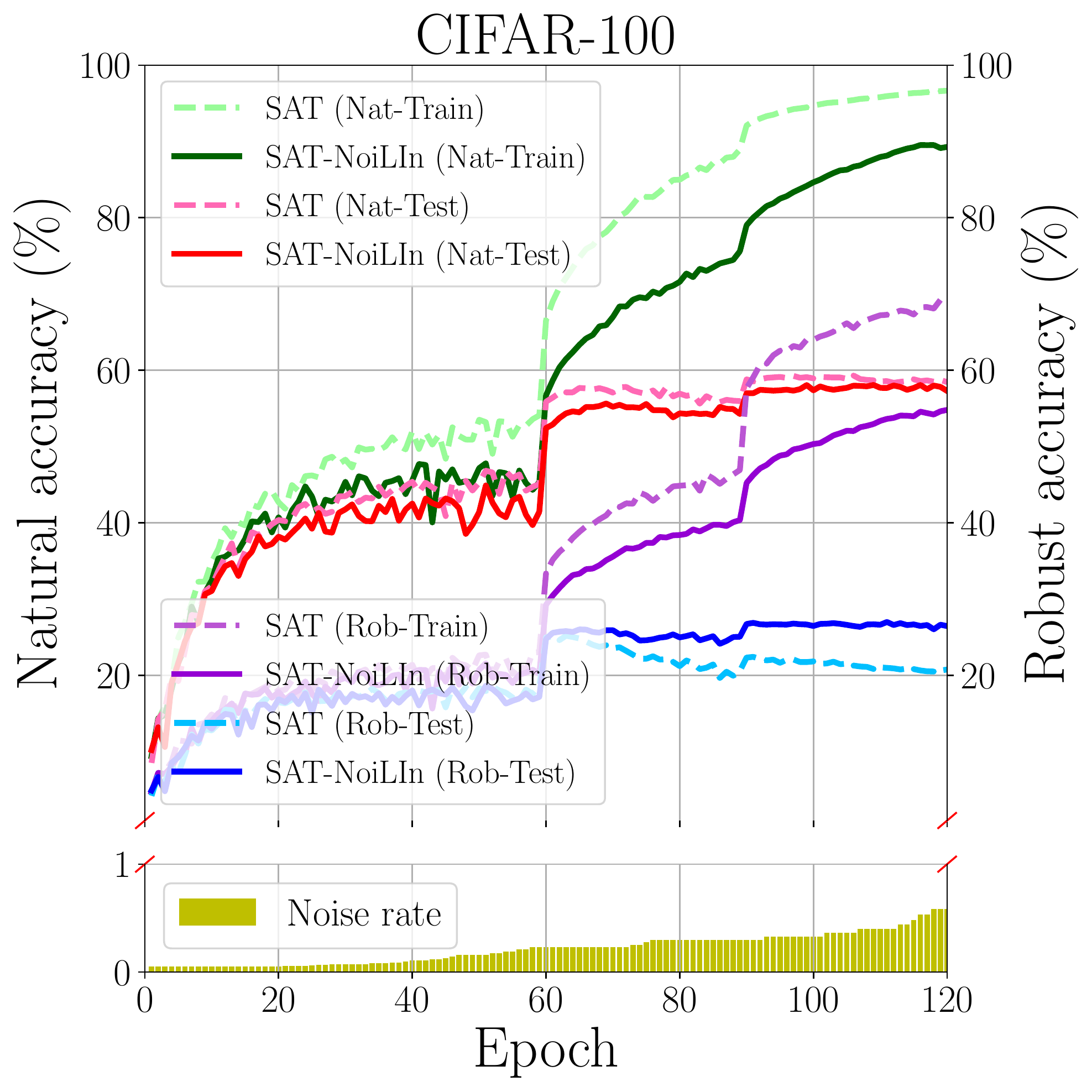} 
	\includegraphics[width=0.32\textwidth]{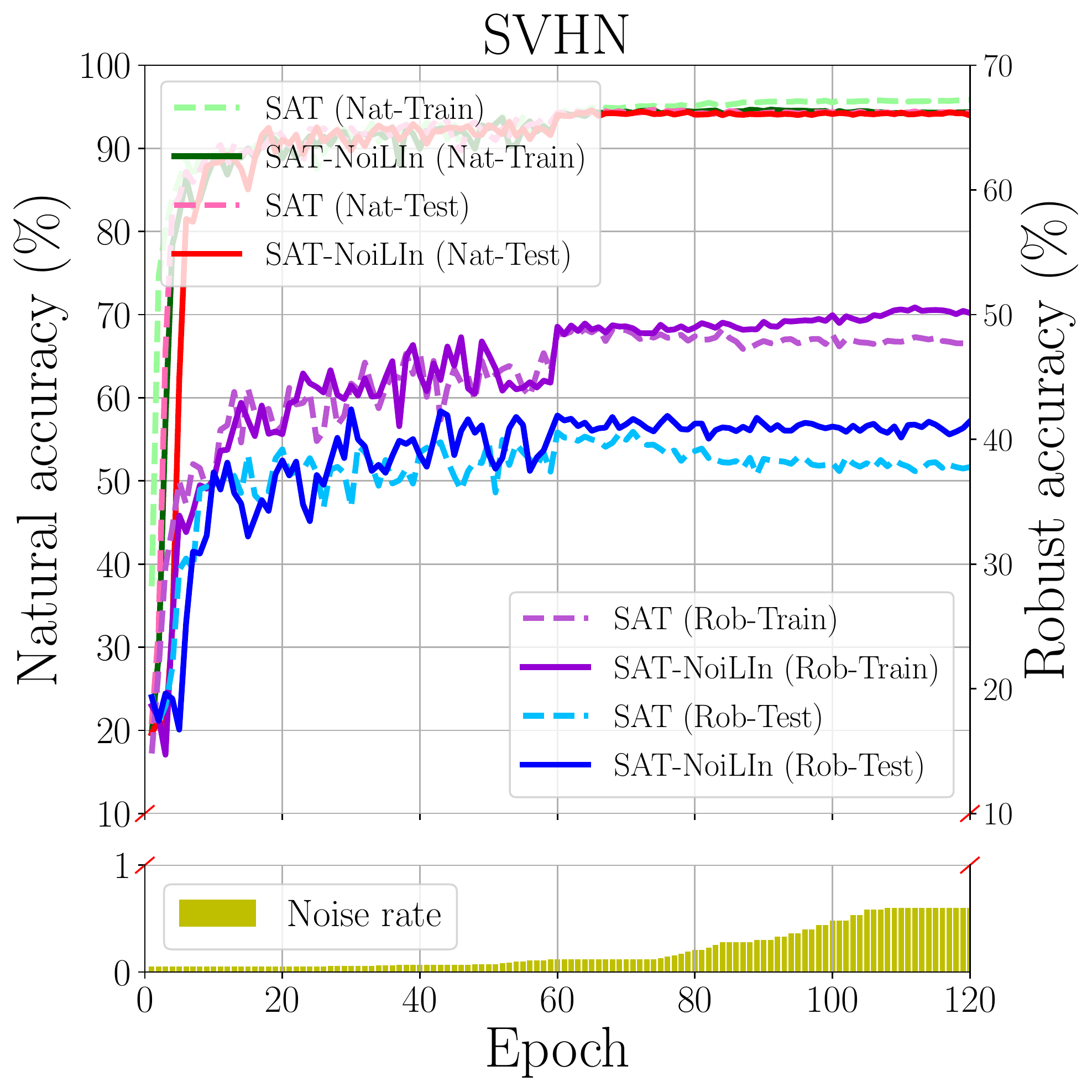}  \\
	\includegraphics[width=0.32\textwidth]{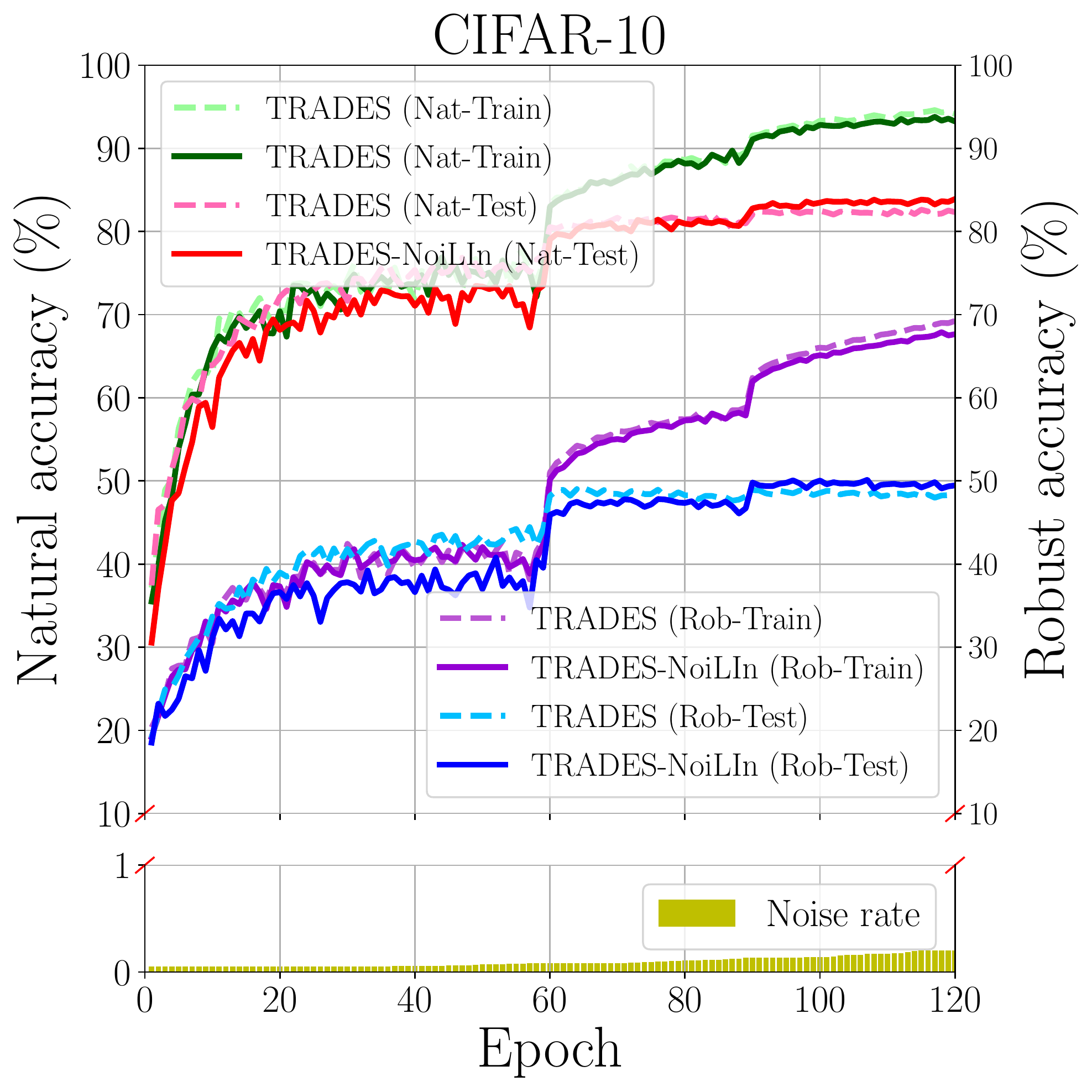} 
	\includegraphics[width=0.32\textwidth]{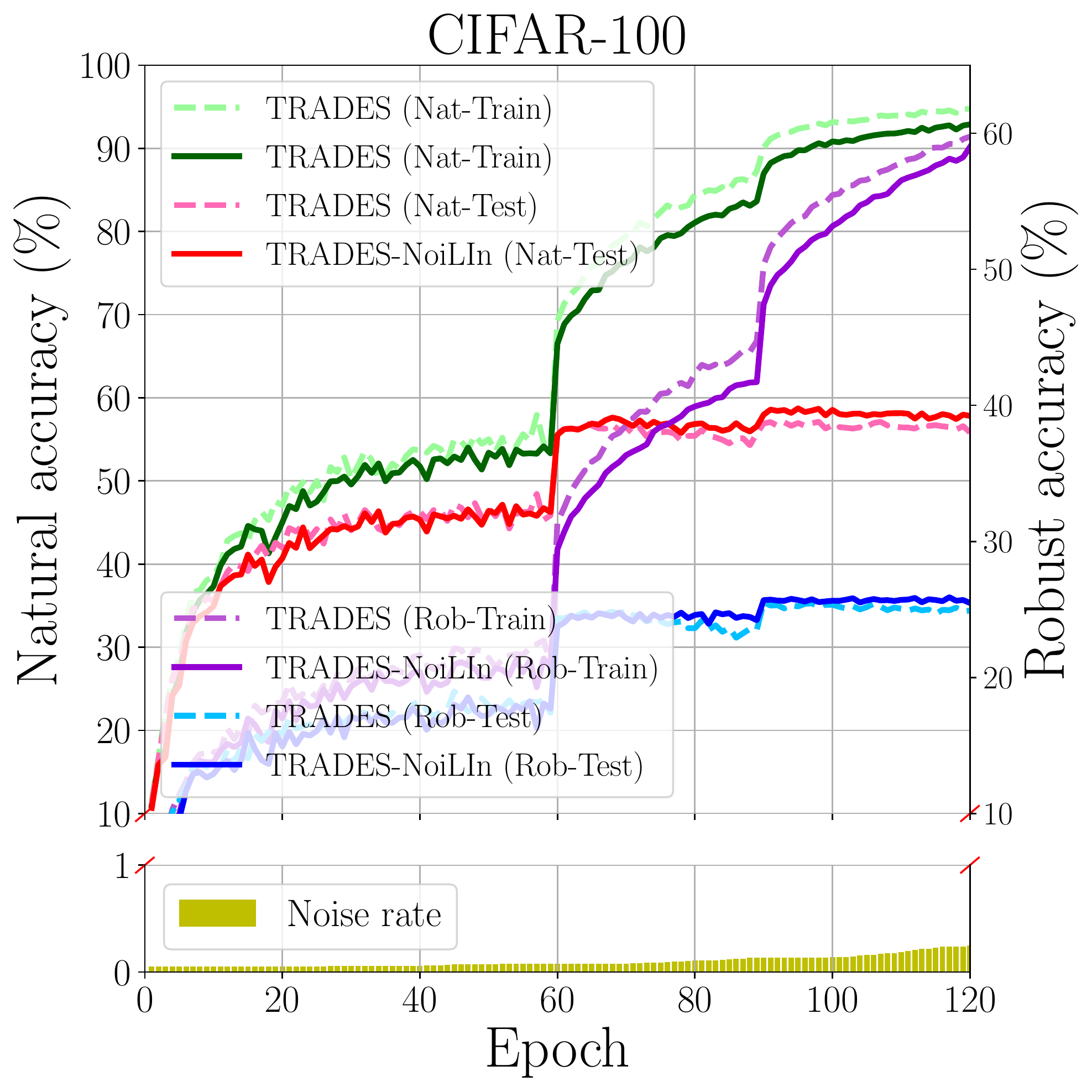}
	\includegraphics[width=0.32\textwidth]{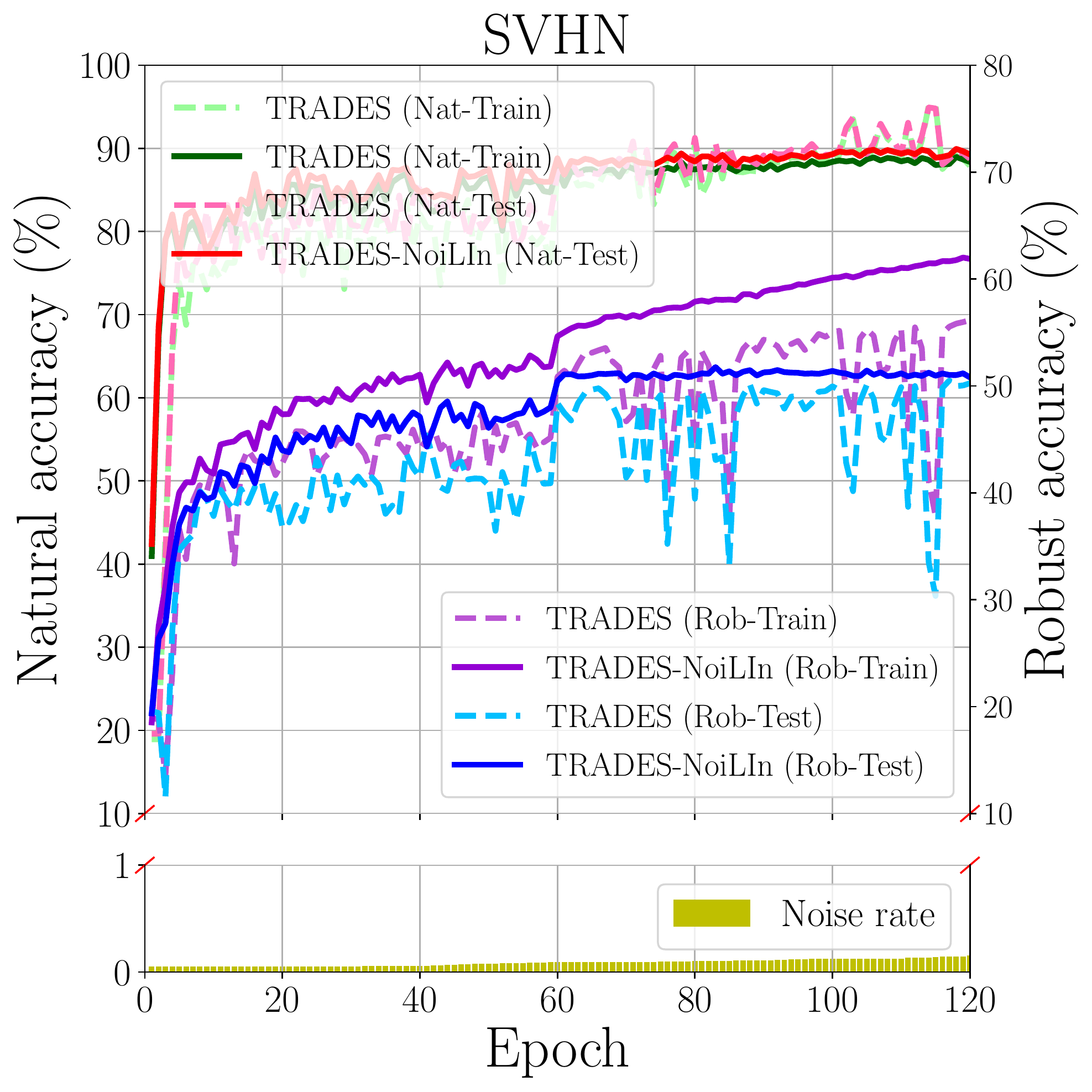}
	\vspace{-2mm}
	\caption{Evaluations on ResNet-18. We compare generalization (green/red lines for training/test data) and robustness (purple/blue lines for training/test data, evaluated by the strongest AA attack) between our NoiLIn (solid lines) and two AT methods such as SAT and TRADES (dash lines) on three datasets---CIFAR-10, CIFAR-100, and SVHN. The yellow histogram below each figure reflects the change of NL injection rate over the training process.}
	\label{fig:robust-overfitting-with-trainingacc}
	\vspace{-2mm}
\end{figure*}

\subsection{Relation with Label Smoothing}
\label{appendix:LS_diff_level}
AT-LS~\cite{cheng2020cat,pang2021bag} has been comprehensively investigated among different smoothing levels $\rho$.
\cite{pang2021bag} pointed out AT with LS only in outer minimization (``AT-LS-outer'') under a mild smoothing level (e.g., $\rho=0.1$) can slightly improve adversarial robustness.
Therefore, we sampled $\rho$ from $\{ 0.05,0.1,0.2,0.3\}$.
Training details (e.g., the optimizer and learning rate schedule) of SAT, AT-LS-outer under different $\rho$, and SAT-NoiLIn (Symmetric) kept same as Section~\ref{sec:robust-overfitting}.
We compared the performance of SAT, AT-LS-outer under different $\rho$, and SAT-NoiLIn (Symmetric) using ResNet-18 on CIFAR-10 dataset in Table~\ref{tab:r18_at_LS_diff_rho}.
We only reported AT-LS-outer ($\rho=0.1$) in Table~\ref{tab:r18_at_LS} for the reason that AT-LS-outer ($\rho=0.1$) achieves the most robustness against AA attacks at the best checkpoint among AT-LS-outer under different $\rho$.

\begin{table}[t!]
	\centering
	\caption{Comparisons between SAT, AT-LS-outer~\cite{pang2021bag} under different $\rho$ and SAT-NoiLIn (Symmetric). We reported the test accuracy of the best checkpoint and that of the last checkpoint as well as the gap between them---``best/last (gap)''.}
	\label{tab:r18_at_LS_diff_rho}
	\begin{tabular}{c|ccc}
		\hline
		Defense               & Natural      & C$\&$W$_{\infty}$-100 & AA \\ \hline
		SAT~\cite{Madry_adversarial_training}   & 81.97/84.76 ($+2.79$) & 49.53/45.12 ($-4.71$) & 48.09/43.30 ($-4.79$)\\
		AT-LS-outer ($\rho=0.05$)    & 82.38/85.19 ($+2.81$) & 49.60/44.95 ($-4.65$) & 47.86/43.01 ($-4.79$) \\
		AT-LS-outer ($\rho=0.1$)     & 82.76/85.15 ($+2.39$)  & 50.06/45.40 ($-4.66$) & 48.60/43.44 ($-5.16$) \\
		AT-LS-outer ($\rho=0.2$)    &  82.80/85.19 ($+2.39$) & 49.87/45.71 ($-4.16$) & 48.42/43.96 ($-4.46$) \\
		AT-LS-outer ($\rho=0.3$)     & 82.45/\textbf{85.70} \textbf{($+$3.25)} & 49.70/45.33 ($-4.37$) & 48.26/43.93 ($-4.33$)\\ 
		SAT-NoiLIn (Symmetric) & \textbf{82.97}/83.86 ($+0.89$)  & \textbf{51.21}/\textbf{50.36} \textbf{($-$0.85)} & \textbf{48.74}/\textbf{47.93} \textbf{($-$0.81)} \\
		\hline
	\end{tabular}
\end{table}

\paragraph{AT-NoiLIn relieves robust overfitting while AT-LS does not.}
Compared with AT-LS under different $\rho$, AT-NoiLIn indeed alleviates robust overfitting. The absolute value of the robust test accuracy gap (in the parentheses at the right two columns of Table~\ref{tab:r18_at_LS}) obtained by AT-NoiLIn is consistently smaller than that obtained by AT-LS under different $\rho$.
In addition,
we observed AT-NoiLIn achieves better performance of robustness evaluations bsed on C$\&$W attacks and AA attacks at the best epoch than AT-LS under all different $\rho$. 

\paragraph{AT-NoiLIn is more computationally efficient than AT-LS.}
We have stated that NoiLIn saves $(C-1)$ log operations for each data compared with LS.
However, when it comes to empirical verification, we found AT-NoiLIn uses the comparable training time with AT-LS. Both AT-NoiLIn and AT-LS using ResNet-18 on CIFAR-10 dataset spent about 256 seconds per training epoch evaluated on a GeForce RTX 2080 Ti GPU. It is owing to that the GPU can parallelly conduct the log operation for each data. That is, for each adversarial data $\xadv$, the GPU only needs to conduct one log operation on all $C$ elements of $p(\xadv)$ simultaneously.  

But when AT-LS is conducted on the dataset which contains more classes ($C$ is larger), such as ImageNet~\cite{deng2009imagenet} dataset, the GPU may need to conduct at least more than one log operation for each data. Since the whole tensor of $p(\xadv)$ is too large for the limited GPU memory, $p(\xadv)$ has to be segmented into several small tensors that satisfying GPU memory for log operation. By contrast, AT-NoiLIn only needs to conduct one log operation on the $\yadv$-th element of $p(\xadv)$ in this case. Therefore, NoiLIn framework is better than AT-LS from the perspective of computational efficiency.

\subsection{NoiLIn under Different Learning Rate Schedulers}
\label{appendix:noilin_diff_lr}
\begin{figure}[h!]
	\centering
	\begin{minipage}[h!]{0.24\textwidth}
		\centering
		\includegraphics[width=0.99\textwidth]{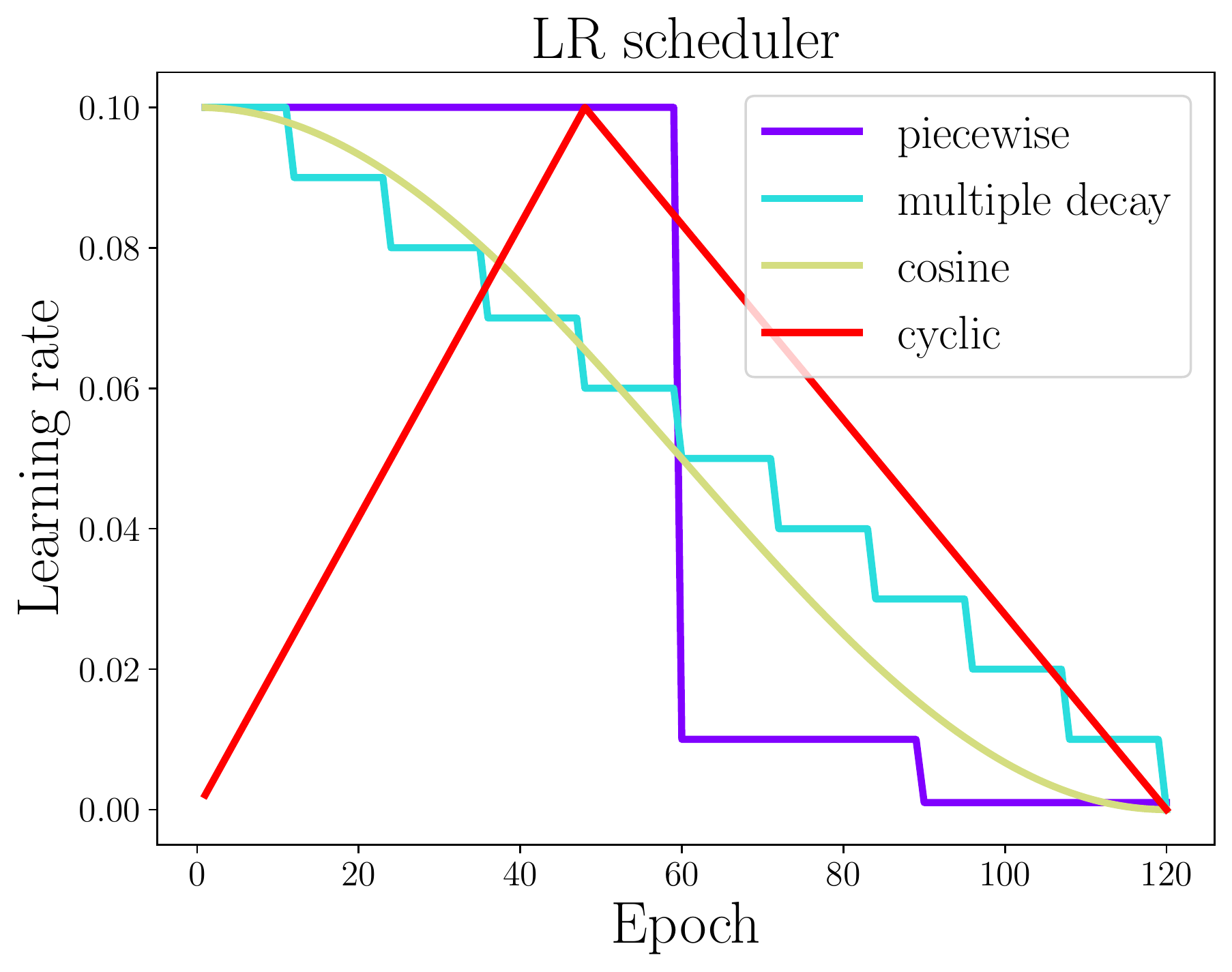}
	\end{minipage}
	\begin{minipage}[h!]{0.74\textwidth}
		\centering
		\includegraphics[width=0.3\textwidth]{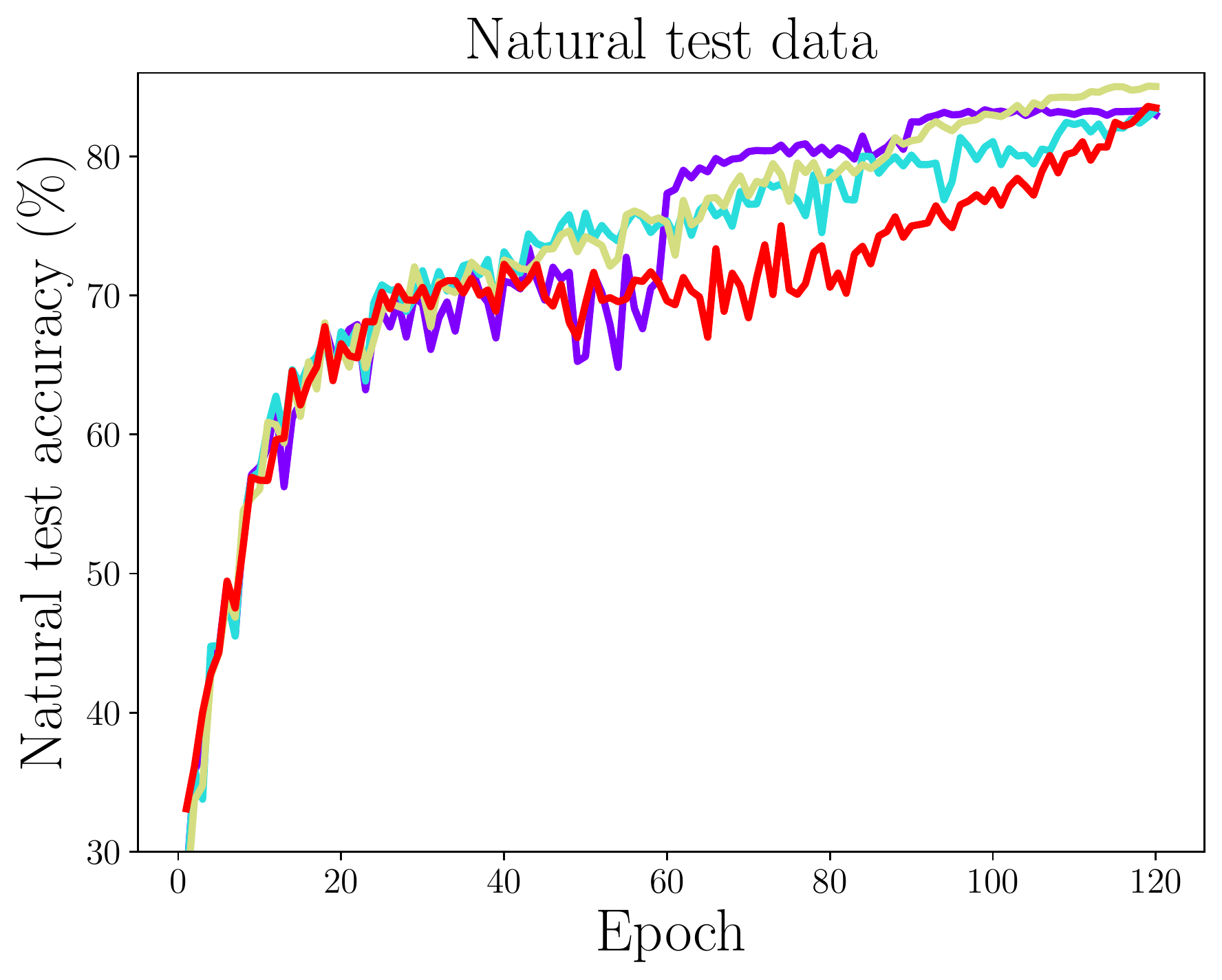}
		\includegraphics[width=0.3\textwidth]{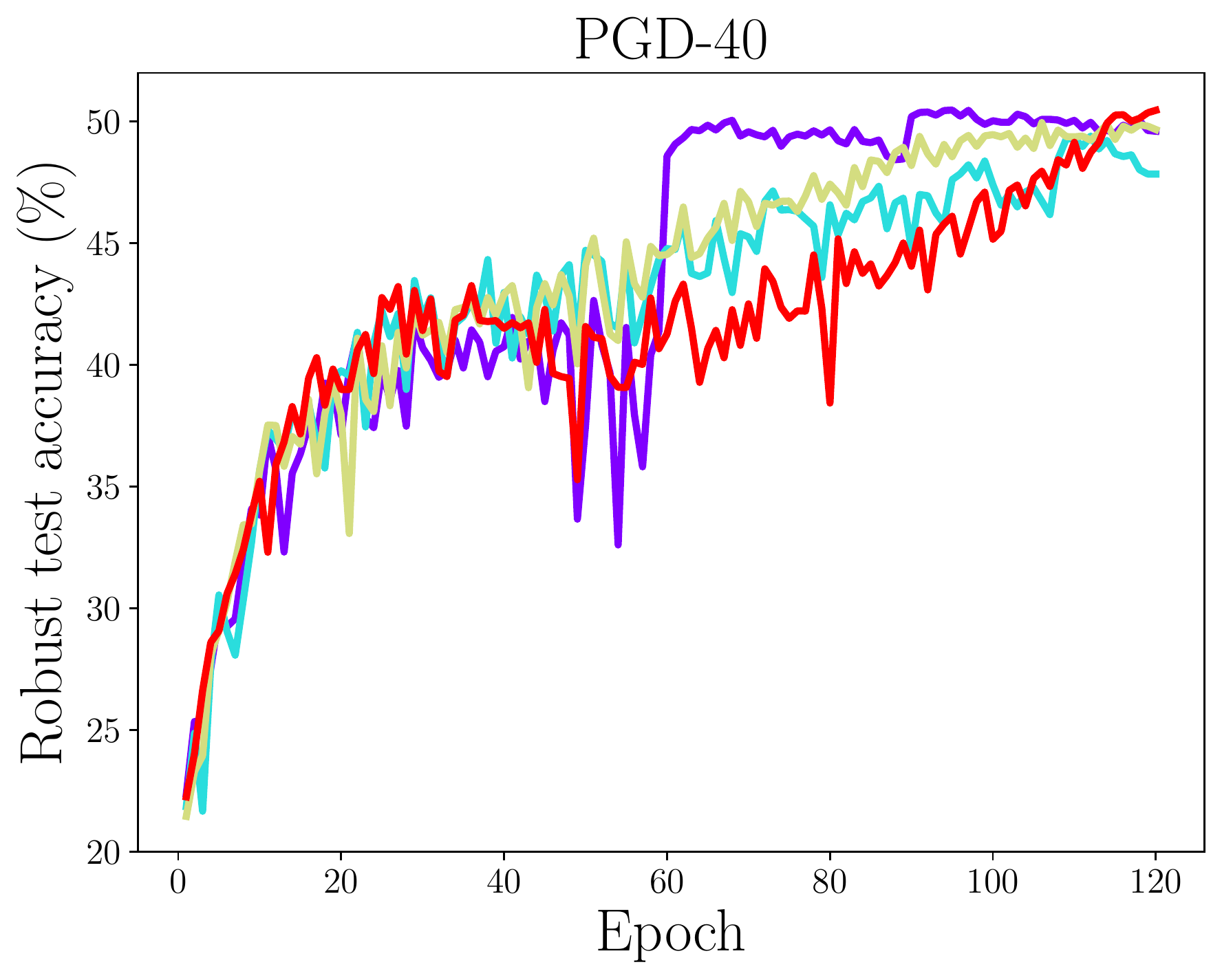}
		\includegraphics[width=0.3\textwidth]{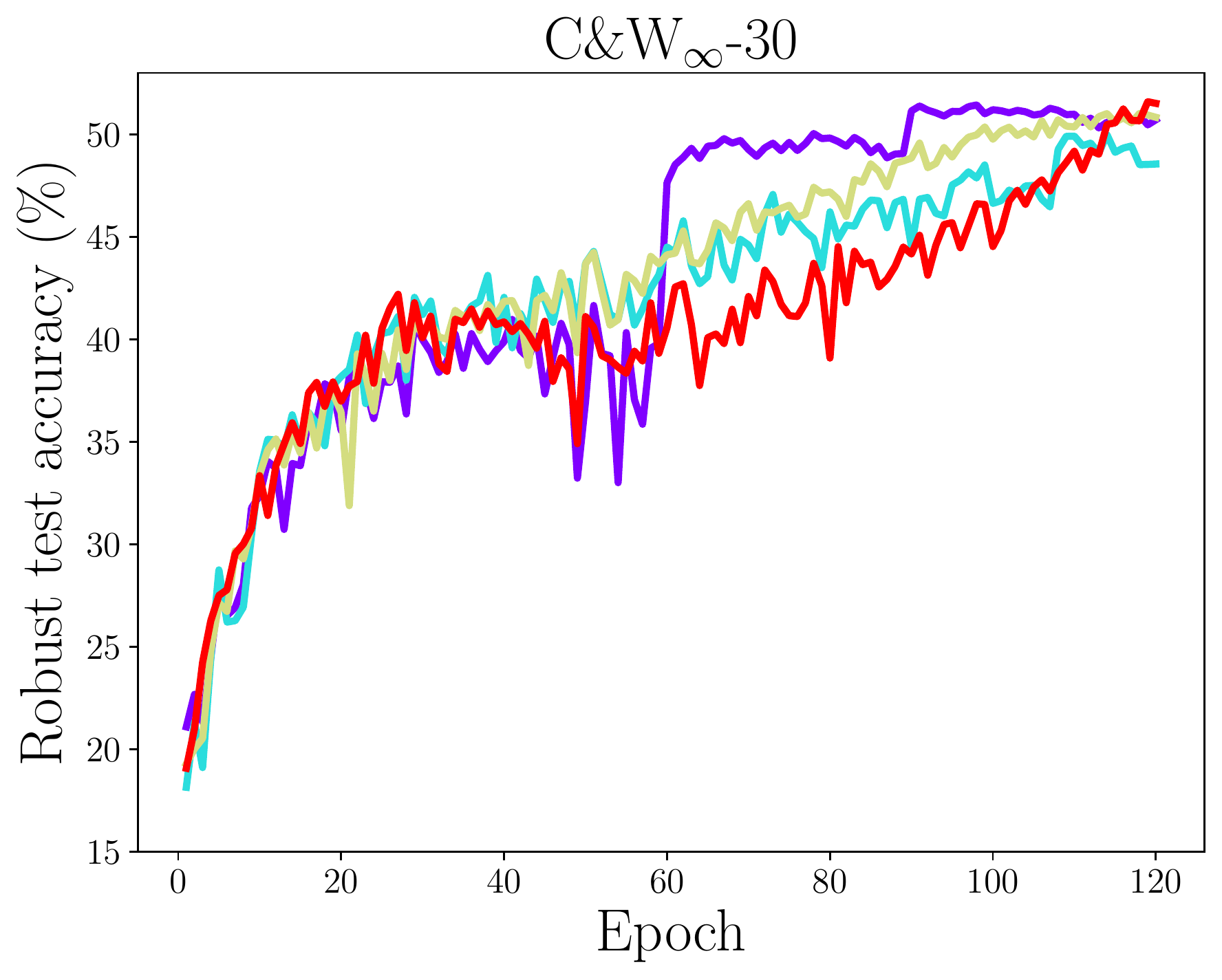}
		
		\includegraphics[width=0.3\textwidth]{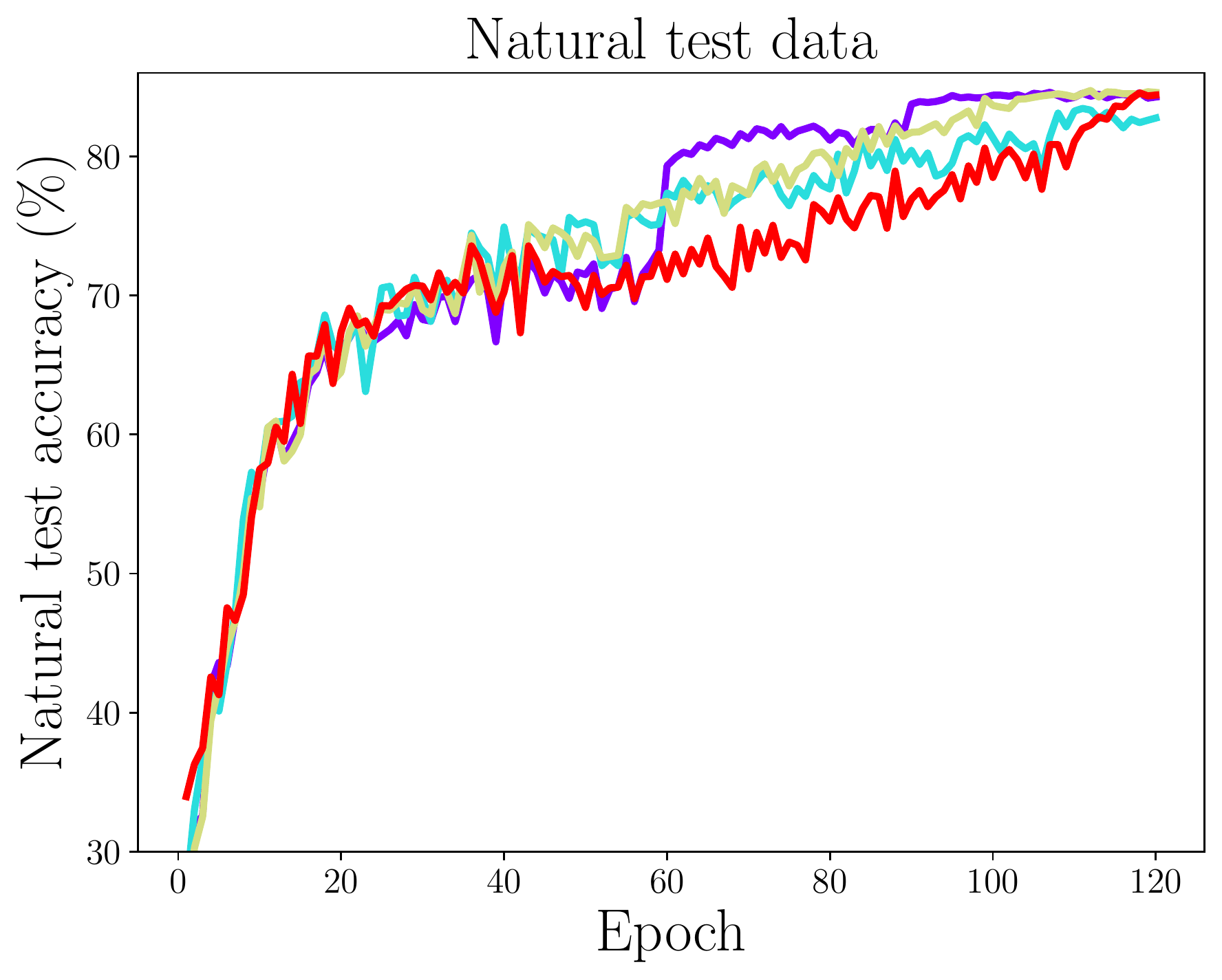}
		\includegraphics[width=0.3\textwidth]{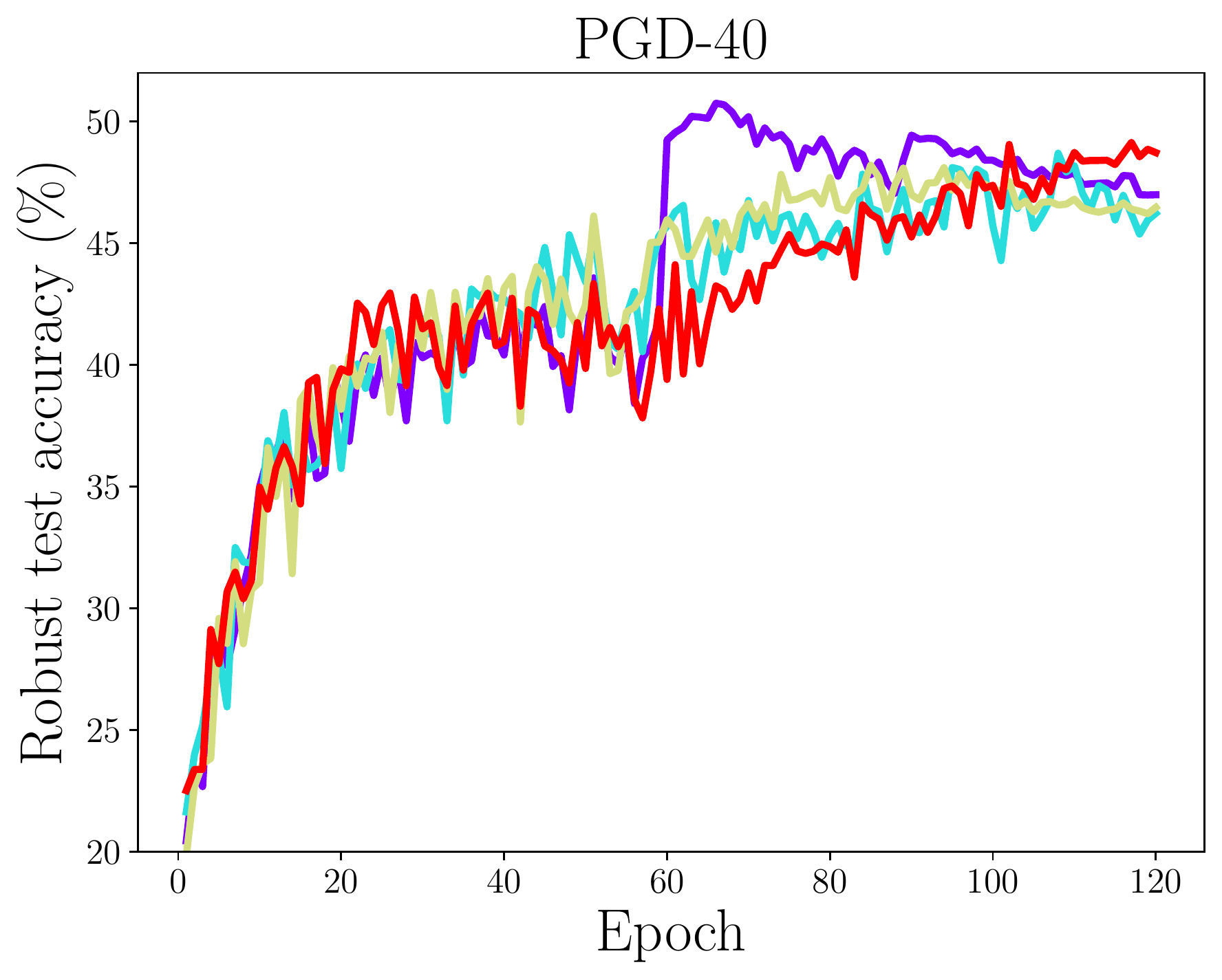}
		\includegraphics[width=0.3\textwidth]{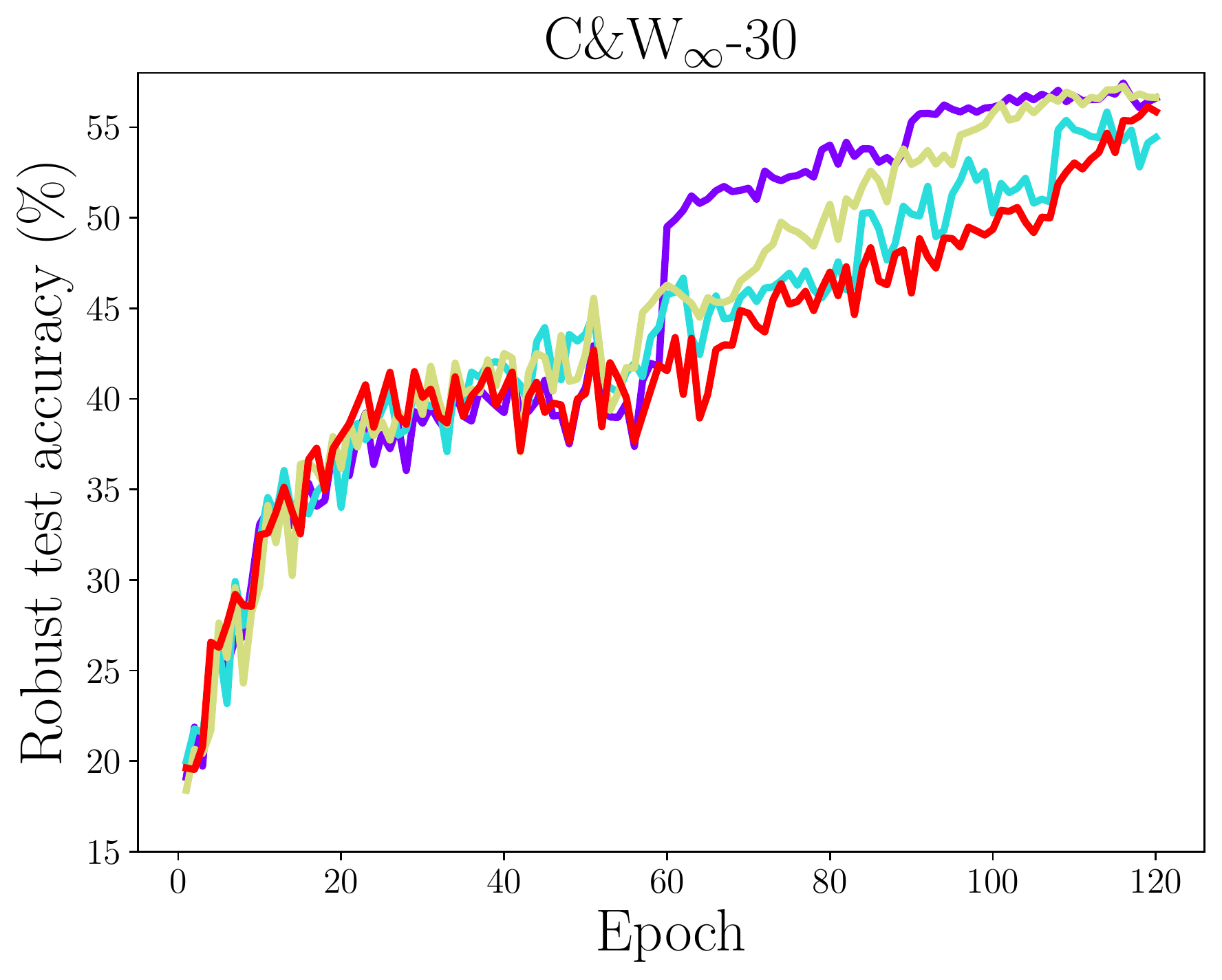}
		\label{fig:AT_NN_lr_schedule}
	\end{minipage}
	\vspace{-3mm}
	\caption{The leftmost panel shows the learning rate w.r.t. epoch under different LR schedulers. The right panels demonstrate the learning curves of AT-NoiLIn under various LR schedulers using symmetric-flipping NL (the upper panels) and pair-flipping NL (the lower panels), respectively.}
	\label{fig:AT_NN_lr_schedule}
	\vspace{-2mm}
\end{figure}

\begin{table}[h!]
	\centering \small 
	\caption{Evaluations of SAT-NoiLIn with pair-flipping NL using ResNet-18 on CIFAR-10 dataset under various LR schedulers. We reported the test accuracy of the best checkpoint and that of last checkpoint as well as the gap between them---``best/last (gap)''.}
	\label{tab:r18_at_diff_lr_pair}
	\resizebox{\columnwidth}{!}{
		\begin{tabular}{c|c|ccc}
			\hline
			Defense & LR scheduler & Natural      & C$\&$W$_{\infty}$-100 & AA \\ \hline
			SAT~\cite{Madry_adversarial_training}    & piecewise  & 81.97/84.76 ($+2.79$)  & 49.53/45.12 ($-4.71$) & 48.09/43.30 ($-4.79$)\\ \hline
			\multirow{4}{*}{\begin{tabular}[c]{@{}c@{}} SAT-NoiLIn\\ (Pair)\end{tabular}} & piecewise & 81.29/84.26 \textbf{($+2.97$)}  & 51.35/\textbf{56.27} \textbf{($+4.92$)} & \textbf{48.30}/46.59 ($-1.71$)\\ 
			& multiple decay & 81.21/82.75 ($+1.54$)& 50.87/54.01 ($+3.14$)& 46.28/45.14 ($-1.14$) \\
			& cosine & 81.69/84.59 ($+2.90$) & 53.09/56.06 ($+2.97$) & 46.75/45.31 ($-1.44$) \\
			& cyclic & \textbf{84.34}/84.34 ($+0.00$) & \textbf{55.58}/55.58 ($+0.00$) & 47.75/\textbf{47.75} \textbf{($+0.00$)}\\
			\hline
		\end{tabular}
		}
\end{table}

We conducted SAT-NoiLIn using pair-flipping NL under different LR schedulers using ResNet-18 on CIFAR-10 dataset.
All the training settings, such as the optimizer and hyperparameters of noise rate, kept exactly same as Section~\ref{sec:robust-overfitting} except the LR scheduler. The learning rate w.r.t. epoch under different LR schedulers is shown in the leftmost panel of Figure~\ref{fig:AT_NN_lr_schedule}. 
We demonstrated the learning curves of AT-NoiLIn (Symmetric) in the upper row of Figure~\ref{fig:AT_NN_lr_schedule} and AT-NoiLIn (Pair) in the lower row of Figure~\ref{fig:AT_NN_lr_schedule} under various LR schedulers. Figure~\ref{fig:AT_NN_lr_schedule} clearly indicates NoiLIn can mitigate robust overfitting under all different LR schedulers. 

Further, we reported the test accuracy of the best checkpoint and that of the last checkpoint as well as the gap between them achieved by NoiLIn using pair-flipping NL in Table~\ref{tab:r18_at_diff_lr_pair}. We observed the gap between test accuracy of the best checkpoint and that of the last checkpoint largely narrows with automatic NL injection, which validates NoiLIn can relieve overfitting.

Moreover, Table~\ref{tab:r18_at_diff_lr} and Table~\ref{tab:r18_at_diff_lr_pair} show AT-NoiLIn under cyclic LR decay simply obtains the best checkpoint at the last epoch (the gap is exactly $+0.00$), which suggests AT-NoiLIn under cyclic LR scheduler can help save the time for selecting the best checkpoint. Note that for selecting the best checkpoint, it is time-consuming to gain the robust test accuracy based on AA attack for ResNet-18 and PGD-10 attack for Wide ResNet over all training epochs.



\begin{figure}[h!]
	\centering
	\includegraphics[width=0.28\textwidth]{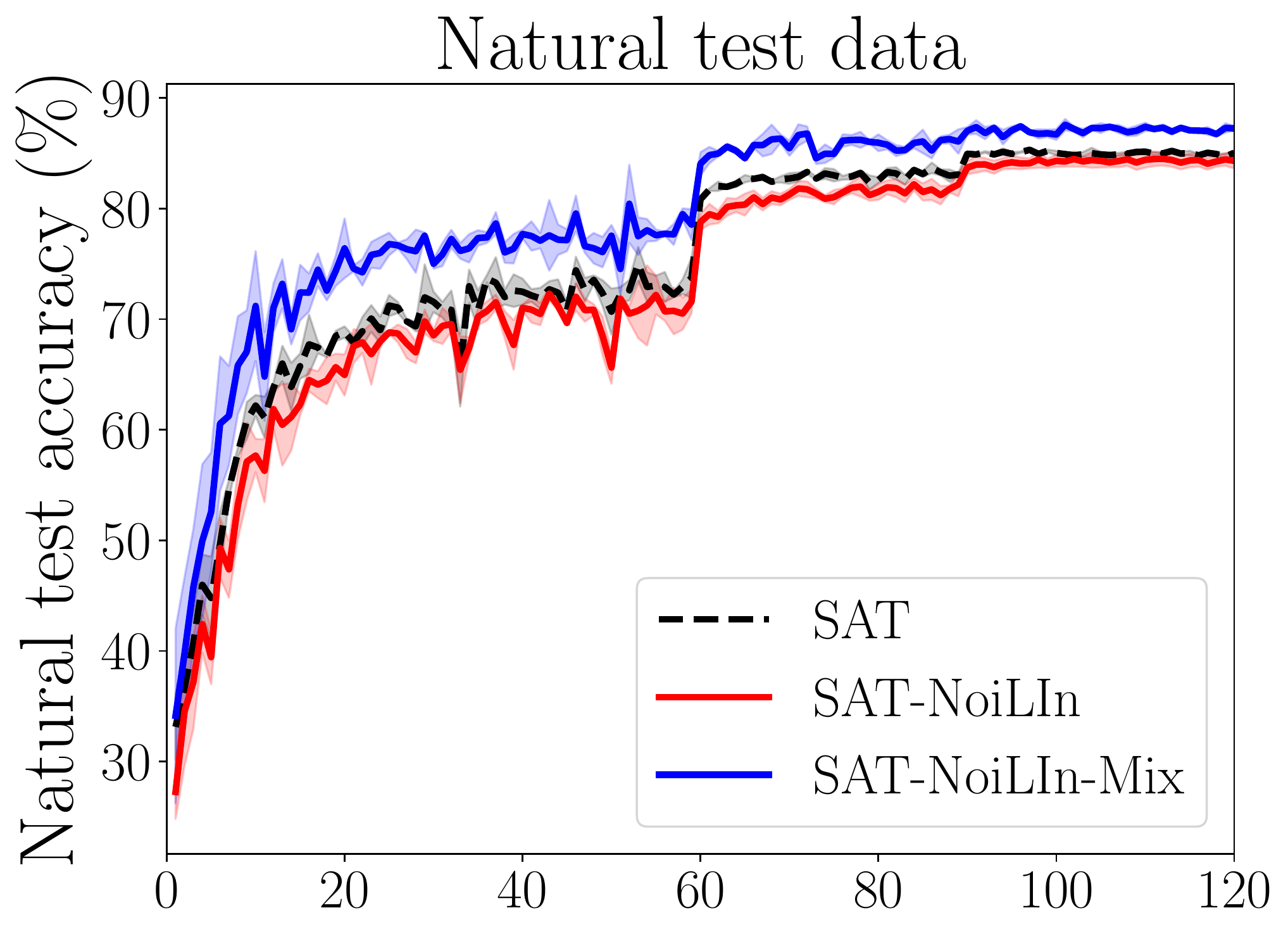}
	\includegraphics[width=0.28\textwidth]{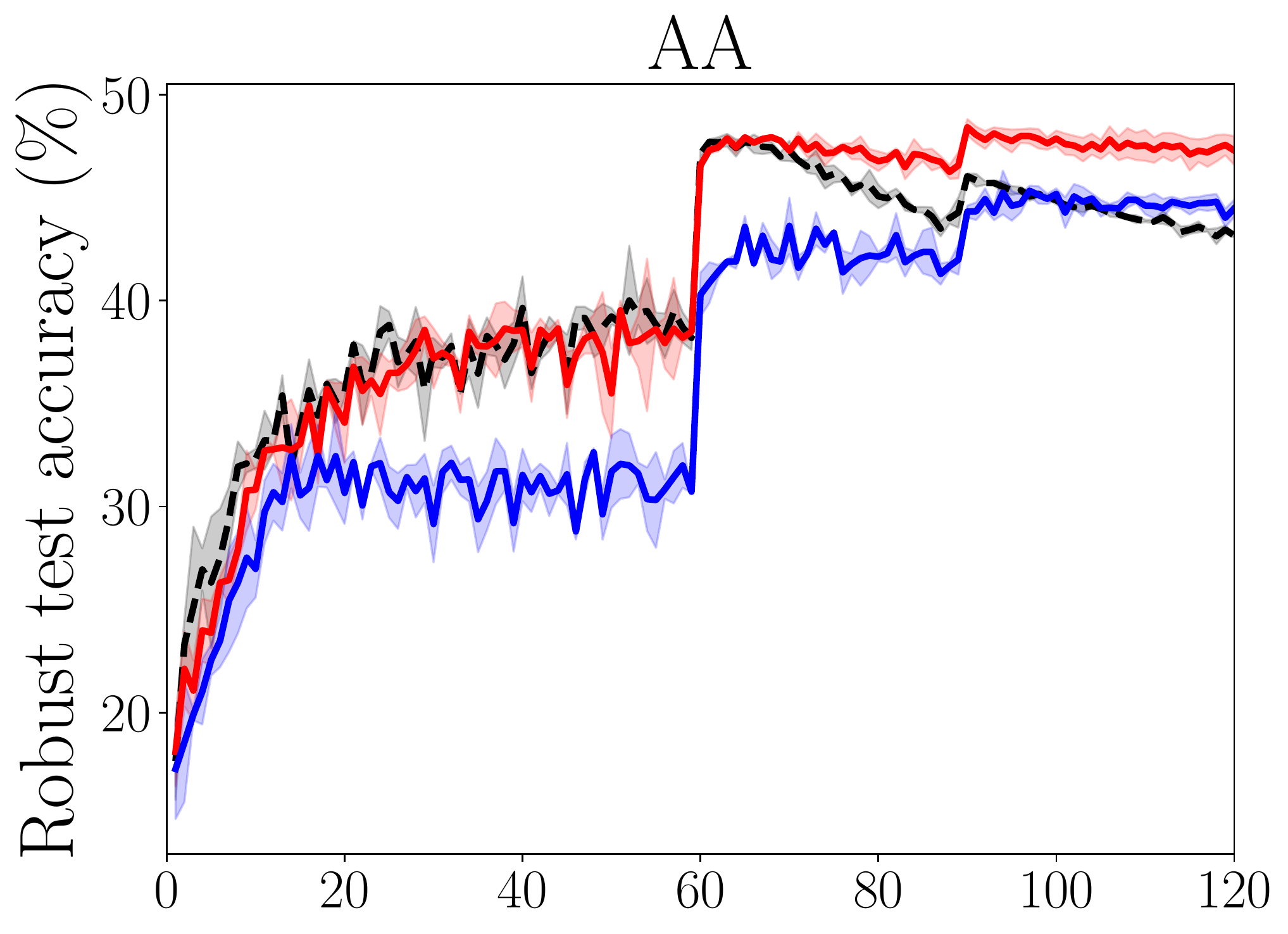}
	\vspace{-0mm}
	\caption{Comparisons between SAT, SAT-NoiLIn, SAT-NoiLIn with batches composed of both natural and adversarial examples (SAT-NoiLIn-Mix).}
	\label{fig:NL_bacth_mix}
	\vspace{-0mm}
\end{figure}

{\subsection{NoiLIn with Batches that Consist of both Natural and Adversarial Examples}
We evaluate the performance of ResNet-18 trained with batches composed of both natural and adversarial examples via NoiLIn (namely, SAT-NoiLIn-Mix). At each epoch, the model parameters are updated via minimizing the sum of natural data's cross-entropy loss and adversarial data's cross-entropy loss. We keep other training configurations (e.g., batch size) same as NoiLIn in Section~\ref{sec:exp}. We repeat the experiments 3 times with different random seeds. In Figure~\ref{fig:NL_bacth_mix}, we show the learning curve of SAT-NoiLIn-Mix.}

{For one thing, Figure~\ref{fig:NL_bacth_mix} shows that the natural test accuracy of SAT-NoiLIn-Mix is consistently above that of SAT and SAT-NoiLIn. This could be attributed to incorporating natural data into each batch, thus helping to improve natural accuracy. For another thing, we find that NoiLIn still can mitigate robust overfitting when each batch is composed of natural and adversarial data, although the robust test accuracy of SAT-NoiLIn-Mix is significantly lower than that of NoiLIn. Therefore, to achieve better robustness, we recommend NoiLIn with batches that consist of only adversarial examples. }

\end{document}